\documentclass[accepted]{uai2025} 
                        
\usepackage[american]{babel}

\usepackage{natbib} 
\usepackage{mathtools} 
\usepackage{booktabs} 
\usepackage{tikz} 

\usepackage{amsmath,amssymb,amsfonts,amsthm, mathtools}

\usepackage{graphicx}
\usepackage{caption}
\graphicspath{{Figures/}} 
\usepackage{wrapfig}

\usepackage{multirow}
\usepackage{booktabs}

\usepackage{algorithm}
\usepackage{algorithmic}

\usepackage{subcaption}

\usepackage{comment}

\def\m{\boldsymbol{m}}
\def\c{\boldsymbol{c}}
\def\z{\boldsymbol{z}}

\def\x{\boldsymbol{x}}
\def\X{\boldsymbol{X}}

\def\0{\boldsymbol{0}}

\def\t{\boldsymbol{\theta}}
\def\e{\boldsymbol{\eta}}

\def\E{\mathbb{E}}

\usepackage{acronym}
\acrodef{vae}[VAE]{Variational Autoencoder}
\acrodef{dvae}[DVAE]{Discrete Variational Autoencoder}
\acrodef{cvae}[CVAE]{Coded Variational Autoencoder}
\acrodef{vi}[VI]{Variational Inference}
\acrodef{elbo}[ELBO]{Evidence Lower Bound}
\acrodef{kl}[KL]{Kullback-Leibler}
\acrodef{iwae}[IWAE]{Importance Weighted Autoencoder}
\acrodef{ldpc}[LDPC]{Low-Density Parity-Check}
\acrodef{vqvae}[VQ-VAE]{Vector Quantized-Variational Autoencoder}
\acrodef{bp}[BP]{Belief Propagation}
\acrodef{nn}[NN]{Neural Network}
\acrodef{ais}[AIS]{Annealed Importance Sampling }
\acrodef{sis}[SIS]{Sequential Importance Sampling}
\acrodef{mse}[MSE]{Mean Squared Error}
\acrodef{psnr}[PSNR]{Peak Signal-To-Noise Ratio}
\acrodef{wer}[WER]{Word Error Rate}
\acrodef{ber}[BER]{Bit Error Rate}
\acrodef{ecc}[ECC]{Error Correcting Code}
\acrodef{cnn}[CNN]{Convolutional Neural Networks}
\acrodef{se}[SE]{Squeeze-and-Excitation}
\acrodef{cdf}[CDF]{Cumulative Distribution Function}
\acrodef{mlp}[MLP]{Multilayer Perceptron}
\acrodef{map}[MAP]{Maximum a Posteriori}
\acrodef{iwae}[IWAE]{Importance Weighted Autoencoder}
\acrodef{ldpc}[LDPC]{Low Density Parity Check}
\acrodef{siso}[SISO]{soft-in soft-out}
\acrodef{cdf}[CDF]{Cumulative Density Function}
\acrodef{fid}[FID]{Fréchet Inception Distance}
\acrodef{rd}[RD]{rate-distortion}
\acrodef{ll}[LL]{log-likelihood}
\acrodef{ddpm}[DDPM]{Denoising Diffusion Probabilistic Models}
\acrodef{ssim}[SSIM]{Structural Similarity Index Measure}

\usepackage[page,header]{appendix}
\usepackage{titletoc}

\title{Improved Variational Inference in Discrete VAEs using Error Correcting Codes}

\author[1,2]{\href{mailto:<martinez-garcia@cs.uni-saarland.de>?Subject=Improved Variational Inference in Discrete VAEs using Error Correcting Codes} Mar\'ia Mart\'inez-Garc\'ia{}}
\author[3]{Grace Villacr\'es}
\author[4]{David Mitchell}
\author[2,5]{Pablo M. Olmos}

\affil[1]{%
    Dept. of Computer Science\\
    Saarland University\\
    Saarbrücken, Germany
}
\affil[2]{%
    Dept. of Signal Theory and Communications\\
    Universidad Carlos III de Madrid\\
    Leganés, Spain
}

\affil[3]{%
    Dept. of Signal Theory and Communications, Universidad Rey Juan Carlos\\
    Fuenlabrada, Spain
}
\affil[4]{%
    Klipsch School of Electrical and Computer Engineering, New Mexico State University, Las Cruces, USA
}

\affil[5]{%
Instituto de Investigación Sanitaria Gregorio Marañón, Madrid, Spain
}
  
\begin{document}
\maketitle

\begin{abstract}

Despite advances in deep probabilistic models, learning discrete latent representations remains challenging. This work introduces a novel method to improve inference in discrete Variational Autoencoders by reframing the inference problem through a generative perspective. We conceptualize the model as a communication system, and propose to leverage Error Correcting Codes (ECCs) to introduce redundancy in latent representations, allowing the variational posterior to produce more accurate estimates and reduce the variational gap. We present a proof-of-concept using a Discrete Variational Autoencoder with binary latent variables and low-complexity repetition codes, extending it to a hierarchical structure for disentangling global and local data features. Our approach significantly improves generation quality, data reconstruction, and uncertainty calibration, outperforming the uncoded models even when trained with tighter bounds such as the Importance Weighted Autoencoder objective. We also outline the properties that ECCs should possess to be effectively utilized for improved discrete variational inference. 

\end{abstract}

\section{Introduction}\label{sec:intro}

Discrete latent space models represent data using a finite set of features, making them particularly well-suited for representing inherently discrete data modalities. Additionally, they provide benefits in terms of interpretability and computational efficiency compared to continuous representations \citep{vqvae, gumbel-softmax, dvae_sharp}. Consequently, recent advancements in deep generative models have increasingly embraced discrete latent representations \citep{vqvae2, dalle, latent_diffusion}. 

However, model training and inference of low-dimensional discrete latent representations remains technically challenging due to non-differentiability, which often requires the use of approximations that can lead to unstable optimization \citep{gumbel-softmax, reinforce_loo_2}. 
Currently, \acp{vqvae} \citep{vqvae, vqvae2} stand out as state-of-the-art discrete latent variable models. They rely on a non-probabilistic encoder trained with a straight-through estimator, which lacks uncertainty quantification in the latent space. In contrast, fully probabilistic discrete \acp{vae} \citep{vae} employ continuous relaxations such as Gumbel-Softmax \citep{gumbel-softmax} or Concrete \citep{concrete} for differentiable sampling \citep{dalle, hierarchical_dvae}. However, these methods can be unstable, as gradient variance is highly sensitive to the temperature parameter. The \ac{dvae} framework \citep{dvae, dvae_sharp, dvae++} addresses this issue by augmenting binary latent representations with continuous variables, allowing for stable reparameterization after marginalizing the latent bits. In this setting, Boltzmann machines serve as discrete priors to enhance model flexibility, at the cost of degraded interpretability.

This work introduces a novel approach to improve variational inference in discrete latent variable models, particularly discrete \acp{vae}, by offering an alternative perspective on the inference problem. Rather than viewing \acp{vae} as lossy compression methods employed to minimize data reconstruction error, we adopt a generative perspective, where inference seeks to recover the latent vector that generated a given observation. This allows us to conceptualize the generative model as a communication system, where we reinterpret latent vectors as \textit{messages} transmitted through a nonlinear \textit{channel} given by the \ac{vae} decoder, and then recovered by the \ac{vae} encoder. Based on this analogy, we propose to incorporate \acp{ecc} to introduce redundancy into latent representations before they are processed by the decoder to generate data. 

\begin{figure*}[ht]
    \centering
     \begin{subfigure}[b]{0.51\textwidth}
         \centering
         \includegraphics[width=1\linewidth]{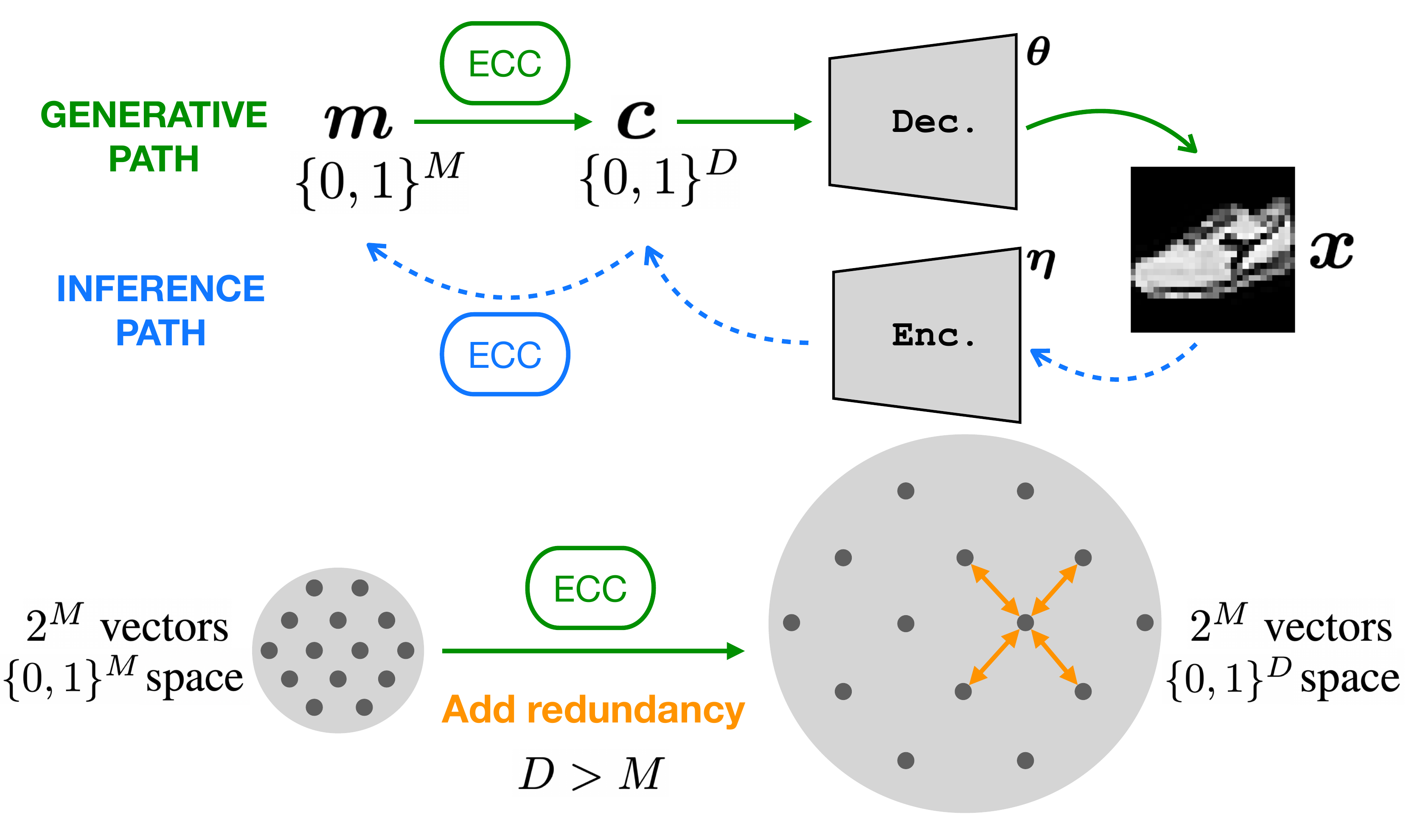}
         \caption{}
         \label{fig:fancy_scheme}
    \end{subfigure}
    \centering
     \begin{subfigure}[b]{0.43\textwidth}
         \centering
         \includegraphics[width=1\linewidth]{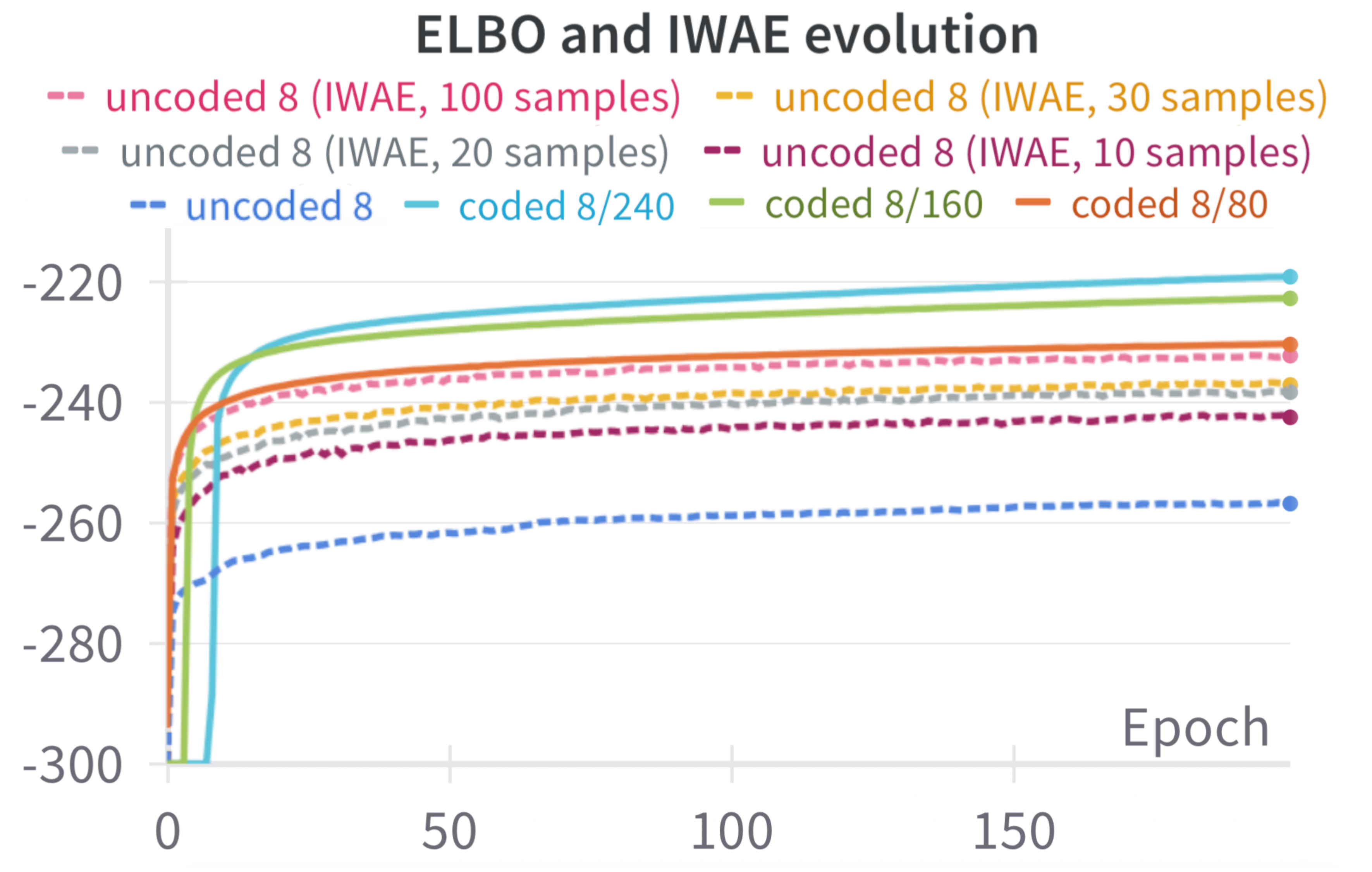}
         \vfill
         \caption{}
         \label{fig:loss_evol}
    \end{subfigure}

    \caption{\textbf{ECC within a DVAE}. Fig. (a) illustrates the method and how adding redundancy transforms a dense $M$-dimensional latent space into a sparse $D$-dimensional one with $2^M$ valid vectors, enabling error correction in the variational posterior via minimum distance. Fig. (b) compares the ELBO and IWAE objectives for coded and uncoded models, where coding schemes with $M=8$ and $D=80,160,240$ introduce $D/8$ redundancy bits per latent bit.}
    \label{fig:fancy_figure}
\end{figure*}

Our approach builds on well-established techniques from digital communications and data storage, where information is \textit{protected} using \acp{ecc} before transmission or storage to reduce the overall error rate during recovery \citep{Moon}. Notably, Shannon's landmark work demonstrated that properly designed error correction schemes can achieve arbitrarily low error rates \citep{shannon}. We use \acp{ecc} to deterministically introduce controlled redundancy to the latent vectors, increasing their Hamming distance and creating a higher-dimensional but sparser space where only a subset of vectors are valid. As illustrated in Fig. \ref{fig:fancy_scheme}, this increased separation between latent codes allows for some errors in inference while still enabling recovery of the correct latent code by identifying the closest valid vector. Hence, this structure facilitates error correction during inference. 

The integration of \acp{ecc} into discrete latent variable models is a novel approach that remains compatible with existing training methods for discrete generative models. Note that, while deep learning has been widely explored in digital communications \citep{DeepCom3, DeepCom4, DeepCom5, DeepCom1, DeepCom2}, the use of \acp{ecc} as a design tool in machine learning remains largely unexplored. This work introduces a new research direction, leveraging communications theory to enhance deep generative models.

In our experiments, we use a simplified version of \ac{dvae}++ \citep{dvae++} (referred to as \textit{uncoded} \ac{dvae}) and show that the added redundancy enables a more accurate variational approximation to the true posterior, using simple independent priors. We highlight that error rates and variational gaps are linked through bounds derived from mismatch hypothesis testing, showing that minimizing the variational gap tightens the upper bound on the error rate \citep{error_bounds_f_divergence, error_bounds_refined}. Across different datasets, our results indicate that the \ac{dvae} with \acp{ecc} (Coded-\ac{dvae}) leads to reduced error rates in inference, resulting in smaller variational gaps. This improvement translates into superior generation quality, improved data reconstruction, and critically calibrated uncertainty in the latent space. Notably, Fig. \ref{fig:loss_evol}  shows that Coded-\ac{dvae} achieves a significantly tighter training bounds even when the baseline is trained using the \ac{iwae} objective \citep{iwae}. 

\textbf{Contributions.} The primary contribution of this work is presenting a new perspective on the inference problem by framing the generative model as a communication system. We present proof-of-concept results that show how training discrete latent variable models can be improved by incorporating \ac{ecc} techniques, an approach that, to the best of our knowledge, is novel in the existing literature. We then introduce a coded version of a discrete \ac{vae}, utilizing block repetition codes, and show that encoding and decoding can be done with linear complexity. Our results show that the Coded-\ac{dvae} model significantly reduces error rates during inference, leading to better data reconstruction, generation quality, and improved uncertainty calibration in the latent space. We further generalize this approach to other coding schemes and propose a hierarchical structure inspired by polar codes \citep{polarcodes}, which effectively separates high-level information from finer details.  \footnote{Code:  \href{https://github.com/mariamartinezgarcia/codedVAE}{https://github.com/mariamartinezgarcia/codedVAE}}

\section{Our baseline: the uncoded DVAE}\label{uncoded_case}

This section introduces a simplified version of the \ac{dvae}++ \citep{dvae++} (\textit{uncoded} \ac{dvae}), which serves as the foundational model upon which the subsequent aspects of our work are constructed. A key advantage of this framework is its stable training process, facilitated by reparameterization through smoothing transformations, in contrast to more unstable methods like Gumbel-Softmax \citep{gumbel-softmax}. To ensure that any performance improvements in the Coded-\ac{vae} stem solely from the coding scheme rather than a complex prior, we adopt a simple independent prior.

\begin{wrapfigure}{r}{0.13\textwidth}
    \centering
    \includegraphics[width=0.1\textwidth]{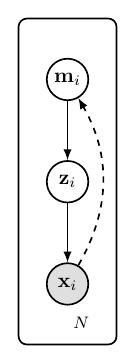}
    \caption{Graphical model of the uncoded DVAE.}
    \label{fig:DAG_uncoded}
\end{wrapfigure}

Let $\X=\{\x_0,\ldots,\x_N\}$ denote a collection of unlabeled data, where $\x_i \in \mathbb{R}^K$ represents a $K$-dimensional feature vector. While \citet{dvae}, \citet{dvae++} and \citet{dvae_sharp} use Boltzmann machine priors, we consider a generative probabilistic model characterized by a simple low-dimensional binary latent variable $\m \in \{0,1\}^M$ comprising independent and identically distributed (i.i.d.) Bernoulli components $p(\m) = \prod_{j=1}^{M} p(m_j) = \prod_{j=1}^{M}\text{Ber}(\nu)$, where $\nu = p(m_j=1)$. Since backpropagation through discrete samples is generally not possible, a smoothing transformation $p(\z|\m)=\prod_{j=1}^{M}p(z_j|m_j)$ of these binary variables is introduced. Following \citet{dvae++}, we introduce truncated exponential distributions given by
\begin{align}\label{smoothing_transformations_dvae}
   p(z|m=1) = \frac{e^{\beta (z-1)}}{Z_\beta}, \quad  p(z|m=0) = \frac{e^{-\beta z}}{Z_\beta},
\end{align}
for $m\in\{0,1\}$, $z\in[0,1]$, and $Z_\beta=(1-e^{-\beta})/\beta$. The parameter $\beta$ serves as an inverse temperature term, similar to the one in the Gumbel-Softmax relaxation \citep{gumbel-softmax}. Given the simplicity of the defined binary prior, the complexity of the model is primarily determined by the likelihood function $p_{\t}(\x|\z) = p(f_{\t}(\z))$, the decoder function, where the likelihood is a \ac{nn} (referred to as the decoder) with parameter set $\t$.

\subsection{Variational family and inference}
Following \citet{dvae}, we assume an amortized variational family of the form $q_{\e}(\m,\z|\x) = q_{\e}(\m|\x)p(\z|\m)$, where 
\begin{align}\label{VIuncoded} 
q_{\e}(\m|\x) &= \prod_{j=1}^M \text{Ber}(m_j; q_j) = \prod_{j=1}^M \text{Ber}(m_j; g_{j,\e}(\x))
\end{align}
where $\text{Ber}(m_j;q_j)$ is a Bernoulli with parameter $q_j=q_{\e}(m_j=1|\x)$ and $g_{\e}(\x)$ is a \ac{nn} (referred to as the data encoder) with parameter set $\e$. Inference is achieved by maximizing the \ac{elbo}, which can be expressed as
\begin{equation}
    \begin{gathered}
    \log p(\x)\geq\int q_{\e}(\m,\z|\x) \log\left(\frac{p_{\t}(\x,\z,\m)}{q_{\e}(\m,\z)}\right)d\m d\z \\
    = \E_{q_{\e}(\m,\z|\x)}\log p_{\t}(\x|\z) - \mathcal{D}_{KL}\big(q_{\e}(\m|\x)||p(\m)\big),
    \end{gathered}\label{elbo_dvae}
\end{equation}

where the first term corresponds to the reconstruction of the observed data and the second is the \ac{kl} Divergence between the variational family and the prior distribution, which acts as a regularization term. This can be computed in closed form as $\mathcal{D}_{KL}\big(q_{\e}(\m|\x)||p(\m)\big) = \sum_{j=1}^{M} \Big[q_{j} \log\frac{q_{j}}{\nu}+(1-q_{j})\log\frac{1-q_{j}}{1-\nu}\Big]$, 
with $\nu = p(m_j=1)$. Since $p_{\t}(\x|\z)$ does not depend on the binary latent variable $\m$, we can marginalize the posterior distribution as 
\begin{equation}
    \begin{gathered}
    q_{\e}(\z|\x) = \prod_{j=1}^{M} q_{\e}(z_j|\x), \\
     \quad q_{\eta}(z_j|\x) = \sum_{k=0}^1 q_{\eta}(m_j=k|\x)p(z_j|m_j=k).
     \end{gathered}
\end{equation}
As shown in \citet{dvae++}, the corresponding inverse \ac{cdf} is given by
\begin{equation}\label{CDF_dvae}
    F^{-1}_{q_{\e}(z_j|\x)}(\rho) =  -\frac{1}{\beta}\log\left(\frac{-b+\sqrt{b^2-4c}}{2}\right),
\end{equation}
where $b = \big(\rho+e^{-\beta}(q_{j}-\rho)\big)/(1-q_{j})-1$ and $c=-[q_{j}e^{-\beta}]/(1-q_{j})$. Equation \eqref{CDF_dvae} defines a differentiable function that converts a sample $\rho$ from an independent uniform distribution $\mathcal{U}(0,1)$ into a sample from $q_{\e}(\z|\x)$. Thus, we can apply the reparameterization trick to sample from the latent variable $\z$ and optimize the ELBO. 

\section{On the variational gap and the inference error}\label{inf_as_communication}

While \acp{vae} are typically viewed as lossy compression methods, this work introduces an alternative perspective on the inference problem from a generative viewpoint: we first sample a latent vector $\m$, generate an observation $\x$, and aim to minimize the error rate when recovering $\m$ from $\x$. This requires approximating the true, unknown posterior distribution $p_{\t}(\m|\x)$ with a sufficiently accurate proposed distribution $q_{\e}(\m|\x)$. The gap between $q_{\e}(\m|\x)$ and $p_{\t}(\m|\x)$ is precisely the variational gap, which can be related to the error rate in inference, i.e. comparing the true $\m$ with samples drawn from $q_{\e}(\m|\x)$.

In statistical classification, particularly in the context of multiple hypothesis testing, classification error is the primary performance metric. Bayes' decision rule (also known as Bayes' test) minimizes classification error, assuming the true probability distributions governing the classification problem are known. In the case of model mismatch, e.g., when a posterior distribution is learned using variational inference, statistical bounds can be used to relate the classification error and the model estimation error \citep{error_bounds_f_divergence, error_bounds_refined}. In particular, for a generative model $\x\sim p_{\t}(\x|\m)$ with discrete latent vector $\m$, let $1-A_{\t}$ be the error rate when estimating $\m$ from $\x$ using the true posterior $p_{\t}(\m|\x)$ and $1-A_{\e}$ be the error rate using the variational approximate posterior $q_{\e}(\m|\x)$. Then, $\Delta \doteq A_{\t} - A_{\e}$ 
can be bounded as
\begin{align}\label{bound1}
    0\leq\Delta^2 &\leq 1 - e^{-2\mathcal{D}_{KL}(q_{\e}(\m,\x)||p_{\t}(\m,\x))},
\end{align}
where $\mathcal{D}_{KL}(q_{\e}(\m,\x)||p_{\t}(\m,\x)) = \E_{p(\x)}[\mathcal{D}_{KL}(q_{\e}(\m|\x)||p_{\t}(\m|\x))]$
is the variational gap. For fixed $\t$ (and thus for fixed $A_{\t}$), any improvement in $\e$ that reduces the expected variational gap also tightens an upper bound on $\Delta \doteq A_{\t} - A_{\e}$. 

In fields where reliable data transmission or storage is crucial, introducing \acp{ecc} is a well-established approach to reduce the error rate when estimating a discrete source $\m$ transmitted through a noisy channel with output $\x$. Building on this, we propose using \acp{ecc} to protect $\m$ with controlled redundancy that the variational posterior $q_{\e}(\m|\x)$ can leverage. This way, it is possible to reduce error rates in inference and refine the variational parameters $\e$ on the enhanced model, thus narrowing the variational gap. Notably, according to \eqref{bound1}, a smaller variational gap results in a tighter upper bound on the error rates. In the following sections, we show that simple coding schemes enhance inference, improving generation, reconstruction, \acp{ll}, and uncertainty calibration in the latent space.

\section{Coded-DVAE}\label{coded_dvae}

\begin{wrapfigure}{r}{0.14\textwidth}
    \centering
    \includegraphics[width=0.1\textwidth]{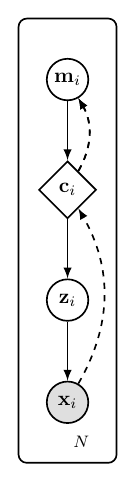}
    \caption{Graphical model of the Coded-DVAE.}
    \label{fig:DAG_bit}
\end{wrapfigure}

\begin{figure*}[!t]
\begin{subfigure}[c]{0.37\textwidth}
\centering
  \includegraphics[width=1\linewidth]{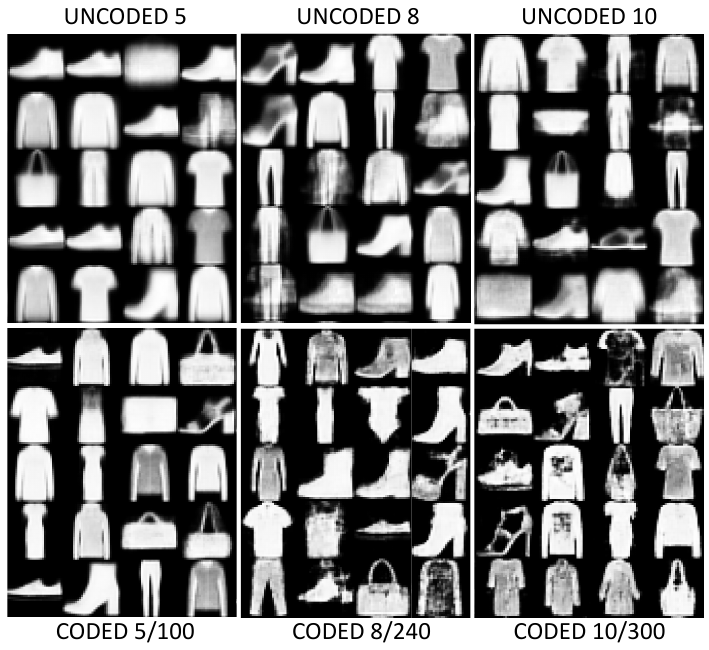}
\end{subfigure}
\begin{subfigure}[c]{0.63\textwidth}
    \centering
    \begin{tabular}{cccccc}
      \toprule 
      \bfseries Model & \bfseries BER &\bfseries BER MAP & \bfseries WER & \bfseries WER MAP & \bfseries LL test\\
      \midrule 
      uncoded 5 & 0.051 & 0.051 & 0.195 & 0.190 & -267.703\\
      coded 5/50 & 0.011 & 1.960$\cdot10^{-4}$ & 0.046 & 9.800$\cdot10^{-4}$ & -241.882\\
      coded 5/80 & 0.008 & 4.000$\cdot10^{-6}$ & 0.039 & 2.000$\cdot10^{-5}$ & -232.992\\
      coded 5/100 & 0.010 & 2.120$\cdot10^{-4}$ & 0.049 & 9.600$\cdot10^{-4}$ & -241.404\\
      \midrule
      uncoded 8 & 0.089 & 0.088 & 0.384 & 0.376 &  -249.880\\
      coded 8/80 & 0.021 & 0.003 & 0.144 & 0.021 & -232.992\\
      coded 8/160 & 0.027 & 0.001 & 0.189 & 0.009 & -235.819\\
      coded 8/240 & 0.037 & 0.004 & 0.231 & 0.023 & -238.459\\
      \midrule
      uncoded 10 & 0.142 & 0.144 & 0.622 & 0.644 & -244.997\\
      coded 10/100 & 0.040 & 0.005 & 0.321 & 0.036 & -230.772\\
      coded 10/200 & 0.044 & 0.006 & 0.341 & 0.039 & -234.849\\
      coded 10/300 & 0.045 & 0.002 & 0.349 & 0.014 & -238.647\\
      \bottomrule 
    \end{tabular}
\end{subfigure}
\caption{\textbf{Evaluation of generation in FMNIST.} Uncurated randomly generated samples (left) and evaluation metrics (right), including BER and WER sampled from the variational posterior, BER and WER using the MAP value, and test \ac{ll}.}\label{fig:generation_fmnist}
\end{figure*}

This section extends the previously described \ac{dvae}, introducing an \ac{ecc} over $\m$. We refer to this model as Coded-\ac{dvae}. The corresponding graphical model is shown in Fig. \ref{fig:DAG_bit}, where the diamond node represents a deterministic transformation. In \acp{ecc}, we augment the dimensionality of the binary latent space from $M$ to $D$, introducing redundancy in a controlled and deterministic manner, where $R=M/D$ is the \emph{coding rate}. An \ac{ecc} is designed to maximize the separation between the $2^M$ possible codewords within the $D$-bit binary space, enabling efficient error correction by searching for the nearest valid codeword (see Fig. \ref{fig:fancy_scheme}). A random selection of codewords results in \emph{random block codes} \citep{shannon}, which are very robust and amenable to theoretical analysis. However, their lack of structure makes them impractical, as encoding and decoding require exhaustive codeword enumeration. Appendix \ref{coded_word} provides a detailed formulation of the encoding and decoding process for a random code within the Coded-\ac{dvae}.

We adopt a much simpler linear coding scheme, namely repetition codes, where each bit of the original message $\m$ (i.e., each information bit) is duplicated multiple times to form the encoded message $\c$. Intuitively, the more times an information bit is repeated, the better it is protected. Our experiments consider uniform $(M, D)$ repetition codes where all bits are repeated $L$ times, resulting in codewords of dimension $D=ML$ and a coding rate of $R=1/L$. Note that repetition codes represent a special case of linear \acp{ecc} since each codeword can be computed by multiplying a binary vector $\m$ by an $M\times D$ \emph{generator matrix} $\mathbf{G}$, such that $\c = \m^T\mathbf{G}$, where $k$-th row, with $k=1,\ldots,M$, has entries equal to one at columns $L(k-1)+1,L(k-1)+2,\dots,Lk$, and zero elsewhere. For example, for $M=3$ and $L=2$, the generator matrix of the $(3,6)$ repetition code is
\begin{equation}
\mathbf{G} = \begin{bmatrix}
1 & 1 & 0 & 0 & 0 & 0 \\
0 & 0 & 1 & 1 & 0 & 0 \\
0 & 0 & 0 & 0 & 1 & 1 
\end{bmatrix}  .
\label{example}
\end{equation}

The generative process of the Coded-\ac{dvae} is similar to the uncoded version and is illustrated in Fig. \ref{fig:DAG_bit}. We assume the same prior distribution $p(\m)$, but in this case the samples $\m$ are deterministically encoded using $\mathbf{G}$. Consequently, the smoothing transformations are now defined over $\c$ 
\begin{align}
p(\z|\c)=\prod_{j=1}^{D}p(z_j|c_j),
\end{align}
with $p(z_j|c_j)$ following Eq. \eqref{smoothing_transformations_dvae}. The likelihood $p(\x|\z)$ is again of the form $p_{\t}(\x|\z) = p(f_{\t}(\z))$. Compared to the uncoded case, the input dimensionality to the decoder $f_{\t}(\z)$ is $L$ times larger due to the introduced redundancy. Therefore, the overall architecture of the decoder (detailed in Appendix \ref{architecture}) remains unchanged, except for the first \ac{mlp} layer that processes the input $\z$. When using a repetition code with rate $R=1/L$, the additional parameters of $f_{\t}(\z)$ amount to $(L-1)\times h$, where $h$ is the dimension of the first hidden layer. 

\subsection{Variational family and inference}

 Since $\c$ is a deterministic transformation of $\m$, its randomness is entirely determined by $\m$. Therefore, following the recognition model described in Fig. \ref{fig:DAG_bit}, we assume a variational family factorizing as 
\begin{equation}\label{VI1rep}
    q_{\e}(\m,\z|\x) = q_{\e}(\m|\x)p(\z|\c),
\end{equation}
where $q_{\e}(\m|\x)=\prod_{j=1}^{M}q_{\e}(m_j|\x)$ is computed in two steps. First, we construct an encoder $g_{\e}(\x)$ similar to the one in \eqref{VIuncoded}, which estimates $\c$ from $\x$ 
\begin{align}\label{equalizer}
q^{u}_{\e}(\c|\x) &= \prod_{j=1}^D q^{u}_{\e}(c_j|\x) = \prod_{j=1}^D \text{Ber}(c_j;g_{j,\e}(\x)),
\end{align}
\noindent
where the superscript $u$ indicates that this posterior does not utilize the \ac{ecc} (\textit{uncoded}).
Now, we leverage the known redundancy introduced by the \ac{ecc} to constrain the solution of $q_{\e}^u(\c|\x)$, given that each bit from $\m$ has been repeated $L$ times to create $\c$. To do so, we follow a \emph{soft decoding} approach, where the marginal posteriors of the information bits are derived from the marginal posteriors of the encoded bits, exploiting the code's known structure. In the case of repetition codes, we compute the all-are-zero and the all-are-ones products of probabilities of the bits in $\c$ that are copies of the same message bit and re-normalize as

\begin{equation}
\begin{gathered}
\label{products}
q(m_k=1|\x) = \frac{1}{Z} \prod_{j=L(k-1)+1}^{Lk}q^u_{\e}(c_j=1|\x), \\
q(m_k=0|\x)=  \frac{1}{Z} \prod_{j=L(k-1)+1}^{Lk}q^u_{\e}(c_j=0|\x),
\end{gathered}
\end{equation}
for $k=1,\ldots,M$ and $Z$ is the normalization constant. For the $M=3$, $L=2$ toy example in \eqref{example}
\begin{align*}
q_{\e}(m_1=1|\x)\propto q_{\e}^{u}(c_1=1|\x) q_{\e}^{u} (c_2=1|\x),\\
q_{\e}(m_2=1|\x) \propto q_{\e}^{u}(c_3=1|\x) q_{\e}^{u} (c_4=1|\x),\\
q_{\e}(m_3=1|\x) \propto q_{\e}^{u}(c_5=1|\x) q_{\e}^{u} (c_6=1|\x).
\end{align*}
This approach can be seen as a soft majority voting strategy. All operations in \eqref{products} are differentiable and implemented in the log domain. We consider the same architecture for the encoder $g_{\e}(\x)$ in \eqref{equalizer} as in the uncoded case, with the only difference being the final \ac{mlp} layer. The additional overhead in the coded case requires $(L-1)\times h'$ parameters in the last layer, where $h'$ is the output dimension. In Appendix \ref{ablation_parameters}, we conduct an ablation study on the number of trainable parameters to show that performance gains do not stem from the increased parameter count.

\begin{figure*}[h]
   \begin{subfigure}[c]{0.5\textwidth}
     \centering
     \includegraphics[width=0.77\linewidth]{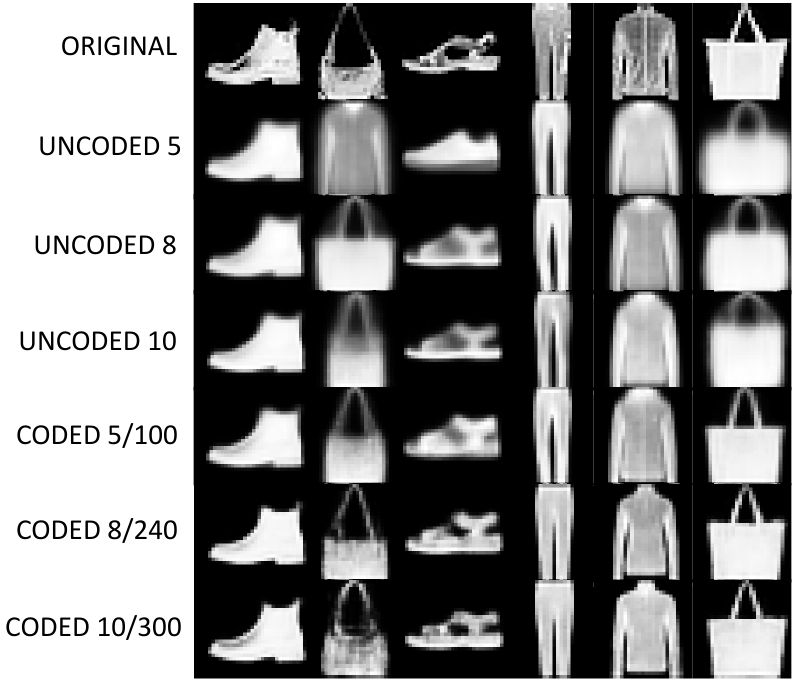}
   \end{subfigure}%
  \begin{subfigure}[c]{0.5\textwidth}
     \centering
        \begin{tabular}{ccccc}
          \toprule 
          \bfseries Model  & \bfseries PSNR  & \bfseries Acc  & \bfseries Conf. Acc & \bfseries Entropy\\
          \midrule 
          uncoded 5  & 14.477& 0.536 & 0.536 &  0.237\\ 
          coded 5/50 & 16.241  & 0.647 & 0.700 & 1.899\\
          coded 5/80 & 16.624 & 0.688 & 0.748 & 2.180\\
          coded 5/100  & 16.702 & 0.700 & 0.757 & 2.256\\
          \midrule
          uncoded 8  & 15.598 & 0.594 & 0.595 & 0.467\\
          coded 8/80  & 17.318 & 0.750 & 0.816 & 2.905 \\
          coded 8/160 & 17.713 & 0.783 & 0.831 & 3.637\\
          coded 8/240 & 17.861 & 0.799 & $\mathbf{0.893}$ & 4.000 \\
          \midrule
          uncoded 10 & 16.000 & 0.644 & 0.648 & 0.659 \\
          coded 10/100 & 17.694 & 0.790 & 0.850 & 3.879  \\
          coded 10/200 & 18.009 & 0.814 & 0.871  & 4.609\\
          coded 10/300 & $\mathbf{18.111}$ & $\mathbf{0.817}$ & 0.870 & 5.076\\
          \bottomrule 
     \end{tabular} 
   \end{subfigure}
   \caption{\textbf{Reconstruction performance in FMNIST}. The figure at the left shows an example of reconstructed test images obtained with different model configurations. Observe that more details are visualized as we increase the number of bits in the latent space and decrease the coding rate. The table at the right includes reconstruction metrics. Acc is the semantic accuracy, and Conf. Acc the confident semantic accuracy. Entropy is the average entropy of $q_{\e}(\m|\x)$ in the test set.}\label{fig:rec_fmnist}
 \end{figure*}

\subsection{Efficient reparameterization}

Given the variational family in \eqref{VI1rep}, the \ac{elbo} remains unchanged from \eqref{elbo_dvae}. However, the reparameterization trick in \eqref{CDF_dvae} assumes independent bits, which does not hold for $\c$. To address this issue efficiently during training, we adopt a \textit{soft encoding} approach. We compute a marginal probability for each bit in $\c$, leveraging the \ac{ecc} structure and the marginal posterior probabilities of the information bits in $\m$. For a repetition code, this involves simply replicating the posterior probabilities $q_{\e}(m_k|\x), k=1,\ldots,M$, for each copy of the same information bit\footnote{For the example in \eqref{example}, this results in $p_{\t}(\c|\x)\approx q_{\e}(\c|\x) = \text{Ber}(c_1;q_1)\text{Ber}(c_2;q_1)\text{Ber}(c_3;q_2)\text{Ber}(c_4;q_2)\text{Ber}(c_5;q_3)\text{Ber}(c_6;q_3)$.}.Hence, we treat the bits in $\c$ as independent but distributed according to $q_{\e}(\m|\x)$. The training algorithm can be found in  Appendix \ref{coded_train_algorithm}.

When marginalizing $\c$ using the \textit{soft encoding} marginals, we disregard potential correlations between the coded bits. Consequently, sampling from the marginals may produce an invalid codeword. However, since we do not sample the coded bits during training but instead propagate their marginal probabilities, we consider this method to have minimal negative impact. In fact, it can be seen as a form of probabilistic dropout, which enhances robustness during training. Since encoded words may contain inconsistent bits, the decoder learns to utilize the correlated inputs from repeated bits rather than disregarding them. Remark that when sampling from the generative model in test time, we use hard bits encoded into valid codewords, yielding visually appealing samples, indicating effective training.

\section{Results}\label{experiments}

This section empirically evaluates the Coded-\ac{dvae} with repetition codes and its uncoded counterpart on reconstruction and generation tasks. In particular, we present results for MNIST \citep{mnist}, FMNIST \citep{fmnist}, CIFAR10 \citep{cifar10}, and Tiny ImageNet \citep{tinyimagenet}. Additionally, we compare the coded model to the uncoded \ac{dvae} trained with the \ac{iwae} objective (see Fig. \ref{fig:fancy_figure} and Appendix \ref{iwae_results}). All experiments use a fixed architecture (see Appendix \ref{architecture}), modifying the encoder’s output and decoder’s input to accommodate the repetition code. We identify the models based on the number of information bits for uncoded models and the code rate for coded models. 

\subsection{Generation}\label{results_generation} 

We first evaluate the models using the image generation task. Fig. \ref{fig:generation_fmnist} presents examples of randomly generated images using various model configurations on FMNIST. Additional results for other datasets are provided in the Supplementary Material. All models produce more detailed and diverse images as the number of information bits increases. However, when the number of latent vectors becomes too high for the dataset's complexity, some codebook entries remain unspecialized during training. This results in generation artifacts, where images contain overlapping features from different object classes. A visual inspection of Fig. \ref{fig:generation_fmnist} suggests that such artifacts occur more frequently in the uncoded model.

\textbf{Error metrics}. The effectiveness of the \ac{ecc}-based inference can be assessed by generating images and measuring errors in recovering $\m$ by either drawing one sample from the variational posterior $q_{\e}(\m|\x)$ or using the \ac{map} estimate. As shown in the table included in Fig. \ref{fig:generation_fmnist}, the coded models significantly reduce both \ac{ber} and \ac{wer} compared to the uncoded case for the same number of latent bits. Note also that the error rates grow with the number of latent bits, which is expected due to the increased complexity of the inference process.

\textbf{Log-likelihood (LL)}. Aditionally, we estimated the data \ac{ll} using importance sampling with 300 samples per observation. As shown in the table included in Fig. \ref{fig:generation_fmnist}, coded models consistently outperform their uncoded counterparts on both train and test sets, aligning with the rest of the results. \ac{ll} values generally improve with more information bits, reflecting increased model flexibility. However, reducing the code rate does not enhance \ac{ll}, suggesting possible decoder overfitting, where reconstruction improves, but \ac{ll} declines. This may be due to the use of feed-forward networks at the decoder’s input, which might not fully capture the correlations in coded words (see Appendix \ref{augmented_model} for details).

\subsection{Reconstruction}\label{results_reconstruction}

We also assess the model's performance in image reconstruction. In the table in Fig. \ref{fig:rec_fmnist}, we report the \ac{psnr} of reconstructions on the FMNIST test set, calculated as
\begin{equation}
\text{PSNR} = 20\log\Big(\frac{\text{max}(x_{i,j})}{\text{RMSE} (\x, \x')}\Big),
\end{equation}
where $\text{max}(x_{i,j})$ is the highest pixel value, RMSE represents the Root Mean Square Error, and $\x'$ the reconstructed image. Results for the other datasets can be found in the Supplementary Material. In all cases, coded models outperform their uncoded counterparts in reconstruction quality, as indicated by higher \ac{psnr} values, a trend also visible in Fig. \ref{fig:rec_fmnist}. As the number of information bits increases, \ac{psnr} improves, reflecting greater model flexibility to capture data structure. We also observe improved \ac{psnr} with reduced code rates, as more redundancy is added. However, this increased redundancy does not enhance the model's flexibility, which is determined by the number of information bits.

\begin{figure}[b!]
  \centering
  \includegraphics[width=0.95\linewidth]{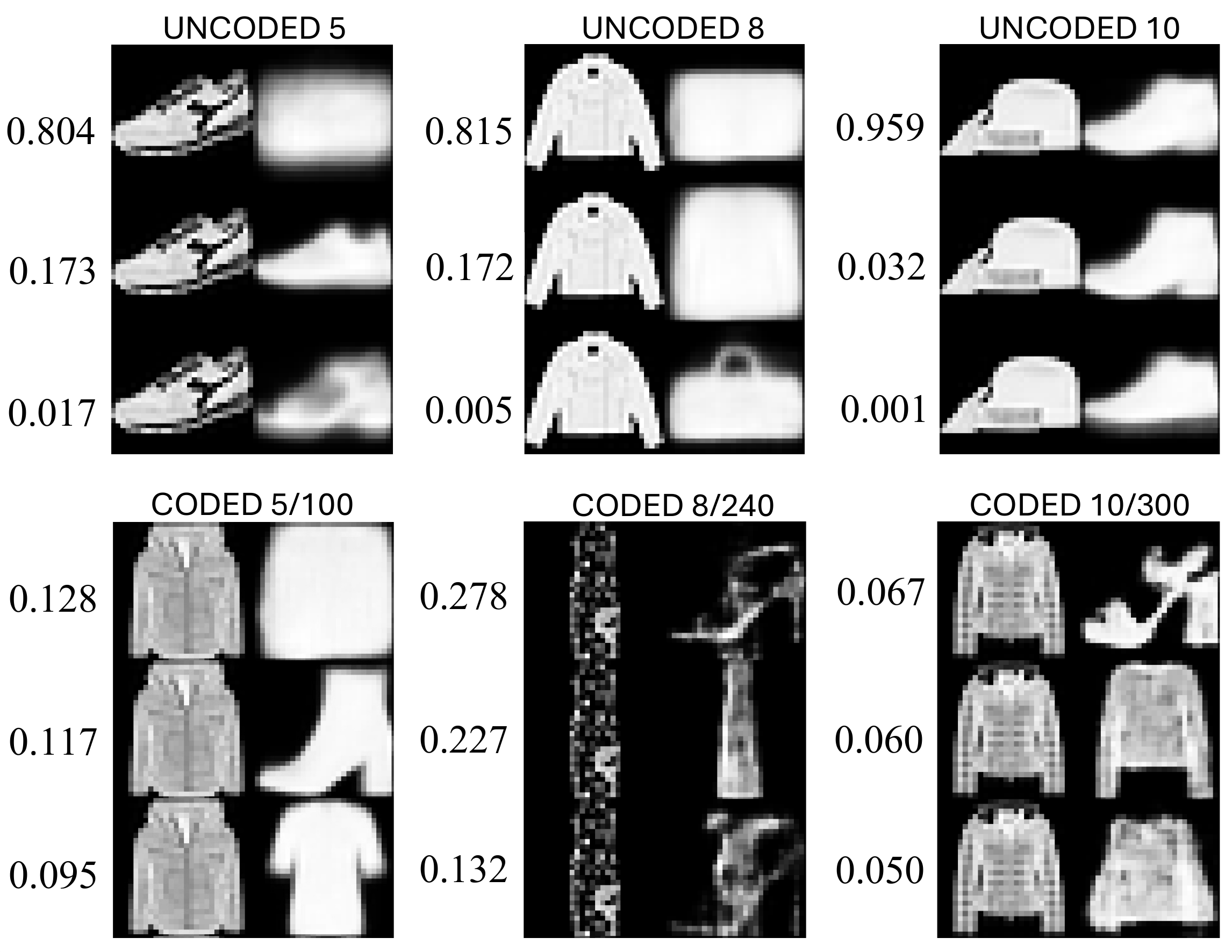}
  \caption{\textbf{Example of erroneous reconstructions in FMNIST}. The first column in each image shows the original images, while the second column displays the reconstructions. The $q_{\e}(\m|\x)$ probability is indicated in each row.}\label{fig:errors_fmnist}
\end{figure}

\textbf{Semantic Accuracy}. Since the \ac{psnr} operates at the pixel level, it does not account for the \textit{semantic} errors committed by the model. To address this, we evaluate reconstruction accuracy in Fig. \ref{fig:rec_fmnist}, checking whether the model reconstructs images within the correct semantic class (type of clothing in FMNIST). For this, we trained an image classifier for each dataset and compared the predicted labels of the reconstructed images with the ground truth labels. We also report a \textit{confident} reconstruction accuracy, which is calculated by considering only the images projected into a latent vector with posterior probability above 0.4 \footnote{We do not count errors when the \ac{map} of $q_{\e}(\m|\x) < 0.4$.}. Results indicate that incorporating an \ac{ecc} enables the model to produce latent spaces that better capture the semantics of the data.

\textbf{Posterior Uncertainty Calibration}. \hspace{0.2cm}Finally, we report the average entropy of the variational posterior $q_{\e}(\m|\x)$ over the test set in the table included in Fig.~\ref{fig:rec_fmnist}. The low entropy in the uncoded models indicates low uncertainty when the model maps data points into the latent space, which could be beneficial if the model consistently assigns high probability to the correct latent vectors. However, the semantic accuracy results show that this is not the case for the uncoded model, as it often projects images into vectors corresponding to the wrong semantic class with high confidence.

\begin{figure}[!t]
    \centering
     \begin{subfigure}[b]{0.48\textwidth}
         \centering
         \includegraphics[width=1\linewidth]{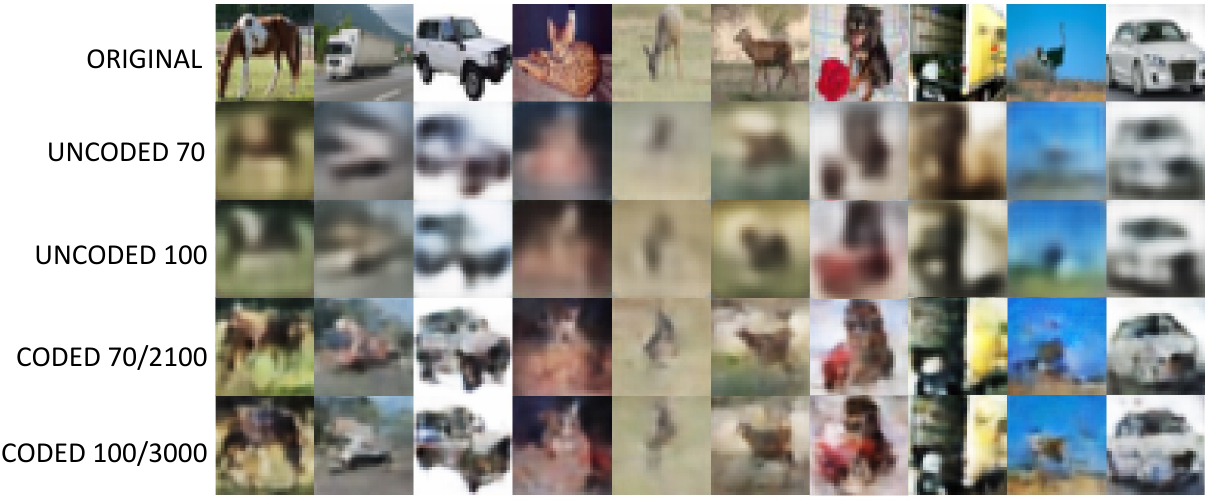}
    \end{subfigure}
    \centering
     \begin{subfigure}[b]{0.48\textwidth}
         \centering
         \includegraphics[width=1\linewidth]{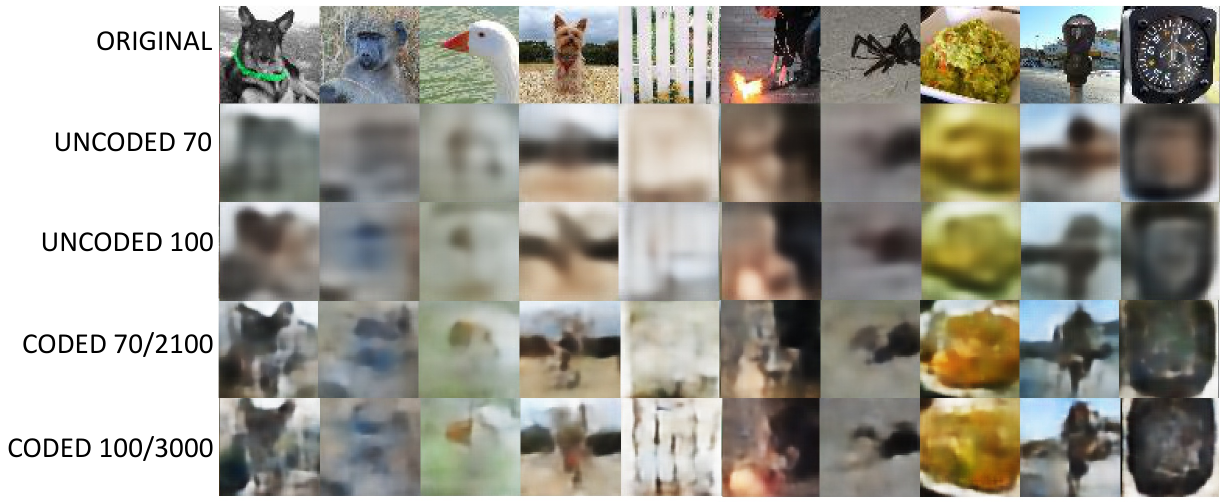}
    \end{subfigure}
    \caption{\textbf{Reconstruction results} for CIFAR10 (top) and Tiny ImageNet (bottom).} \label{fig:cifar_imagenet_rec}
\end{figure}

\begin{figure*}[ht]
    \centering
     \begin{subfigure}[b]{0.46\textwidth}
         \centering
         \includegraphics[width=1\linewidth]{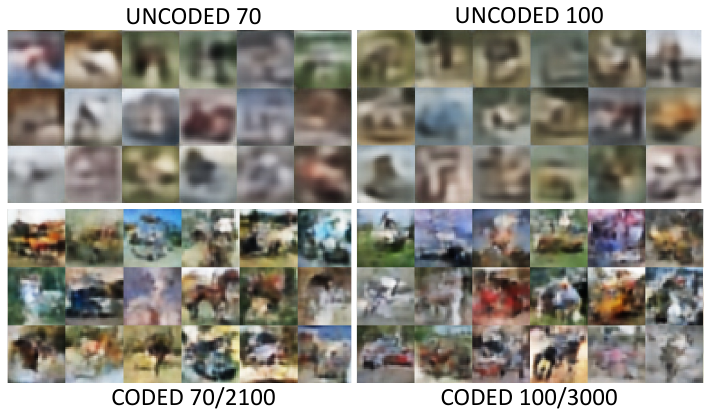}
    \end{subfigure}
    \hspace{1cm}
    \centering
     \begin{subfigure}[b]{0.46\textwidth}
         \centering
         \includegraphics[width=1\linewidth]{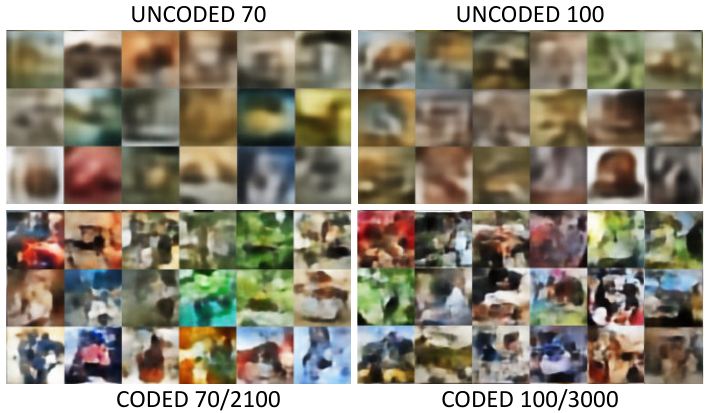}
    \end{subfigure}
    \caption{\textbf{Generation results} for CIFAR10 (left) and Tiny ImageNet (right).}
    \label{fig:cifar_imagenet_gen}
\end{figure*}

Coded models, in contrast, improve semantic accuracy and exhibit higher entropy. This indicates that (i) the Coded-DVAE recognizes multiple latent vectors that might be related to a given semantic class, and (ii) the model’s posterior demonstrates significant uncertainty (high entropy) for images where the class is not well identified. This is illustrated in Fig. \ref{fig:errors_fmnist}, where we show images with class reconstruction errors caused by the MAP latent word from $q_{\e}(\m|\x)$. The reconstructions of the three most probable latent vectors, along with their respective probabilities, are displayed. Notably, the uncoded model shows high confidence regardless of the reconstruction result, while the coded posterior shows greater uncertainty. Note that this does not imply an uninformative distribution, as information bits remain recoverable. Additionally, increasing the number of latent bits does not lead to an excessive increase in entropy, despite the exponential growth in the number of vectors. 

\subsection{Results on CIFAR10 and ImageNet}\label{results_bound}

Since MNIST-like datasets are relatively simple, it is challenging to fully assess the benefits of introducing \acp{ecc} in our model. To address this, we present additional results on more complex datasets, CIFAR10 and Tiny ImageNet, which feature colored images with more intricate shapes, patterns, and greater diversity. We trained both uncoded and coded models with different configurations to better understand the impact of the \ac{ecc}. For context, the DVAE++ model \citep{dvae++} required 128 binary latent variables to achieve state-of-the-art performance on these datasets, using a more complex model with Boltzmann Machine priors. In Fig. \ref{fig:cifar_imagenet_rec}, we show reconstruction examples, and in Fig. \ref{fig:cifar_imagenet_gen}, we present randomly generated images. Additional results are available in Appendices \ref{cifar_results_supp} and \ref{tinyimagenet_results_supp}. The results align with those from previous sections, but the performance difference is even more pronounced in this case. We find that the uncoded \ac{dvae} struggles to decouple spatial information in the images, while the Coded-\ac{dvae} shows particular promise for learning low-dimensional discrete latent representations, even for complex datasets. It is important to note that we used a simple architecture to isolate the gains achieved purely through the introduction of \acp{ecc} in the latent space, keeping a simple independent prior.

\section{Beyond repetition codes}\label{beyond_repetition}

We have presented compelling proof-of-concept results that incorporating \acp{ecc}, like repetition codes, into \acp{dvae} can improve performance. We believe this opens a new path for designing latent probabilistic models with discrete latent variables. Although a detailed analysis of the joint design of \ac{ecc} and encoder-decoder networks is beyond the scope of this work, in Appendix \ref{properties_eccs} we outline key properties that any \acp{ecc} must satisfy to be integrated within this framework.

\begin{figure}[t!]
  \centering
  \includegraphics[width=1\linewidth]{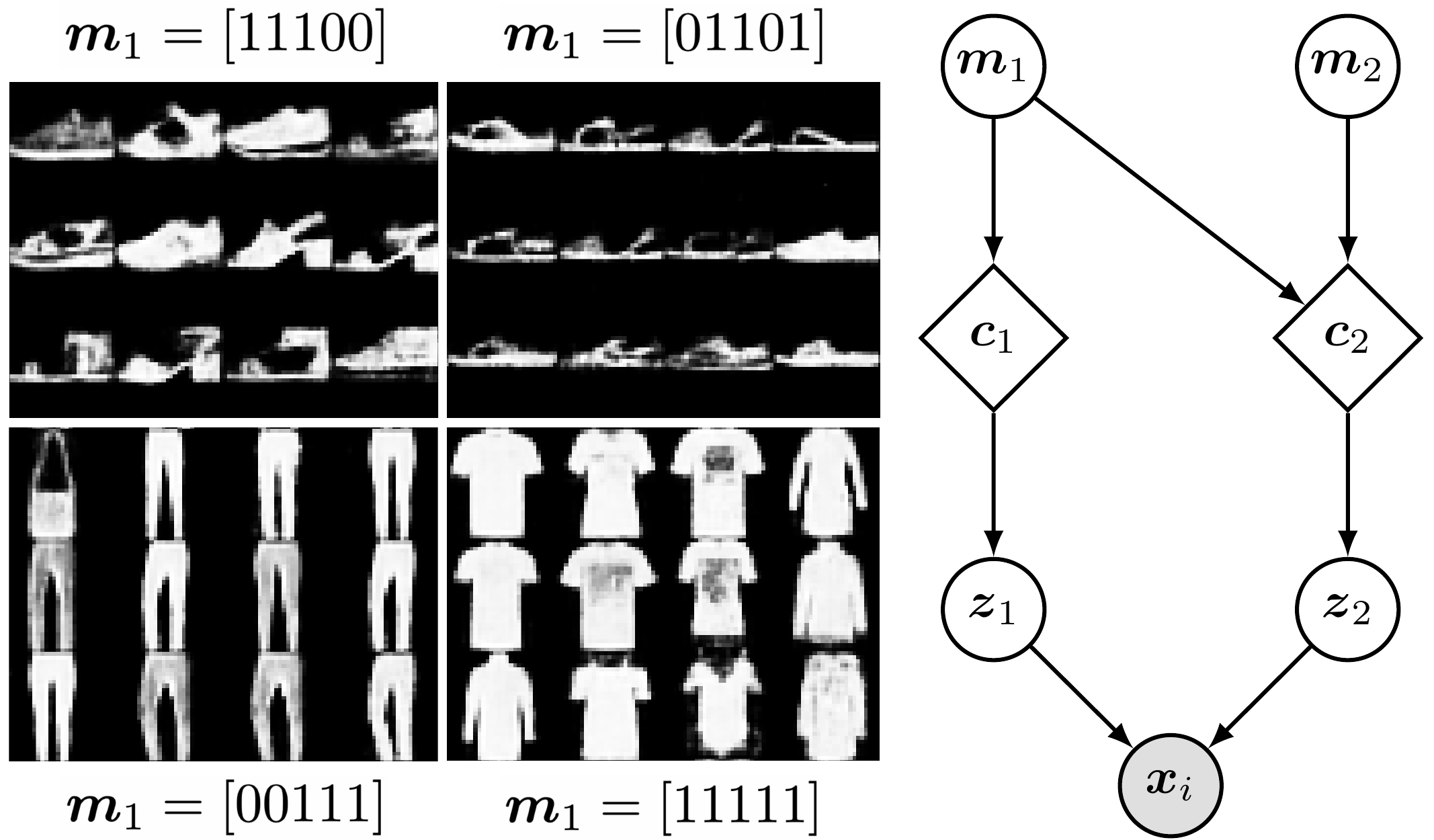}
  \caption{\textbf{Hierarchical Coded-DVAE}. Graphical model (right) and generated images with a 5/100 code per branch (left). Here, $\m_1$ is fixed, and $\m_2$ is randomly sampled.}\label{fig:hierarchical_main}
\end{figure}

Drawing inspiration from polar codes \citep{polarcodes}, we introduce a hierarchical Coded-\ac{dvae} with two layers. The graphical model is shown in Fig. \ref{fig:hierarchical_main}. In this model, the latent bits $\m_1$ are encoded using a repetition code in the first layer, producing $\c_1$ and $\z_1$. Simultaneously, the vector $\m_2$ is linearly combined with $\m_1$ using modulo 2 operations $\big(\m_1 \oplus \m_2\big)$ and then encoded using a repetition code, yielding $\c_2$ and $\z_2$. Both soft vectors are concatenated and fed to the decoder to generate $\x$. Since $\m_1$ appears in both generative branches, it receives stronger protection. Inference follows a similar approach to the Coded-\ac{dvae}, incorporating the linear combination of $\m_1$ and $\m_2$ used in the second branch. This hierarchical structure allows the model to separate high-level information from finer details, as we show in the results presented in Appendix \ref{polar_codes_results}. The bits in $\m_1$ offer stronger protection, allowing the model to encode the semantic information (type of clothing in FMNIST). As a result, during inference, it is more likely to project the image into a latent vector $\m_1$ that represents the correct semantic class. In contrast, $\m_2$ provides weaker protection and encodes image details that are less crucial for reconstruction. Fig. \ref{fig:hierarchical_main} demonstrates that by keeping $\m_1$ fixed (semantic information) and sampling $\m_2$ randomly (fine details) we can generate diverse images within the same semantic class.


\section{Conclusion}

This paper presents the first proof-of-concept demonstration that safeguarding latent information with \acp{ecc} within deep generative models holds promise for enhancing overall performance. By integrating redundancy into the latent space, the variational family can effectively refine the inference network's output according to the structure of the \ac{ecc}. Our findings underscore the efficacy of simple and efficient \acp{ecc}, like repetition codes, showcasing remarkable improvements over a lightweight version of the \ac{dvae} introduced in \cite{dvae++}. 

Furthermore, our work reveals numerous avenues for future research. Firstly, investigating architectures capable of efficiently utilizing the correlations and structure introduced by the \acp{ecc}, in contrast to the feed-forward networks employed in this study.  We also contemplate exploring more complex and robust coding schemes, conducting theoretical analyses aligned with Shannon's channel capacity and mutual information to determine the fundamental parameters of the \ac{ecc} needed to achieve reliable inference, exploring different modulations, and integrating these concepts into state-of-the-art models based on discrete representations.


\begin{acknowledgements} 
    
    María Martínez-García was supported by the Generaci\'on de Conocimiento grant PID2022-142506NA-I00. Grace Villacrés was supported by the Comunidad de Madrid within the 2023-2026 agreement with Universidad Rey Juan Carlos for the granting of direct subsidies for the promotion, encouragement of research and technology transfer, line of Action A Emerging Doctors, under Project OrdeNGN (Ref. F1177). David Mitchell was supported by the National Science Foundation under Grant No. CCF-2145917 and he acknowledges travel support from the European Union’s Horizon 2020 research and innovation program under Grant Agreement No. 951847. Pablo M. Olmos was supported by the Comunidad de Madrid IND2022/TIC-23550, IDEA-CM project (TEC-2024/COM-89), the ELLIS Unit Madrid (European Laboratory for Learning and Intelligent Systems), the 2024 Leonardo Grant for Scientific Research and Cultural Creation from the BBVA Foundation, and by projects MICIU/AEI/10.13039/501100011033/FEDER and UE (PID2021-123182OB-I00; EPiCENTER). 

    We would like to thank Isabel Valera for her valuable feedback on earlier drafts of the manuscript, as well as for the discussions that helped us improve the presentation of our method to the machine learning community.
    
\end{acknowledgements}

\bibliography{uai2025-template/references}

\newpage
\onecolumn
\title{Improved Variational Inference in Discrete VAEs using Error Correcting Codes\\(Supplementary Material)}
\maketitle
\appendix
The following Appendices offer further details on the model architecture, implementation, and experimental setup. They also include additional results on the FMNIST \citep{fmnist}, MNIST \citep{mnist}, CIFAR10 \citep{cifar10}, and Tiny ImageNet \citep{tinyimagenet} datasets, along with comparisons to uncoded models trained with the IWAE objective \citep{iwae}, and a description of the hierarchical Coded-\ac{dvae}. Given the length of the material, we have included a Table of Contents for easier navigation.

\startcontents[sections]
\printcontents[sections]{l}{1}{\setcounter{tocdepth}{2}}

\newpage

\section{Uncoded training algorithm}\label{uncoded_train_algorithm}

The following pseudo-code describes the training process for the uncoded \ac{dvae}. It's important to note that the main difference from the training of the Coded-\ac{dvae} lies in the fact that the encoder directly outputs $q_{\e}^{u}(\m|\x_i)$, which is used to sample $\z$. Therefore, we skip the soft decoding and coding steps.

\begin{algorithm}[!h]
   \caption{Training the model with \textit{uncoded} inference.}
   \label{training_uncoded}
\begin{algorithmic}[1]
   \STATE {\bfseries Input:} training data $\x_i$.
   \REPEAT
   \STATE $q_{\e}^{u}(\m|\x_i) \gets$ forward encoder $g_{\e}(\x_i)$ 
   \STATE $\z \gets$ sample from \eqref{CDF_dvae}
   \STATE $p_{\t}(\x|\z) \gets$ forward decoder $f_{\t}(\z)$
   \STATE Compute ELBO according to \eqref{elbo_dvae}
   \STATE $\t, \e \gets Update(ELBO)$ 
   \UNTIL convergence
\end{algorithmic}
\end{algorithm}

\section{Coded training algorithm}\label{coded_train_algorithm}

The following pseudo-code describes the training process for the Coded-\ac{dvae}. Here, we utilize soft decoding to leverage the added redundancy and retrieve the marginal posteriors of the information bits $\m$, correcting potential errors in $q_{\e}^{u}(\c|\x_i)$. We then apply the soft encoding technique to incorporate the structure of the code and sample $\z$ using the reparameterization trick as described in \eqref{CDF_dvae}.

\begin{algorithm}
   \caption{Training the Coded-\ac{dvae} with repetition codes.}
    \label{algorithm_codedDVAE}
\begin{algorithmic}[1]
   \STATE {\bfseries Input:} training data $\x_i$, matrix $\mathbf{G}$.
   \REPEAT
   \STATE $q_{\e}^{u}(\c|\x_i) \gets$ forward encoder $g_{\e}(\x_i)$ 
   \STATE $q_{\e}(\m|\x_i) \gets$ soft decoding by aggregating $q_{\e}^{u}(\c|\x_i)$ according to \eqref{products}
   \STATE $q_{\e}(\c|\x_i) \gets$ repeat posterior bit probabilities  $q_{\e}(\m|\x_i)$ according to $\mathbf{G}$
   \STATE $\z \gets$ sample from \eqref{CDF_dvae}
   \STATE $p_{\t}(\x|\z) \gets$ forward decoder $f_{\t}(\z)$
   \STATE Compute ELBO according to \eqref{elbo_dvae}
   \STATE $\t, \e \gets Update(ELBO)$ 
   \UNTIL convergence
\end{algorithmic}
\end{algorithm}

\section{Architecture}\label{architecture}

In this section, we detail the architecture used to obtain the experimental results with FMNIST and MNIST (28x28 gray-scale images). Note that, across the experiments, we only modify the output layer of the encoder and the input layer of the decoder to adapt to the different configurations of the model. This modification leads to a minimal alteration in the total number of parameters. In Section \ref{ablation_parameters}, we conduct an ablation study on the number of trainable parameters to show that the enhancement in performance is not attributed to the increased dimensionality introduced by redundancy.

For the additional CIFAR10 experiments, we change the input of the encoder and the output of the decoder to process the 32x32 color images. For the Tiny ImageNet experiments, we do the same to process 64x64 color images. The rest of the architecture remains unchanged.

These architectures are comprehensively described in the following subsections.

\subsection{Encoder}

\begin{figure}[!htb]
  \centering
  \includegraphics[width=1\linewidth]{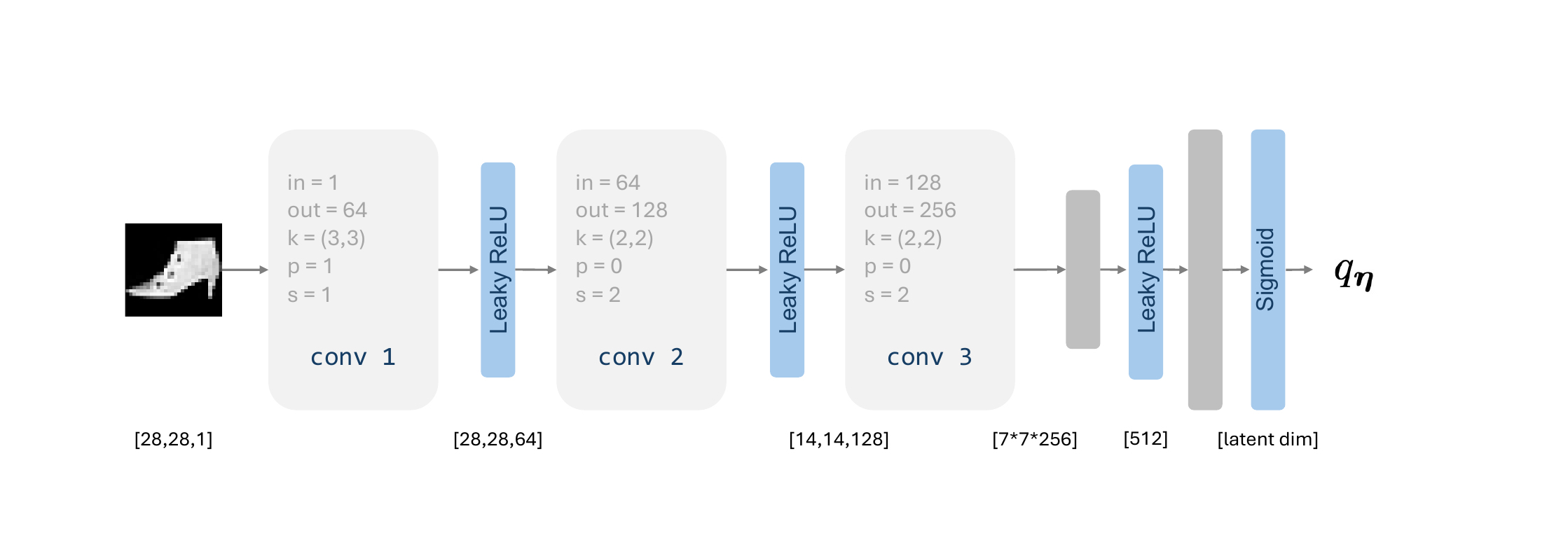}
  \caption{Block diagram of the \textbf{encoder architecture} for FMNIST and MNIST.}\label{fig:architecture_encoder}
\end{figure}

The encoder \ac{nn} consists of 3 convolutional layers followed by two fully connected layers. We employed Leaky ReLU as the intermediate activation function and a Sigmoid as the output activation, since the encoder outputs bit probabilities. The full architecture is detailed in Fig. \ref{fig:architecture_encoder}.

\subsection{Decoder}

\begin{figure}[!htb]
  \centering
  \includegraphics[width=1\linewidth]{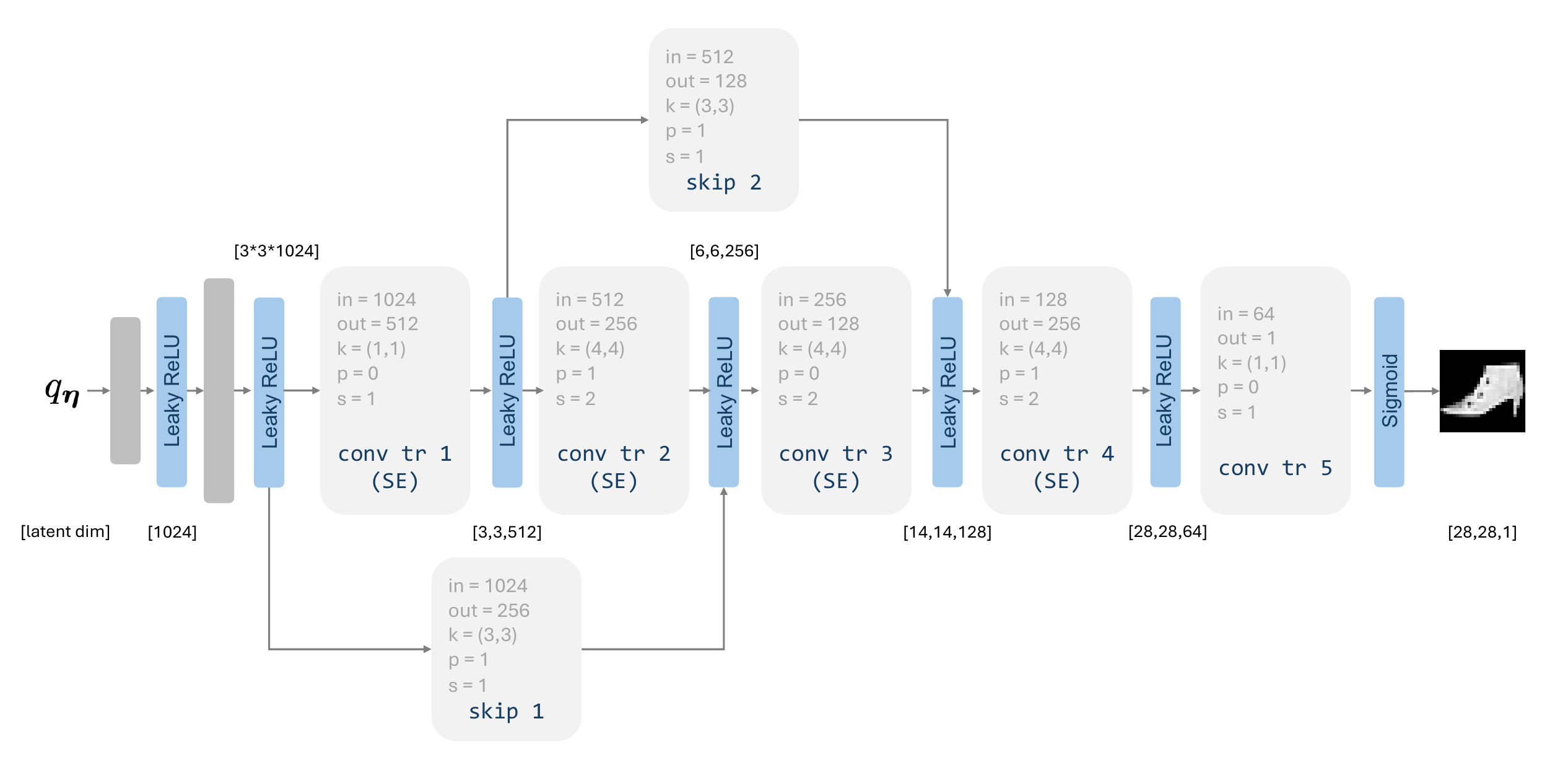}
  \caption{Block diagram of the \textbf{decoder architecture} for FMNIST and MNIST.}\label{fig:architecture_decoder}
\end{figure}

The decoder architecture is inspired by the one proposed in \citet{dgd}. It is composed of two fully connected layers, followed by transposed convolutional layers with residual connections and \ac{se} layers \citep{hu2018squeeze}. We employed Leaky ReLU as the intermediate activation function and a Sigmoid as output activation, given that we consider datasets with gray-scale images. The complete architecture is detailed in Fig. \ref{fig:architecture_decoder}.

\section{FMNIST results}\label{supplementary_fmnist}

In this section, we present supplementary results obtained with the FMNIST dataset.

\subsection{Training}

\begin{figure*}[!hb]
  \centering
  \includegraphics[width=1\linewidth]{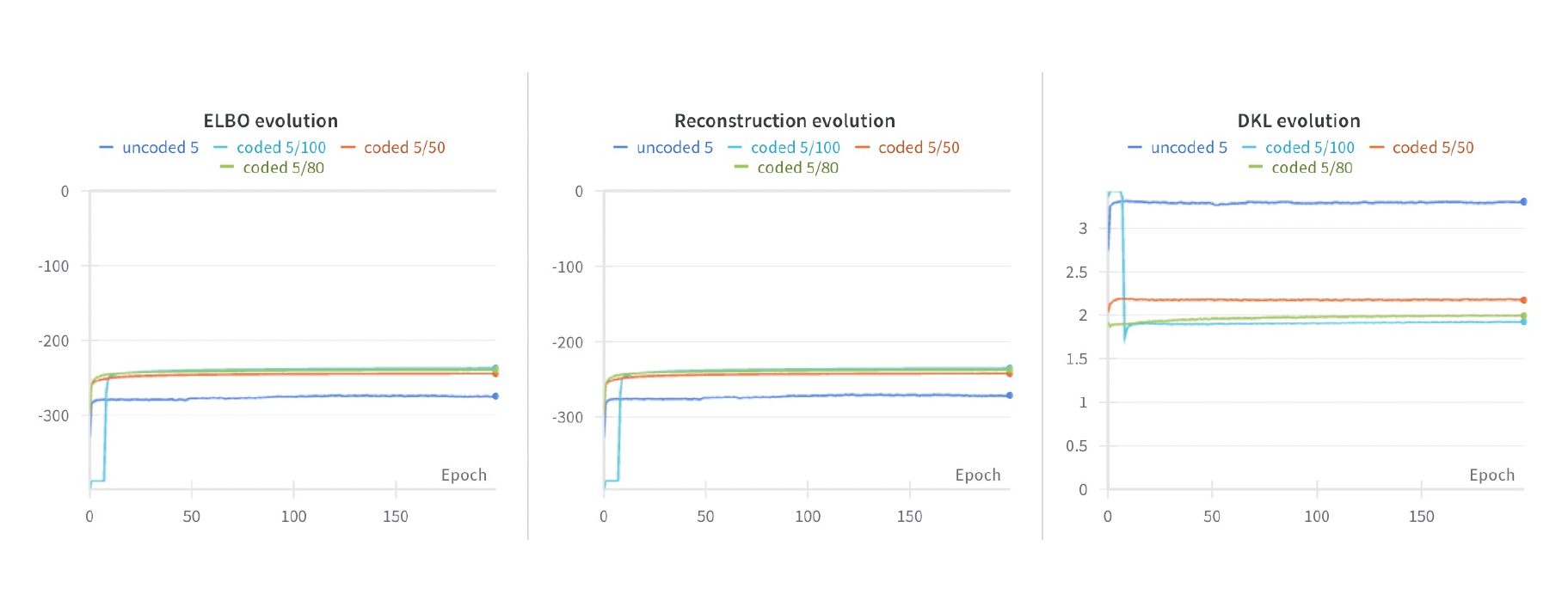}
  \caption{Evolution of the \ac{elbo} during training with 5 information bits on FMNIST.}\label{fig:elbo_fmnist_5}
\end{figure*}

\begin{figure*}[!hb]
  \centering
  \includegraphics[width=1\linewidth]{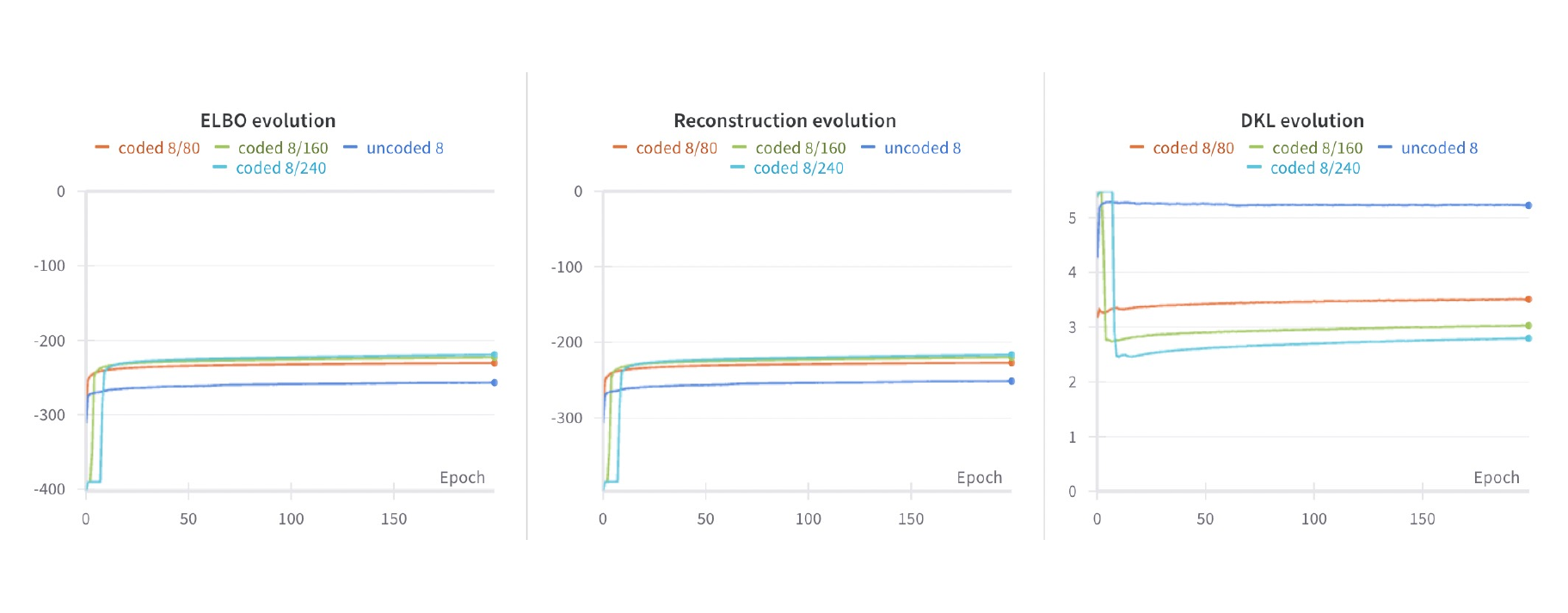}
  \caption{Evolution of the \ac{elbo} during training with 8 information bits on FMNIST.}\label{fig:elbo_fmnist_8}
\end{figure*}

\begin{figure*}[!h]
  \centering
  \includegraphics[width=1\linewidth]{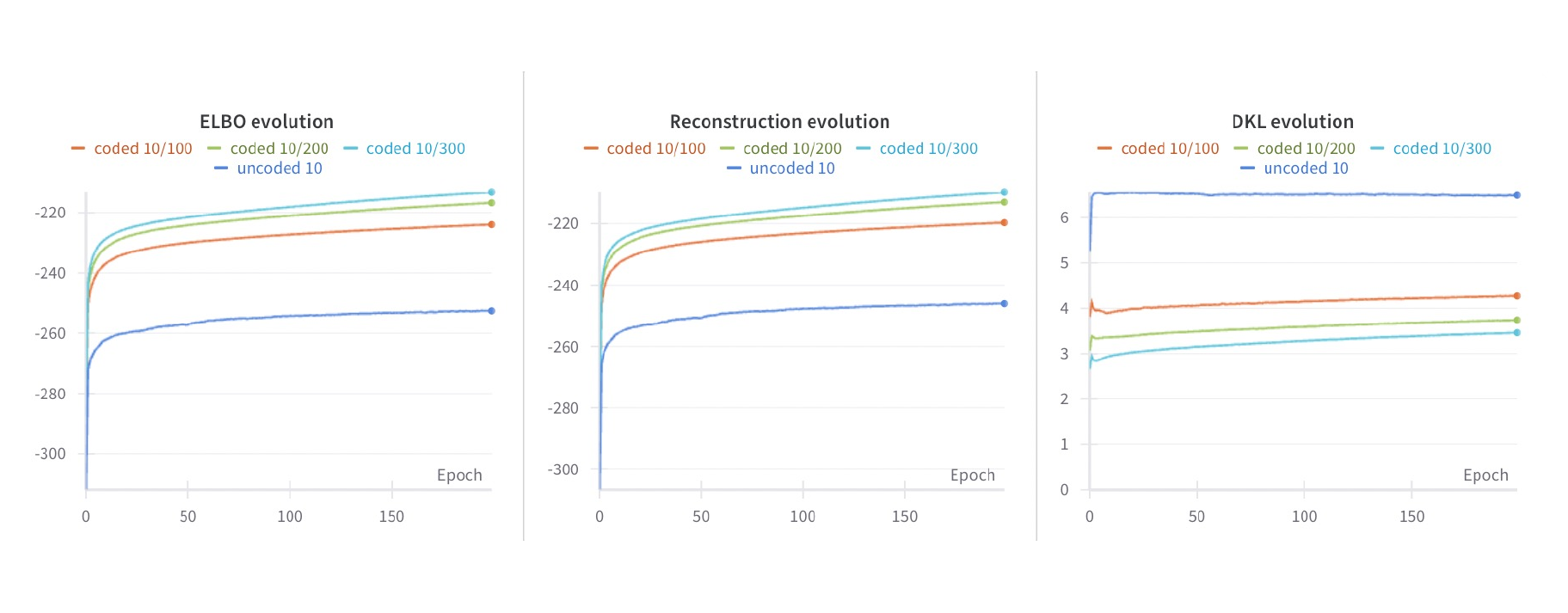}
  \caption{Evolution of the \ac{elbo} during training with 10 information bits on FMNIST.}\label{fig:elbo_fmnist_10}
\end{figure*}

We present the evolution of the \ac{elbo} and its terms throughout the training process. The models were trained for 200 epochs using an Adam optimizer with a learning rate of $10^{-4}$, and a batch size of 128. Fig. \ref{fig:elbo_fmnist_5} displays the results for configurations with 5 information bits, Fig. \ref{fig:elbo_fmnist_8} for 8 information bits, and Fig. \ref{fig:elbo_fmnist_10} for 10 information bits. The colors in all plots represent the various code rates. 

Across all cases, coded models achieve superior bounds. The main differences in the \ac{elbo} come from the different performances in reconstruction. As we have observed in the different experiments, coded models are capable of generating more detailed images and accurate reconstructions. Introducing the repetition code also leads to smaller \ac{kl} values, indicating that the posterior latent features are potentially disentangled and less correlated. 

We observe that, as we decrease the code rate, we obtain better bounds in general. Adding redundancy does not increase the model’s flexibility, since the information bits determine the number of latent vectors. However, the introduction of \acp{ecc} in the model allows for latent spaces that better capture the structure of the images while employing the same number of latent vectors.


\subsection{Reconstruction and generation}

In this section, we augment the results presented in the main text, including outcomes obtained with the training dataset in Table \ref{tab:metrics_rec_fmnist_train}. We include again the results obtained in the test to facilitate comparison. Additionally, we assess reconstruction quality using the \ac{ssim} to provide a more comprehensive evaluation. The results remain consistent across the two data partitions, and the analysis conducted for the test set also applies to training data.

In all the cases, the coded models yield higher \ac{psnr} and \ac{ssim} values than their uncoded counterparts, indicating a superior performance in reconstruction. We observe a general improvement in \ac{psnr} and \ac{ssim} as we increase the number of information bits (i.e., as we augment the latent dimensionality of the model) and decrease the code rate (i.e., as we introduce more redundancy).

As we discussed in the main text, the reconstruction metrics such as the \ac{psnr}  and \ac{ssim} do not account for the semantic errors committed by the model. Therefore, we additionally report the semantic accuracy and the \textit{confident} semantic accuracy. While the reconstruction accuracy is computed across the entire dataset partitions, for the confident accuracy, we only consider those images projected into a latent vector with a probability exceeding $0.4$. We observe that coded models better capture the semantics of the images while employing the same number of latent vectors, significantly outperforming the uncoded models in terms of accuracy in all the cases.

\begin{figure*}[!h]
  \centering
  \includegraphics[width=1\linewidth]{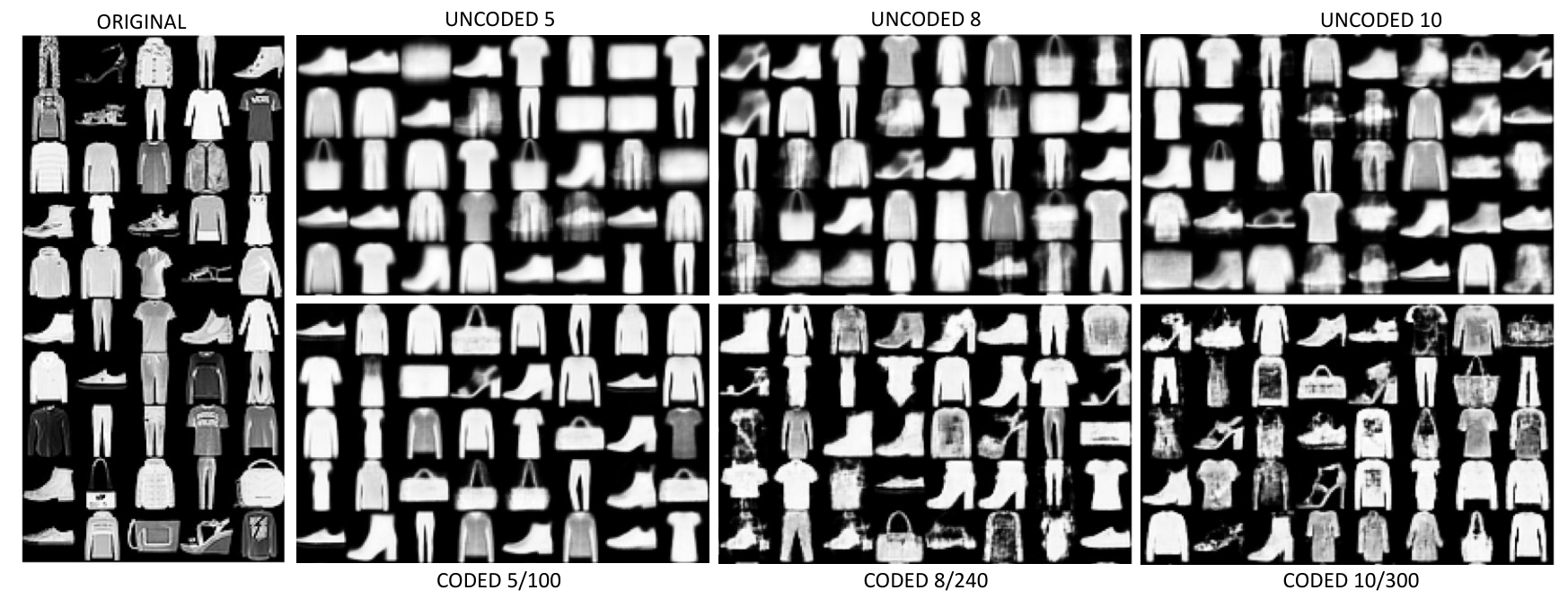}
  \caption{Example of randomly generated, uncurated images using different model configurations.}\label{fig:generation_fmnist_supp}
\end{figure*}

\newpage
In Fig. \ref{fig:generation_fmnist_supp}, we include additional examples of randomly generated images using different model configurations. We observe that coded models can generate more detailed and diverse images than their uncoded counterparts.

\begin{table*}[t]

\centering
    \caption{Evaluation of reconstruction performance in FMNIST.} \label{tab:metrics_rec_fmnist_train}
    \begin{tabular}{ccccccccc}
      \toprule 
      \bfseries Model & \begin{tabular}{@{}c@{}}\bfseries{PSNR}\\ \bfseries{(train)}\end{tabular}&  \begin{tabular}{@{}c@{}}\bfseries{SSIM}\\ \bfseries{(train)}\end{tabular}&\begin{tabular}{@{}c@{}}\bfseries{Acc}\\ \bfseries{(train)}\end{tabular} & \begin{tabular}{@{}c@{}}\bfseries{Conf. Acc.}\\ \bfseries{(train)}\end{tabular} & \begin{tabular}{@{}c@{}}\bfseries{PSNR}\\ \bfseries{(test)}\end{tabular} & \begin{tabular}{@{}c@{}}\bfseries{SSIM}\\ \bfseries{(test)}\end{tabular} &  \begin{tabular}{@{}c@{}}\bfseries{Acc}\\ \bfseries{(test)}\end{tabular}  & \begin{tabular}{@{}c@{}}\bfseries{Conf. Acc.}\\ \bfseries{(test)}\end{tabular} \\
      \midrule 
      uncoded 5  & 14.490 & 0.438 & 0.541 & 0.541 & 14.477 & 0.437 & 0.536 & 0.536 \\ 
      coded 5/50 & 16.375 & 0.559 & 0.656  & 0.702 & 16.241 & 0.551 & 0.647 & 0.700\\
      coded 5/80 & 16.824 & 0.585 & 0.694 & 0.751 & 16.624 & 0.574 & 0.688 & 0.748 \\
      coded 5/100 & 17.001 & 0.596 & 0.708 & 0.760 & 16.702 & 0.580 & 0.700 & 0.757\\
      \midrule
      uncoded 8  & 15.644 & 0.513 & 0.601  & 0.602 & 15.598 & 0.509 & 0.594 & 0.595\\
      coded 8/80 & 17.877 & 0.647 & 0.769  & 0.842 & 17.318 & 0.619 & 0.750 & 0.816\\
      coded 8/160 & 18.828 & 0.690 & 0.807 & 0.878 & 17.713 & 0.641 & 0.783 & 0.831\\
      coded 8/240 & 19.345 & 0.713 & 0.831  & 0.921 & 17.861 & 0.653 & 0.799 & $\mathbf{0.893}$ \\
      \midrule
      uncoded 10 & 16.053 & 0.542 & 0.650  & 0.652 & 16.000 & 0.536 & 0.644 & 0.648 \\
      coded 10/100 & 18.827 & 0.690 & 0.813 & 0.885 & 17.694 & 0.639 & 0.790 & 0.850 \\
      coded 10/200 & 19.937 & 0.735 & 0.846 & 0.897 & 18.009 & 0.659 & 0.814 & 0.871 \\
      coded 10/300 & $\mathbf{20.529}$ & $\mathbf{0.754}$ & $\mathbf{0.855}$ & $\mathbf{0.907}$ & $\mathbf{18.111}$ & $\mathbf{0.662}$ & $\mathbf{0.817}$ & 0.870 \\
      \bottomrule 
    \end{tabular}

\end{table*}

\section{MNIST results}\label{mnist_results}

In this section, we report the results obtained with the MNIST dataset.

\subsection{Training}

\begin{figure*}[!b]
  \centering
  \includegraphics[width=1\linewidth]{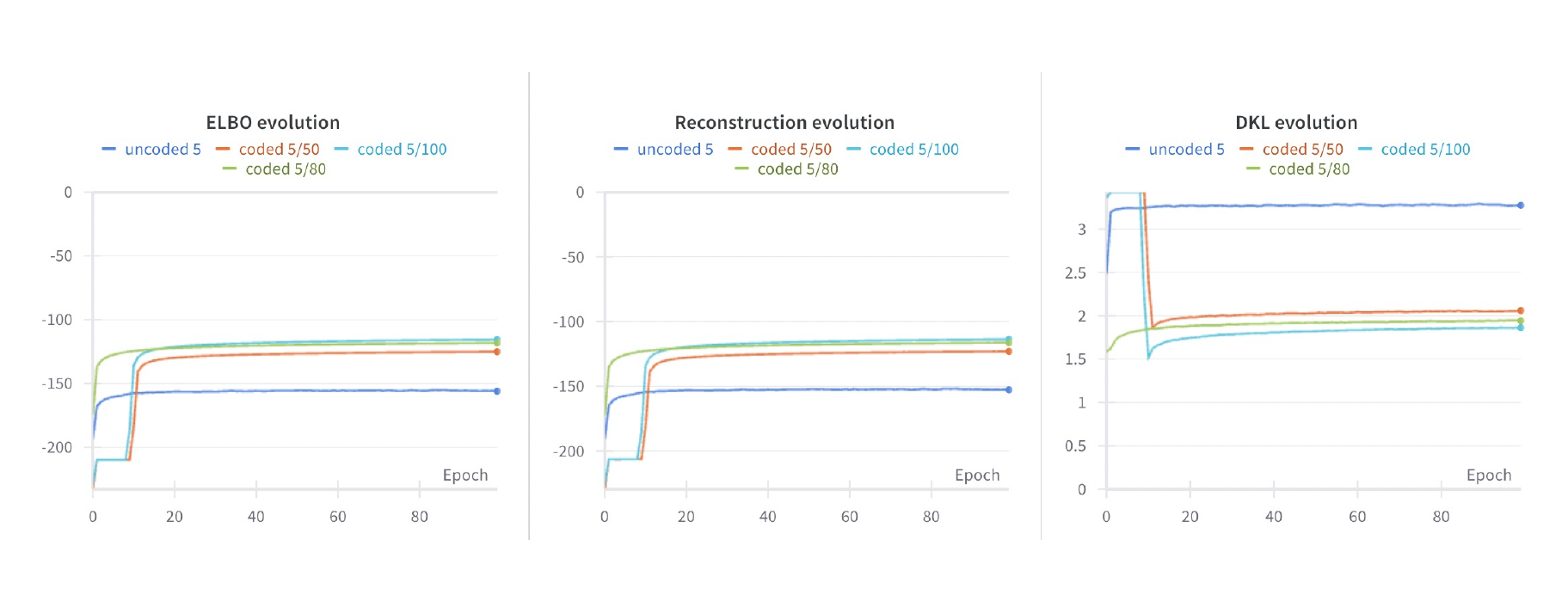}
  \caption{Evolution of the \ac{elbo} during training with 5 information bits on MNIST.}\label{fig:elbo_mnist_5}
\end{figure*}

\begin{figure*}[!ht]
  \centering
  \includegraphics[width=1\linewidth]{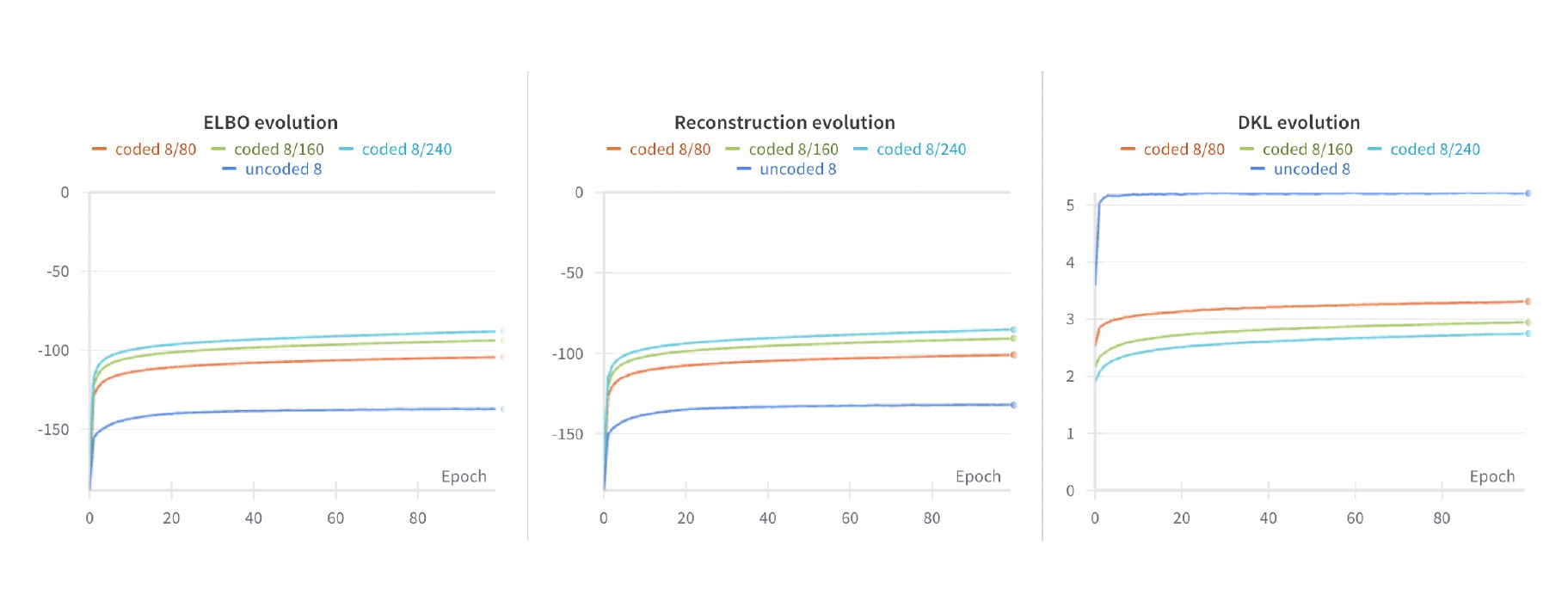}
  \caption{Evolution of the \ac{elbo} during training with 8 information bits on MNIST.}\label{fig:elbo_mnist_8}
\end{figure*}

\begin{figure*}[!ht]
  \centering
  \includegraphics[width=1\linewidth]{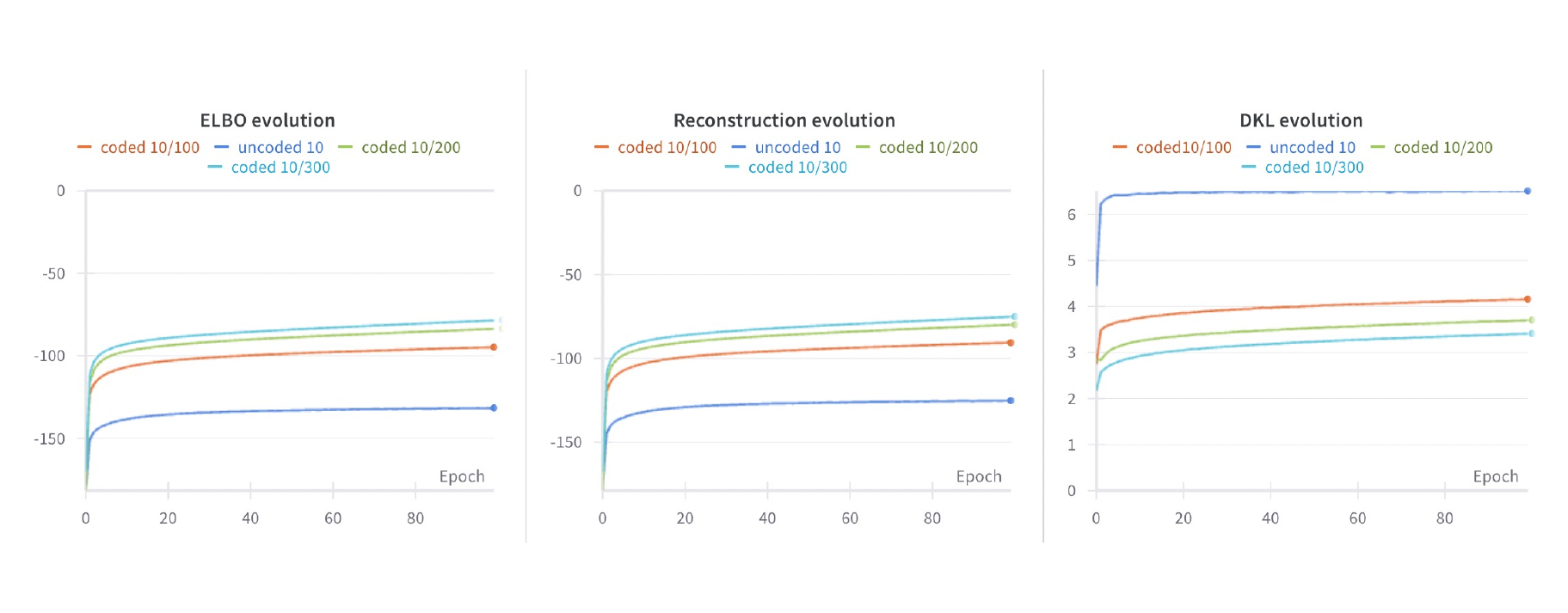}
  \caption{Evolution of the \ac{elbo} during training with 10 information bits on MNIST.}\label{fig:elbo_mnist_10}
\end{figure*}

We present the evolution of the \ac{elbo} and its terms throughout the training process. The models were trained for 100 epochs using an Adam optimizer with a learning rate of $10^{-4}$, and a batch size of 128. Fig. \ref{fig:elbo_mnist_5} displays the results for configurations with 5 information bits, Fig. \ref{fig:elbo_mnist_8} for 8 information bits, and Fig. \ref{fig:elbo_mnist_10} for 10 information bits. The colors in all plots represent the various code rates.


The results are consistent with the ones obtained for FMNIST. Across all the configurations, coded models achieve superior bounds. The main differences in the \ac{elbo} come from the different performances in reconstruction. As we have observed across the different experiments, coded models are capable of better capturing the structure of the data, generating more detailed images and accurate reconstructions.

We observe that, as we decrease the code rate, we obtain better bounds in general. Adding redundancy does not increase the model’s flexibility, since the information bits determine the number of latent vectors. However, the introduction of \acp{ecc} in the model allows for latent spaces that better capture the structure of the images while employing the same number of latent vectors.

\subsection{Reconstruction}

We first evaluate the model’s performance in reconstructing data by examining its \textit{uncoded} and \textit{coded} versions across different configurations, varying the number of information bits and code rates. All the results obtained with MNIST are consistent with those presented in the main text for FMNIST.

\begin{figure*}[!t]
  \centering
  \includegraphics[width=1\linewidth]{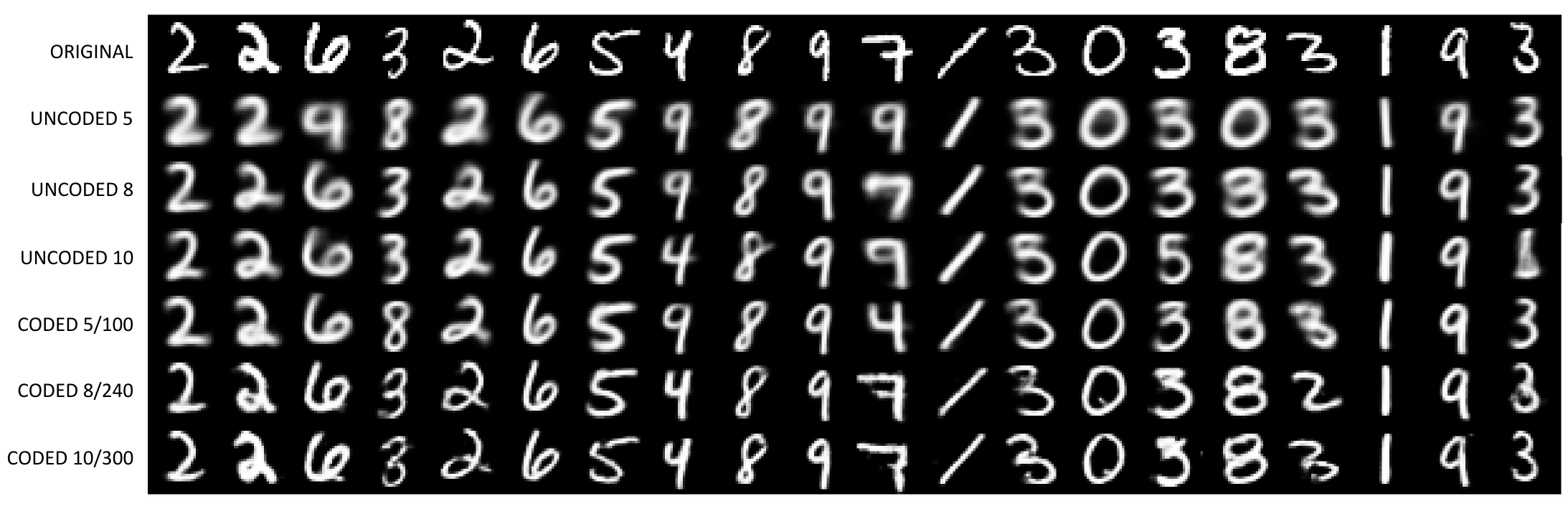}
  \caption{Example of reconstructed test images obtained with different model configurations. Observe that more details are visualized as we increase bits in the latent space and decrease the coding rate.}\label{fig:reconstruction_mnist}
\end{figure*}

In Table \ref{tab:metrics_rec_mnist} we quantify the quality of the reconstructions measuring the \ac{psnr} and the \ac{ssim} in both training and test sets. In all the cases, coded models yield higher \ac{psnr} and \ac{ssim} values, indicating a superior performance in reconstruction. This improvement is also evident through visual inspection of Fig. \ref{fig:reconstruction_mnist}, where the coded models better capture the details in the images. Similar to the results observed on FMNIST, we find that both \ac{psnr} and \ac{ssim} generally improve as the number of information bits increases and the code rate decreases.

\begin{figure}[!t]
  \centering
  \includegraphics[width=1\linewidth]{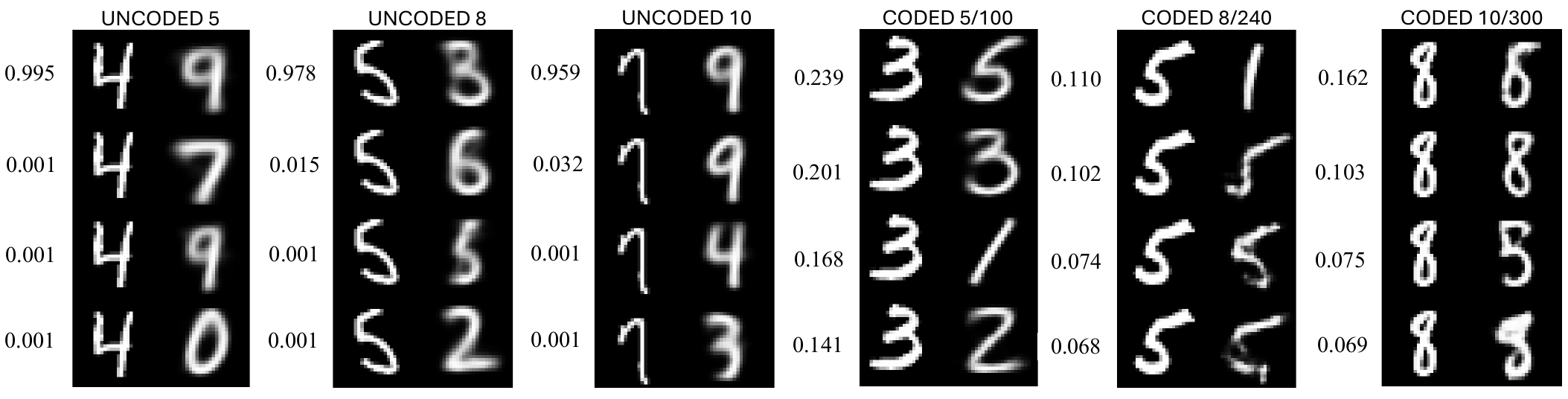}
  \caption{Example of erroneous reconstructions in MNIST using the 4 most-probable words (a posterior). The first column in each image shows the original input to the model, while the second column displays the reconstructions. The a posteriori probability of the word used for reconstruction is indicated in each row.}\label{fig:errors_mnist}
\end{figure}

\begin{table*}[!t]
\centering
    \caption{Evaluation of reconstruction performance in MNIST.} \label{tab:metrics_rec_mnist}
    \begin{tabular}{cccccccccc}
      \toprule 
      \bfseries Model & \begin{tabular}{@{}c@{}}\bfseries{PSNR}\\ \bfseries{(train)}\end{tabular}&\begin{tabular}{@{}c@{}}\bfseries{SSIM}\\ \bfseries{(train)}\end{tabular}& \begin{tabular}{@{}c@{}}\bfseries{Acc}\\ \bfseries{(train)}\end{tabular} & \begin{tabular}{@{}c@{}}\bfseries{Conf. Acc.}\\ \bfseries{(train)}\end{tabular} & \begin{tabular}{@{}c@{}}\bfseries{PSNR}\\ \bfseries{(test)}\end{tabular} & \begin{tabular}{@{}c@{}}\bfseries{SSIM}\\ \bfseries{(test)}\end{tabular} & \begin{tabular}{@{}c@{}}\bfseries{Acc}\\ \bfseries{(test)}\end{tabular}  & \begin{tabular}{@{}c@{}}\bfseries{Conf. Acc.}\\ \bfseries{(test)}\end{tabular} & \bfseries Entropy\\
      \midrule 
      uncoded 5  & 13.491  & 0.434 & 0.702 & 0.703 & 13.483 & 0.430 & 0.701 & 0.702& 0.277\\
      coded 5/50 & 14.983 & 0.601 &  0.887 & 0.923 & 14.888 & 0.598 & 0.887 & 0.920& 2.073\\
      coded 5/80 & 15.436 & 0.641 & 0.899 & 0.936 & 15.263 & 0.634 & 0.895 & 0.929& 2.237 \\
      coded 5/100 & 15.590 & 0.652 & 0.905 & 0.931 & 15.352 & 0.641 & 0.898 & 0.924& 2.382\\
      \midrule
      uncoded 8  & 14.530 & 0.558 & 0.860 & 0.864 & 14.490 & 0.558 & 0.860 & 0.868& 0.513\\
      coded 8/80 & 16.878 & 0.739 & 0.937 & 0.964 & 16.042 & 0.699 & 0.912 & 0.947& 3.105\\
      coded 8/160 & 18.108 & 0.802 & 0.957 & 0.974 & 16.497 & 0.736 & 0.927 & 0.951& 3.645 \\
      coded 8/240 & 19.984 & 0.838 & 0.967 & 0.978 & 16.688 & 0.752 & 0.936 & 0.957& 3.881 \\
      \midrule
      uncoded 10 &  14.879 & 0.591 & 0.888 & 0.891 & 14.816 & 0.589 & 0.887 & 0.890 &  0.636\\
      coded 10/100 & 17.584 & 0.777 & 0.945 & 0.972 & 16.795 & 0.744 & 0.928 & $\mathbf{0.968}$& 4.080 \\
      coded 10/200 & 20.060 & 0.875 & 0.973 & 0.977 & 16.863 & 0.765 & 0.932 & 0.944 & 4.411\\
      coded 10/300 & $\mathbf{21.083}$ & $\mathbf{0.902}$ & $\mathbf{0.979}$ & $\mathbf{0.984}$ & $\mathbf{17.114}$ & $\mathbf{0.781}$ & $\mathbf{0.941}$ & 0.945 & 4.810\\
      \bottomrule 
    \end{tabular}

\end{table*}

As we discussed in the main text, reconstruction metrics such as \ac{psnr} and \ac{ssim} do not account for the \textit{semantic} errors committed by the model. Therefore, we additionally evaluate the reconstruction accuracy, ensuring that the model successfully reconstructs images within the same class as the original ones. We also provide a \textit{confident} reconstruction accuracy, for which we do not count errors when the \ac{map} value of $q(\m|\x)$ is below 0.4. In light of the results, we argue that introducing \acp{ecc} in the model allows for latent spaces that better capture the semantics of the images while employing the same number of latent vectors, outperforming the uncoded models in all the cases.

We also report the average entropy of the variational posterior over the test set in Table \ref{tab:metrics_rec_mnist}. If we consider the entropy together with the semantic accuracy, we can argue that coded \ac{vae} is aware that multiple vectors might be related to the same image class, and that the posterior shows larger uncertainties for images for which the model has not properly identified the class. We illustrate this argument in Fig. \ref{fig:errors_mnist}, where we show some images selected so that the MAP latent word of $q(\m|\x)$ induces class reconstruction errors. We show the reconstruction of the 4 most probable latent vectors and their corresponding probabilities. Observe that the uncoded model is confident no matter the reconstruction outcome, while in the coded posterior, the uncertainty is much larger.

Table \ref{tab:loglik_mnist} shows the log-likelihood values obtained for the MNIST dataset with various model configurations. Coded models consistently outperform their uncoded counterparts for both the training and test sets, consistent with the findings observed using the FMNIST dataset.

\subsection{Generation}

\begin{figure*}[!h]
  \centering
  \includegraphics[width=1\linewidth]{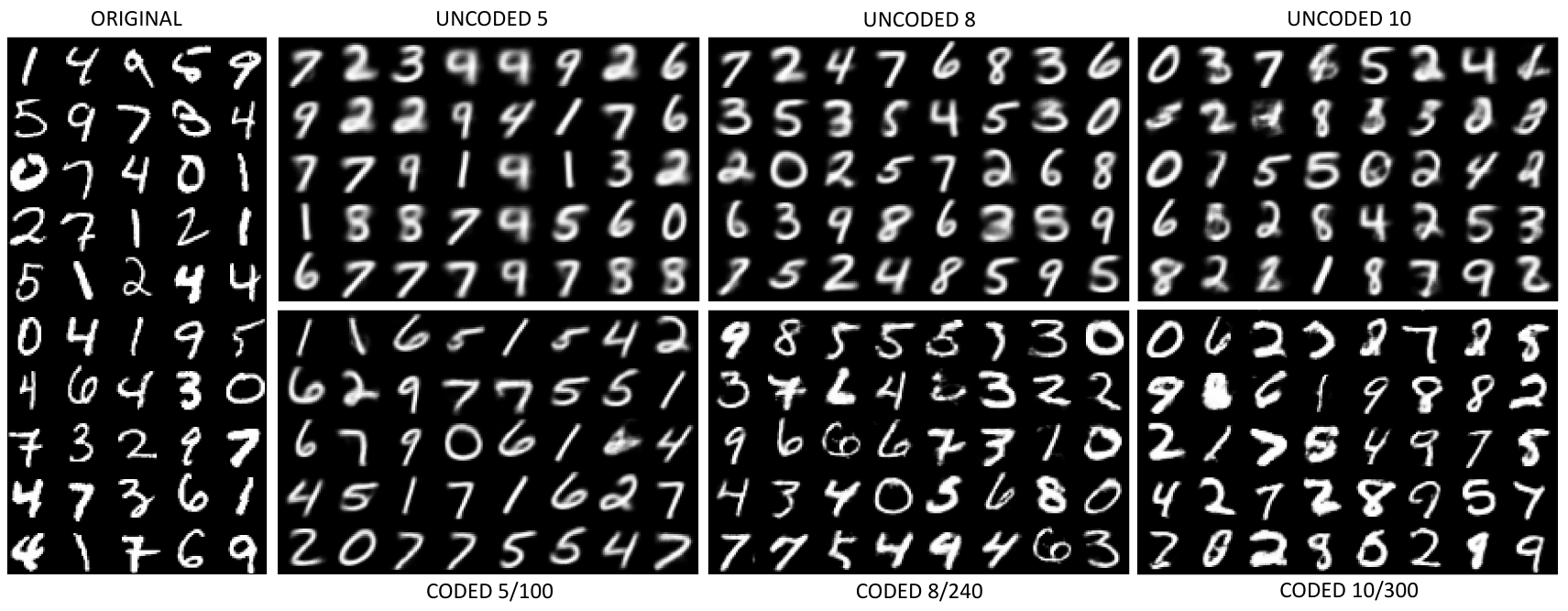}
  \caption{Example of randomly generated, uncurated images using different model configurations.}\label{fig:generation_mnist}
\end{figure*}

In this section, we evaluate the model in the image generation task. In Fig. \ref{fig:generation_mnist}, we show examples of randomly generated images using different model configurations in MNIST. These results are consistent with the ones obtained in reconstruction, and with the ones obtained for FMNIST, where we observe that the coded models can generate more detailed and diverse images.

The improved inference provided by the repetition code can also be tested by generating images using the generative model and counting errors using the MAP solution of the variational posterior distribution, as well as sampling from the approximate posterior distribution. Table \ref{tab:error_mnist} reports the \ac{ber} and \ac{wer} for MNIST. Remarkably, for the same number of latent bits, coded models reduce both the \ac{ber} and \ac{wer} w.r.t. the uncoded case. Note also that the error rates grow with the number of latent bits, but this is expected due to the increased complexity of the inference process. 

\begin{table} [!h]
\parbox{.43\linewidth}{
\centering
\caption{Evaluation of the \ac{ber}, \ac{wer} in MNIST.}\label{tab:error_mnist}
\begin{tabular}{ccccc}
      \toprule 
      \bfseries Model & \bfseries BER & \bfseries BER MAP & \bfseries WER & \bfseries WER MAP\\
      \midrule 
      uncoded 5 & 0.002 & 0.001 & 0.008 & 0.004\\
      coded 5/50 & 0.007 & 0.000 & 0.034 & 0.000\\
      coded 5/80 & 0.004 &0.000 & 0.021 & 0.000\\
      coded 5/100 & 0.009 & 0.000 & 0.045 & 0.000 \\
      \midrule
      uncoded 8 & 0.015 & 0.013 & 0.071 & 0.057\\
      coded 8/80 & 0.020 & 9.750$\cdot10^{-5}$ & 0.147 & 7.800$\cdot10^{-4}$\\
      coded 8/160 & 0.021 & 2.500$\cdot10^{-5}$ & 0.160 & 2.000$\cdot10^{-4}$\\
      coded 8/240 & 0.023 & 7.800$\cdot10^{-4}$& 0.167 & 0.006\\
      \midrule
      uncoded 10 & 0.057 & 0.052& 0.373 & 0.353\\
      coded 10/100 & 0.030 & 2.040$\cdot10^{-4}$ & 0.258 & 0.002\\
      coded 10/200 & 0.034 & 3.260$\cdot10^{-4}$ & 0.282 & 0.003\\
      coded 10/300 & 0.041 & 9.600$\cdot10^{-4}$ & 0.331 & 0.009\\
      \bottomrule 
\end{tabular}
}
\hfill
\parbox{.47\linewidth}{

\centering
\caption{Evaluation of the \ac{ll} in MNIST.} \label{tab:loglik_mnist}
\begin{tabular}{ccc}
      \toprule 
      \bfseries Model & \bfseries LL (train) & \bfseries LL (test) \\
      \midrule 
      uncoded 5  & -149.049 &  -148.997\\
      coded 5/50 & -117.979 & -119.094 \\
      coded 5/80 & -114.911 & -116.639 \\
      coded 5/100 & -115.189 & -117.200\\
      \midrule
      uncoded 8  & -127.079 &  -127.555 \\
      coded 8/80 & -96.554 & -104.692\\
      coded 8/160 & -96.014 & -107.436 \\
      coded 8/240 & -97.316 & -111.312 \\
      \midrule
      uncoded 10 & -120.594 & -121.332 \\
      coded 10/100 & -92.545 & -99.373 \\
      coded 10/200 & -86.072 & -106.249 \\
      coded 10/300 & -88.904 & -110.799 \\
      \bottomrule 
\end{tabular}
}
\end{table}

\newpage
\section{CIFAR10 results}\label{cifar_results_supp}

In this section, we provide additional results using the CIFAR10 dataset with different model configurations.

\subsection{Training}

\begin{figure*}[!b]
  \centering
  \includegraphics[width=1\linewidth]{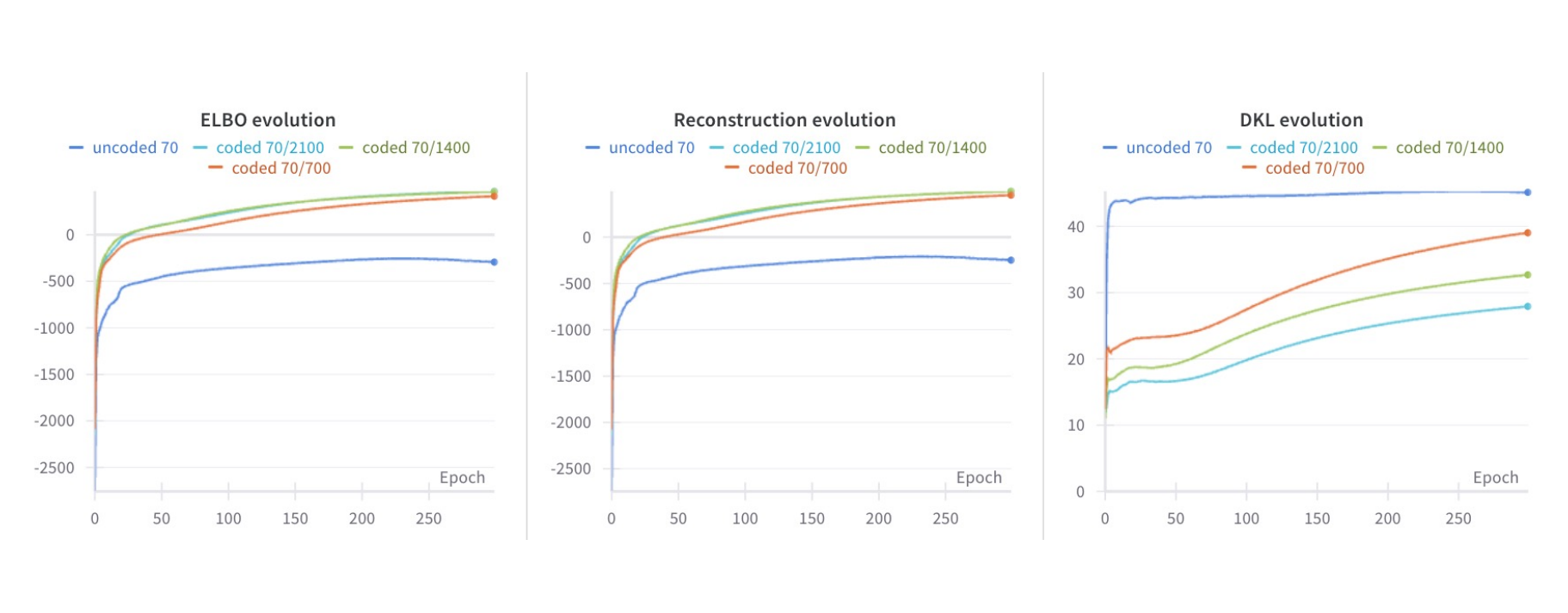}
  \caption{Evolution of the \ac{elbo} during training with 70 information bits on CIFAR10.}\label{fig:elbo_cifar_70}
\end{figure*}

\begin{figure*}[!h]
  \centering
  \includegraphics[width=1\linewidth]{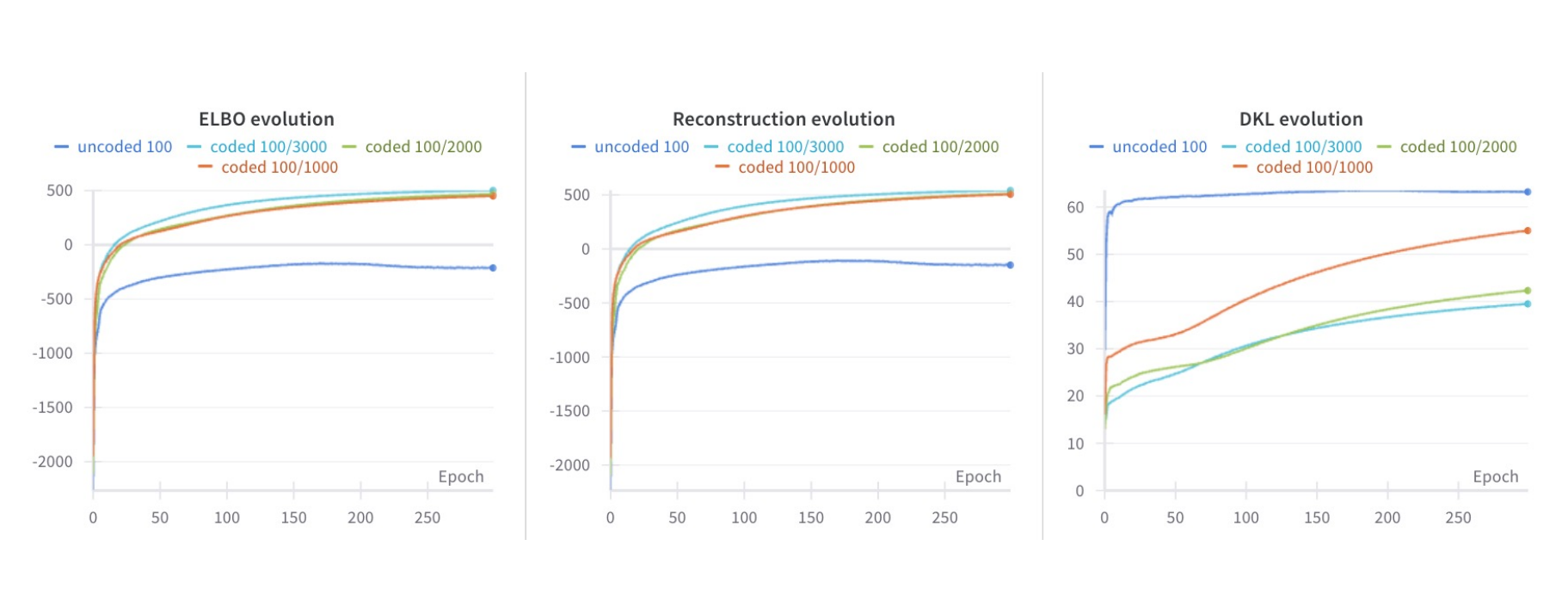}
  \caption{Evolution of the \ac{elbo} during training with 100 information bits on CIFAR10.}\label{fig:elbo_cifar_100}
\end{figure*}

\begin{figure*}[!ht]
  \centering
  \includegraphics[width=1\linewidth]{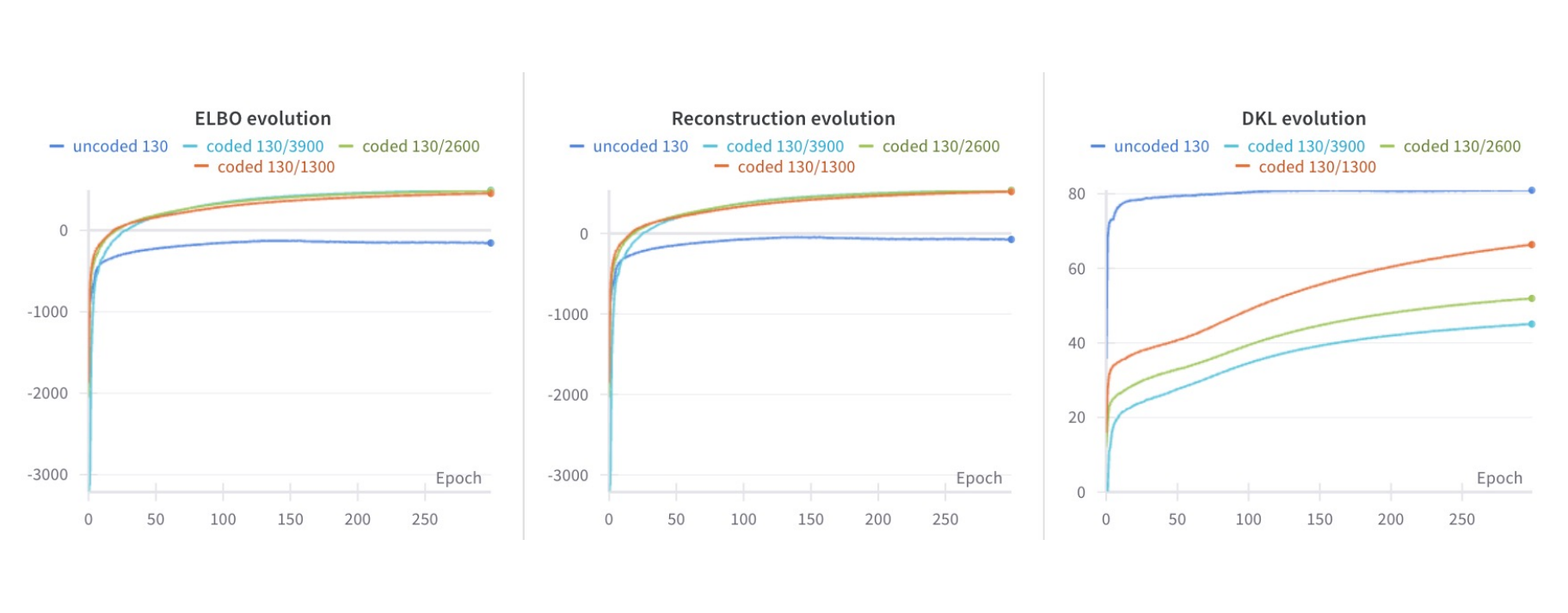}
  \caption{Evolution of the \ac{elbo} during training with 130 information bits on CIFAR10.}\label{fig:elbo_cifar_130}
\end{figure*}

We present the evolution of the \ac{elbo} and its terms throughout the training process. The models were trained for 300 epochs using Adam optimizer with a learning rate of $10^{-4}$, and a batch size of 128. Fig. \ref{fig:elbo_cifar_70} displays the results for configurations with 70 information bits, Fig. \ref{fig:elbo_cifar_100} for 100 information bits, and Fig. \ref{fig:elbo_cifar_130} for 130 information bits. The colors in all plots represent the various code rates.

Across all the configurations, coded models achieve superior bounds. The main differences in the \ac{elbo} come from the different performances in reconstruction. As we have observed across the different experiments, coded models are capable of better capturing the structure of the data, generating more detailed images and accurate reconstructions.

In the coded case, we do not observe significant differences in the obtained bounds as we increase the number of information bits and reduce the code rate. However, the difference is notable if we compare the coded and uncoded models. Adding redundancy does not increase the model’s flexibility, since the information bits determine the number of latent vectors. However, the introduction of \acp{ecc} in the model allows for latent spaces that better capture the structure of the images while employing the same number of latent vectors.

It's important to note that we are currently using feed-forward networks at the decoder's input. However, this approach may not be suitable for the correlations present in our coded words. Utilizing an architecture capable of effectively leveraging these correlations among the coded bits could potentially enable us to better exploit the introduced redundancy.

\subsection{Reconstruction and generation}

We first evaluate the model’s performance in reconstructing data by examining its \textit{uncoded} and \textit{coded} versions across different configurations, varying the number of information bits and code rates. 

\begin{figure*}[!t]
  \centering
  \includegraphics[width=1\linewidth]{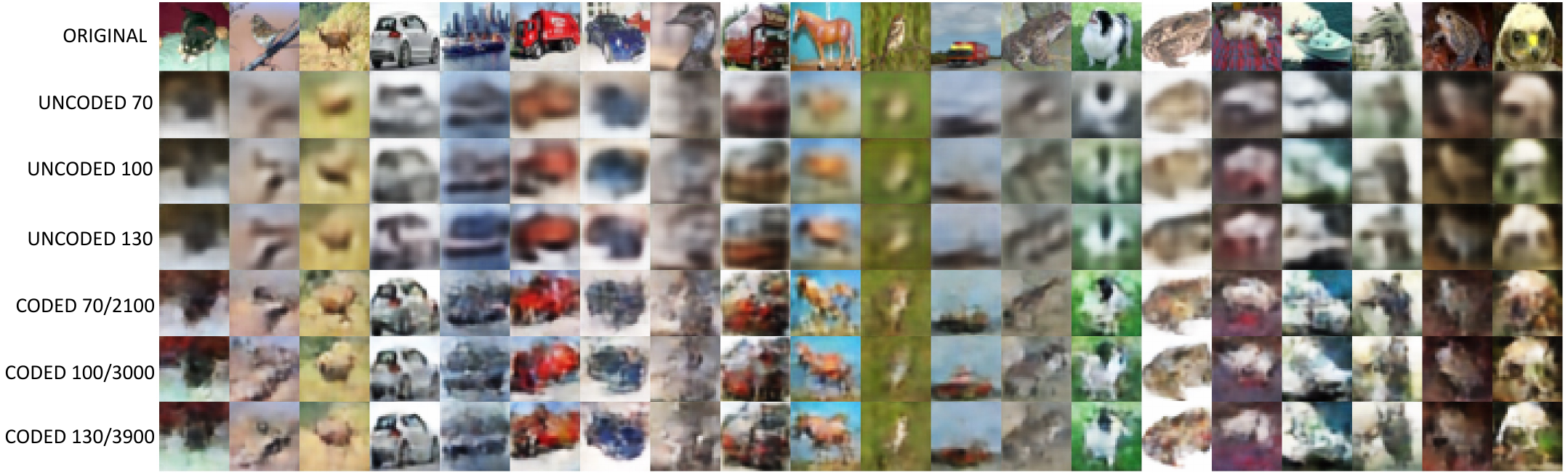}
  \caption{Example of reconstructed test images obtained with different model configurations. Observe that more details are visualized as we increase bits in the latent space and introduce redundancy.}\label{fig:reconstruction_cifar_sup}
\end{figure*}

In Table \ref{tab:metrics_cifar} we quantify the quality of the reconstructions measuring the \ac{psnr} and the \ac{ssim} in both training and test sets. Coded models yield higher \ac{psnr} and \ac{ssim} values in train, and similar values in test, although the coded models with lower rates outperform the rest of the configurations. However, the improvement in reconstruction is evident through visual inspection of Fig. \ref{fig:reconstruction_cifar_sup}, where the coded models better capture the details in the images. We observe that the coded model produces images that better resemble the structure of the dataset, while the uncoded \ac{dvae} cannot decouple spatial information from the images and project it in the latent space. 

We hypothesize that, to adequately model complex images, transitioning to a hierarchical structure may be necessary. This would allow for explicit modeling of both global and local information. However, despite employing this rather simple model, we observe that coded configurations outperform their uncoded counterparts in capturing colors and textures.

\begin{figure}[!t]
  \centering
  \includegraphics[width=1\linewidth]{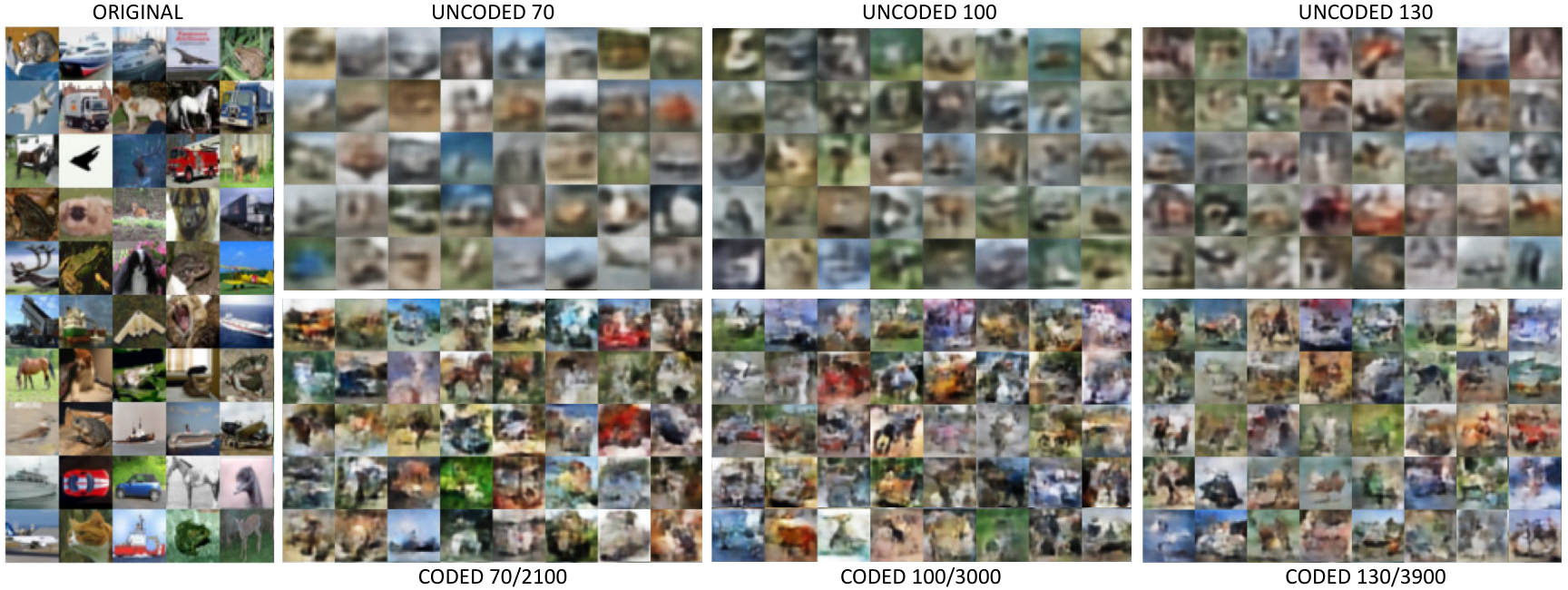}
  \caption{Example of randomly generated, uncurated images using different model configurations.}\label{fig:gen_cifar_sup}
\end{figure}

We also evaluate the model in the image generation task. In Fig. \ref{fig:gen_cifar_sup}, we show examples of randomly generated images using different model configurations in CIFAR10. These results are consistent with the ones obtained in reconstruction, as we observe that the coded models can generate more detailed and diverse images. Additionally, we obtained the \ac{fid} score using the test set and 10k generated samples. For this, we used the implementation available at \url{https://github.com/mseitzer/pytorch-fid}. We can observe that the coded models significantly reduced the \ac{fid} score in all the cases compared to their uncoded counterparts. For the coded models, we do not observe a clear influence of the code rate on the quality of the generations. 

\begin{table}[!ht]
\parbox{.48\linewidth}{

\centering
\caption{Reconstruction metrics in CIFAR10 with different model configurations.}\label{tab:metrics_cifar}
\begin{tabular}{ccccc}
      \toprule 
      \bfseries{Model} & \begin{tabular}{@{}c@{}}\bfseries{PSNR}\\ \bfseries{(train)}\end{tabular} & \begin{tabular}{@{}c@{}}\bfseries{SSIM}\\ \bfseries{(train)}\end{tabular}&  \begin{tabular}{@{}c@{}}\bfseries{PSNR}\\ \bfseries{(test)}\end{tabular} & \begin{tabular}{@{}c@{}}\bfseries{SSIM}\\ \bfseries{(test)}\end{tabular}\\
      \midrule 
      uncoded 70  & 17.985 & 0.340 & 17.596 & 0.324 \\
      coded 70/700  & 23.790 & 0.709 & 17.731& 0.323\\
      coded 70/1400  & 24.555 & 0.748 & 18.008 & 0.357 \\
      coded 70/2100  & 25.551 & 0.748 & 18.401 & 0.385 \\
      \midrule
      uncoded 100  & 18.509 & 0.381 & 18.334& 0.370  \\
      coded 100/1000  & 24.754 & 0.754 & 18.229 & 0.375 \\
      coded 100/2000  & 24.866 & 0.761 & 18.927 & 0.432 \\
      coded 100/3000  & 25.646 & 0.793 & 18.920 & 0.426 \\
      \midrule
      uncoded 130  & 18.951 & 0.419 & 18.758 & .408 \\
      coded 130/1300  & 25.007 & 0.767 & 18.887 & 0.426 \\
      coded 130/2600  & 25.460 & 0.784 & $\mathbf{19.416}$ &$\mathbf{0.464}$ \\
      coded 130/3900  & $\mathbf{25.515}$ & $\mathbf{0.785}$ & 19.292 & 0.456 \\
      \bottomrule 
\end{tabular}
}
\hfill
\parbox{.49\linewidth}{

\centering
\caption{Evaluation of the \ac{ber}, \ac{wer}, and FID in CIFAR10 with different model configurations.}\label{tab:error_cifar}
\begin{tabular}{cccccc}
      \toprule 
      \bfseries{Model} & \bfseries BER & \begin{tabular}{@{}c@{}}\bfseries{BER}\\ \bfseries{MAP}\end{tabular} & \bfseries WER  & \begin{tabular}{@{}c@{}}\bfseries{WER}\\ \bfseries{MAP}\end{tabular} & \bfseries FID\\
      \midrule 
      uncoded 70  & 0.162 & 0.162 & 1.000 & 0.999 & 177.524\\
      coded 70/700  & 0.101 & 0.081 & 1.000 & 0.988&  104.977\\
      coded 70/1400  & 0.088 & 0.044 & 0.999 & 0.917 &104.078\\
      coded 70/2100  & 0.090 & 0.029 & 0.999 & 0.811 & 102.795\\
      \midrule
      uncoded 100  & 0.182 & 0.179 & 1.000 & 1.000 & 172.063\\
      coded 100/1000  & 0.123 & 0.104 & 1.000 & 0.999 & 107.887\\
      coded 100/2000  & 0.114 & 0.060 & 1.000 & 0.990 & 101.182\\
      coded 100/3000  & 0.138 & 0.077 & 1.000 & 0.997 & 107.287\\
      \midrule
      uncoded 130  & 0.197 & 0.194 & 1.000 & 1.000 & 164.138\\
      coded 130/1300  & 0.144 & 0.115 & 1.000 & 0.999 & 109.905\\
      coded 130/2600  & 0.164 & 0.102 & 1.000 & 1.000 & 110.250\\
      coded 130/3900  & 0.185 & 0.132 & 1.000 & 1.000 & 108.561\\
      \bottomrule 
\end{tabular}
}
\end{table}

\newpage
\section{Tiny ImageNet results}\label{tinyimagenet_results_supp}

In this section, we provide additional results using the Tiny ImageNet dataset with different model configurations.

\subsection{Training}

\begin{figure*}[!b]
  \centering
  \includegraphics[width=1\linewidth]{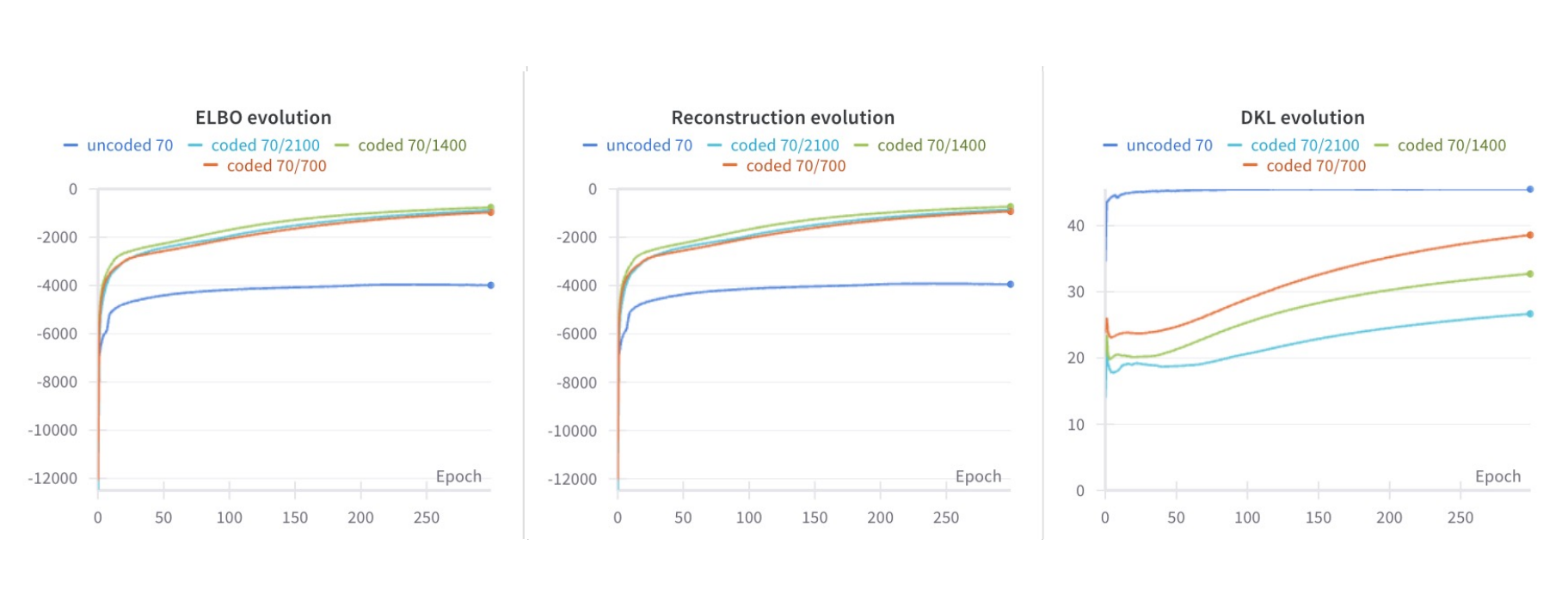}
  \caption{Evolution of the \ac{elbo} during training for the configurations with 70 information bits.}\label{fig:elbo_imagenet_70}
\end{figure*}

\begin{figure*}[!h]
  \centering
  \includegraphics[width=1\linewidth]{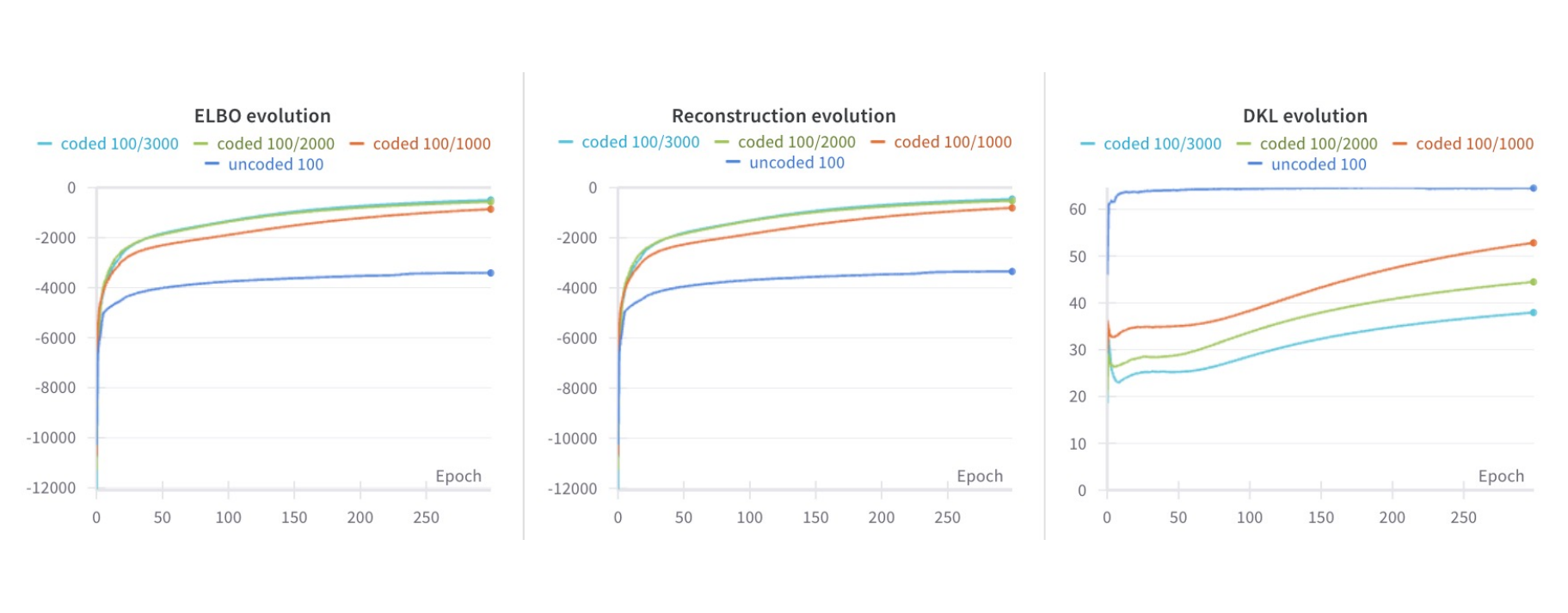}
  \caption{Evolution of the \ac{elbo} during training for the configurations with 100 information bits.}\label{fig:elbo_imagenet_100}
\end{figure*}

\begin{figure*}[!h]
  \centering
  \includegraphics[width=1\linewidth]{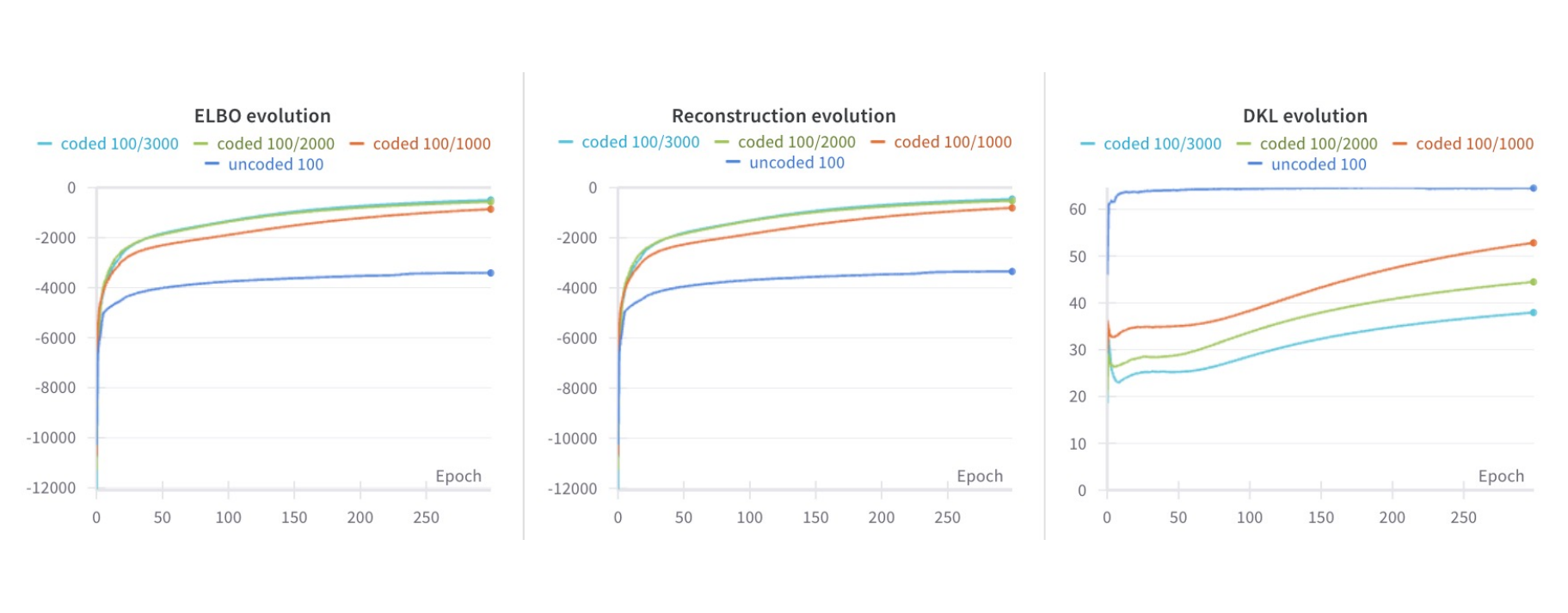}
  \caption{Evolution of the \ac{elbo} during training for the configurations with 130 information bits.}\label{fig:elbo_imagenet_130}
\end{figure*}

We present the evolution of the \ac{elbo} and its terms throughout the training process. The models were trained for 300 epochs using an Adam optimizer with a learning rate of $10^{-4}$, and a batch size of 128. Fig. \ref{fig:elbo_imagenet_70} displays the results for configurations with 70 information bits, Fig. \ref{fig:elbo_imagenet_100} for 100 information bits, and Fig. \ref{fig:elbo_imagenet_130} for 130 information bits. The colors in all plots represent the various code rates.

As in the rest of the datasets, coded models achieve superior bounds across all the configurations. The main differences in the \ac{elbo} come from the different performances in reconstruction. As we have observed across the different experiments, coded models are capable of better capturing the structure of the data, generating more detailed images and accurate reconstructions.

In the coded case, we do not observe significant differences in the obtained bounds as we increase the number of information bits and reduce the code rate. However, the difference is notable if we compare the coded and uncoded models. Adding redundancy does not increase the model’s flexibility, since the information bits determine the number of latent vectors. However, the introduction of \acp{ecc} in the model allows for latent spaces that better capture the structure of the images while employing the same number of latent vectors.

It's important to note that we are currently using feed-forward networks at the decoder's input. However, this approach may not be suitable for the correlations present in our coded words. Utilizing an architecture capable of effectively leveraging these correlations among the coded bits could potentially enable us to better exploit the introduced redundancy.

\subsection{Reconstruction and generation}

\begin{figure*}[!t]
  \centering
  \includegraphics[width=1\linewidth]{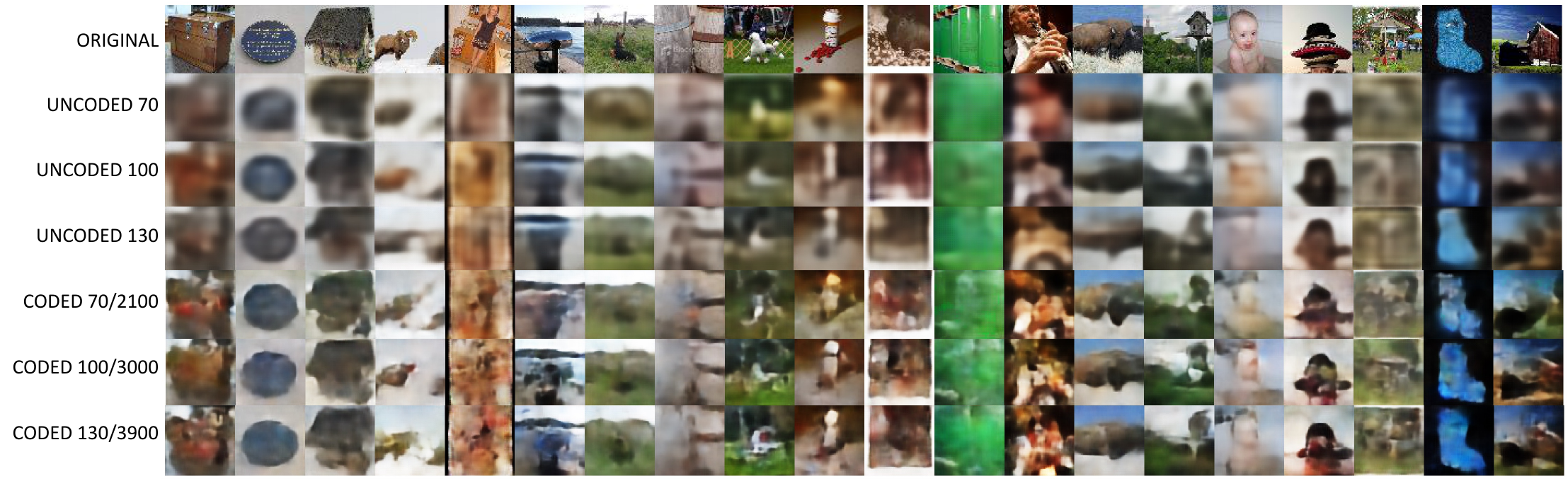}
  \caption{Example of reconstructed test images obtained with different model configurations. Observe that more details are visualized as we increase the bits in the latent space and introduce redundancy.}\label{fig:rec_imagenet_supp}
\end{figure*}

We first evaluate the model’s performance in reconstructing data by examining its \textit{uncoded} and \textit{coded} versions across different configurations, varying the number of information bits and code rates. 

In Table \ref{tab:metrics_imagenet}, we quantify the quality of the reconstructions measuring the \ac{psnr} and the \ac{ssim} in both training and test sets. Coded models yield higher \ac{psnr}  and \ac{ssim} values in train, and similar values in test, although coded models with lower rates outperform the rest of the configurations. However, the improvement in reconstruction is evident through visual inspection of Fig. \ref{fig:rec_imagenet_supp}, where the coded models better capture the details in the images. We observe that the coded model produces images that better resemble the structure of the dataset, while the uncoded \ac{dvae} cannot decouple spatial information from the images and project it in the latent space. 

We hypothesize that, to adequately model complex images, transitioning to a hierarchical structure may be necessary. This would allow for the explicit modeling of both global and local information. However, despite employing this rather simple model, we observe that coded configurations outperform their uncoded counterparts in capturing colors and textures.

\begin{figure*}[tb]
  \centering
  \includegraphics[width=1\linewidth]{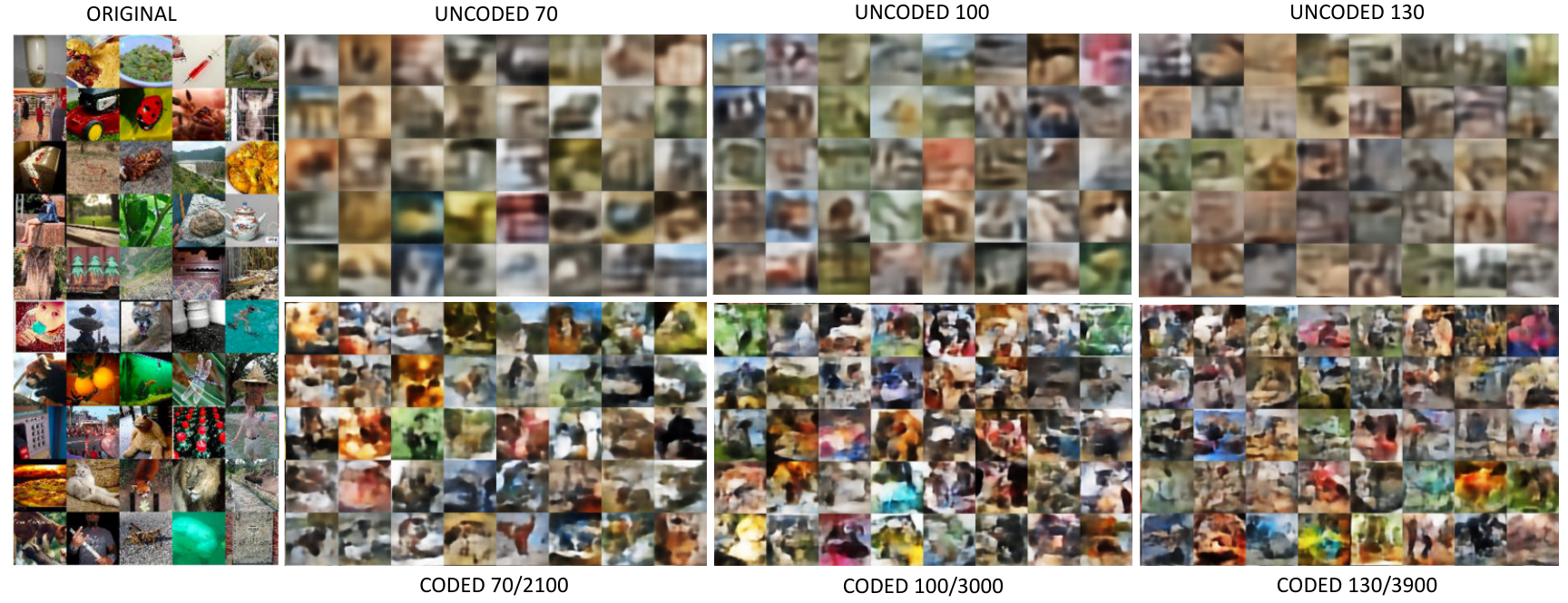}
  \caption{Example of randomly generated, uncurated images using different model configurations.}\label{fig:gen_imagenet_supp}
\end{figure*}

We also evaluate the model in the image generation task. In Fig. \ref{fig:gen_imagenet_supp}, we show examples of randomly generated images using different model configurations in Tiny ImageNet. These results are consistent with the ones obtained in reconstruction, where we observe that the coded models can generate more detailed and diverse images. Additionally, we obtained the \ac{fid} score using the test set and 10k generated samples. For this, we used the implementation available at \url{https://github.com/mseitzer/pytorch-fid}. We can observe that the coded models significantly reduced the \ac{fid} score in all the cases compared to their uncoded counterparts.

For the coded models, we do not observe a clear influence of the code rate on the quality of the generations in CIFAR-10. However, in Tiny ImageNet, smaller code rates produce worse \ac{fid} scores. We hypothesize that this may be due to the presence of artifacts in the generated images. Our experiments indicate that coded models with lower rates attempt to model fine details in images, which can lead to artifacts in generation.

\begin{table}
\parbox{.48\linewidth}{

\centering
\caption{Reconstruction metrics in Tiny ImageNet with different model configurations.}\label{tab:metrics_imagenet}
\begin{tabular}{ccccc}
      \toprule 
      \bfseries{Model} & \begin{tabular}{@{}c@{}}\bfseries{PSNR}\\ \bfseries{(train)}\end{tabular} & \begin{tabular}{@{}c@{}}\bfseries{SSIM}\\ \bfseries{(train)}\end{tabular}&  \begin{tabular}{@{}c@{}}\bfseries{PSNR}\\ \bfseries{(test)}\end{tabular} & \begin{tabular}{@{}c@{}}\bfseries{SSIM}\\ \bfseries{(test)}\end{tabular}\\
      \midrule 
      uncoded 70  & 15.598 & 0.207 & 15.402& 0.196 \\
      coded 70/700  & 18.156 & 0.367 & 15.158 & 0.191\\
      coded 70/1400  & 18.396 & 0.383 & 15.419& 0.203\\
      coded 70/2100  & 18.228 & 0.369 & 15.789& 0.223\\
      \midrule
      uncoded 100  & 16.012 & 0.229 & 15.774& 0.215 \\
      coded 100/1000  & 18.298 & 0.378 & 15.677& 0.218 \\
      coded 100/2000  & 18.647 & 0.404 & 15.892& 0.232\\
      coded 100/3000  & 18.729 & 0.408 & 16.167& 0.248\\
      \midrule
      uncoded 130  & 16.278 & 0.243 & 16.009 & 0.228\\
      coded 130/1300  & 18.719 & 0.409 & 15.900 & 0.233\\
      coded 130/2600  & 18.818 & $\mathbf{0.433}$ & $\mathbf{16.329}$& $\mathbf{0.259}$\\
      coded 130/3900  & $\mathbf{19.020}$ & 0.415 & 16.288 & $\mathbf{0.259}$\\
      \bottomrule 
\end{tabular}
}
\hfill
\parbox{.49\linewidth}{

\centering
\caption{Evaluation of the \ac{ber}, \ac{wer}, and \ac{fid} in Tiny ImageNet with different model configurations.}\label{tab:error_cifar}
\begin{tabular}{cccccc}
      \toprule 
      \bfseries{Model} & \bfseries BER & \begin{tabular}{@{}c@{}}\bfseries{BER}\\ \bfseries{MAP}\end{tabular} & \bfseries WER  & \begin{tabular}{@{}c@{}}\bfseries{WER}\\ \bfseries{MAP}\end{tabular} & \bfseries FID\\
      \midrule 
      uncoded 70  & 0.143 & 0.140 & 1.000 & 0.999 & 265.474\\
      coded 70/700  & 0.096 & 0.072 &  0.998 & 0.978 & 171.993\\
      coded 70/1400  & 0.104 & 0.066 &  1.000 & 0.972 & 170.496\\
      coded 70/2100  & 0.096 & 0.034 & 0.998 & 0.832 & 176.245\\
      \midrule
      uncoded 100  & 0.164 & 0.162 & 1.000 & 1.000 & 234.358\\
      coded 100/1000  & 0.099 & 0.074 & 1.000 & 0.996 & 153.743 \\
      coded 100/2000  & 0.097 & 0.053 & 1.000 & 0.981 & 162.889\\
      coded 100/3000  & 0.098 & 0.035 & 1.000 & 0.925 & 163.049\\
      \midrule
      uncoded 130  & 0.200 & 0.198 & 1.000 & 1.000 & 219.003\\
      coded 130/1300  & 0.129 & 0.107 & 1.000 & 0.999 & 165.064\\
      coded 130/2600  & 0.114 & 0.060 & 1.000 & 0.996 & 164.759\\
      coded 130/3900  & 0.128 & 0.070 & 1.000 & 0.999 & 170.603\\
      \bottomrule 
\end{tabular}
}
\end{table}

\section{IWAE results}\label{iwae_results}

One could draw a parallel between the Coded-\ac{dvae} with repetition codes and the well-known \ac{iwae} \citep{iwae}, but the two approaches are fundamentally different. In the \ac{iwae}, independent samples are drawn from the variational posterior and propagated independently through the generative model to obtain a tighter variational bound on the marginal log-likelihood. In our method, we jointly propagate the output of the \ac{ecc} encoder through the generative model, obtaining a single prediction and exploiting the introduced known correlations in the variational approximation of the posterior. In the case of repetition codes, the \ac{ecc} encoder outputs are repeated bits, or repeated probabilities in the case of soft encoding. However, our approach extends beyond repetition codes, opening a new field for improved inference in discrete latent variable models.

In our approach, we specifically utilize the redundancy introduced by the repetition code to correct potential errors made by the encoder through a soft decoding approach. This results in a more accurate approximation of the posterior $p(\m|\x)$ and an improved proposal for sampling, resulting in improved performance over uncoded models, even those trained with the tighter \ac{iwae} bound. To compute the \ac{iwae} objective in the uncoded case, we draw samples from the posterior using the reparameterization trick described in Eq. \eqref{CDF_dvae}, and compute the importance weights as $\boldsymbol{w}=\frac{p_{\boldsymbol{\theta}}(\x|\z)p(\z)}{q_{\e}(\z|\x)}$. Here, the prior and posterior over $\z$ are obtained by marginalizing out $\m$ in $p(\z, \m) = p(\m)p(\z|\m)$ and $q_{\e}(\m,\z|\x) = q_{\e}(\m|\x)p(\z|\m)$, respectively.

Additionally, we implemented two extensions of the \ac{iwae} bound to confirm that the observed improvements are not due to a known shortcoming of the reparametrized gradient estimator of the \ac{iwae} bound. Following \citet{daudel2023alpha}, we implemented the VR-IWAE generalization, which introduces a hyperparameter $\alpha \in [0,1)$ into the objective function. This formulation interpolates between the \ac{iwae} bound (recovered when $\alpha=0$) and the \ac{elbo} (recovered when $\alpha=1$). For our experiments, we selected a midpoint value of $\alpha = 0.5$. Additionally, we implemented the doubly-reparametrized (DR) gradient estimator associated with the VR-IWAE objective, as described in \citet{daudel2023alpha}.  We trained the models on FMNIST using 10, 20, and 30 posterior samples, matching the number of repetitions used in our coded models. We included an additional experiment training an uncoded model with the IWAE bound considering 100 samples.

From the results, outlined in Table \ref{tab:iwae}, we observe that training uncoded models with tighter bounds (even if we use bounds that are meant to become tighter as the number of samples increases) does not lead to substantial improvements in performance. These models consistently underperform relative to their coded counterparts, which are trained using the \ac{elbo} as the objective function. For a fair comparison, we use the same neural network architecture for the uncoded and the coded posterior (differing only in the final dense layer to match the respective output dimensionalities), the same network architecture for the decoder (differing only in the first dense layer to match the respective input dimensionalities), and the same simple independent prior.

\begin{figure}[!t]
  \centering
  \includegraphics[width=1\linewidth]{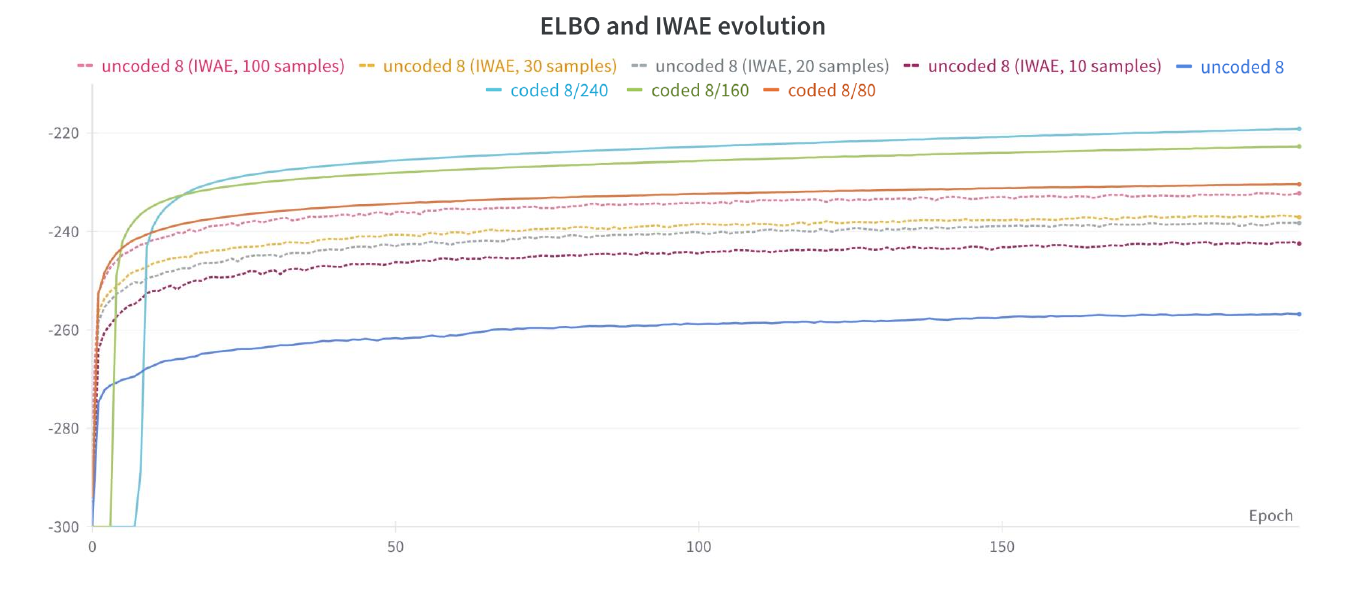}
  \caption{Evolution of the ELBO and the IWAE objectives for various configurations. Observe that the IWAE provides a tighter bound than the ELBO in the uncoded setting. However, coded models obtain even better bounds using the same number of samples/repetitions. }\label{fig:iwae}
\end{figure}

Fig. \ref{fig:iwae} shows the evolution of the different objectives, with coded models obtaining better bounds than the uncoded IWAE models, even when considering 100 samples. In Table \ref{tab:iwae}, we present test metrics, with coded models again overperforming their uncoded counterparts trained with the IWAE objective. We observe that some metrics slightly decline augmenting the number of samples, likely due to overfitting, since we applied a common early stopping point.

\begin{table}[!t]
\centering
    \caption{Comparison of test metrics between our method and the uncoded DVAE trained with the IWAE objective.} \label{tab:iwae}
    \begin{tabular}{cccccccc}
      \toprule 
      \bfseries Model & \bfseries BER & \bfseries WER & \bfseries Entropy & \bfseries Acc. & \bfseries Conf. Acc. & \bfseries PSNR & \bfseries SSIM\\
      \midrule 
      uncoded 8  & 0.089	& 0.384 & 0.467	& 0.594	& 0.595	& 15.598 & 0.509\\
      uncoded 8 IWAE 10 samples	& 0.063	& 0.372	& 1.309	& 0.617	& 0.640	& 14.282 & 0.470\\
      uncoded 8 IWAE 20 samples	& 0.075	& 0.447	& 1.391	& 0.634	& 0.651	& 14.237 & 0.460\\
      uncoded 8 IWAE 30 samples	& 0.074	& 0.438	& 1.564	& 0.619	& 0.641	& 13.757 & 0.457\\
      uncoded 8 IWAE 100 samples & 0.107 & 0.583 & 2.274 & 0.596 & 0.663 & 12.996 & 0.433\\
      uncoded 8 IWAE ($\alpha=0.5$, 10 samples) & 0.059 & 0.353 & 0.929 & 0.630 & 0.638 & 14.582 & 0.488\\
      uncoded 8 IWAE ($\alpha=0.5$, 20 samples) & 0.067 & 0.388 & 1.066 & 0.638 & 0.644 & 14.432 & 0.479\\
      uncoded 8 IWAE ($\alpha=0.5$, 30 samples) & 0.053 & 0.349 & 1.165 & 0.630 & 0.644 & 14.178 & 0.472\\
      uncoded 8 IWAE (DR, $\alpha=0.5$, 10 samples) & 0.052 & 0.312 & 1.004 &0.623 & 0.629 & 14.542 & 0.483\\
      uncoded 8 IWAE (DR, $\alpha=0.5$, 20 samples) & 0.056 & 0.353 & 1.041 & 0.600 & 0.614 & 14.231 & 0.463\\
      uncoded 8 IWAE (DR, $\alpha=0.5$, 30 samples) & 0.057 & 0.352 & 1.068 & 0.602 & 0.607 & 14.016 & 0.449\\
      coded 8/80	& 0.021 & 0.144	& 2.905	& 0.750	& 0.816	& 17.318 & 0.619\\
      coded 8/160	& 0.027	& 0.189	& 3.637	& 0.783	& 0.831	& 17.713 & 0.641\\
      coded 8/240	& 0.037	& 0.231	& 4.000	& 0.799	& 0.893	& 17.861 & 0.653\\
      \bottomrule 
    \end{tabular}
\end{table}

\section{Beyond Repetition Codes}\label{properties_eccs}

We have presented compelling proof-of-concept results that incorporating \acp{ecc}, like repetition codes, into \acp{dvae} can improve performance. We believe this opens a new path for designing latent probabilistic models with discrete latent variables. Although a detailed analysis of the joint design of \ac{ecc} and encoder-decoder networks is beyond the scope of this work, we will outline key properties that any \acp{ecc} must satisfy to be integrated within this framework.

\begin{itemize}
\item \textbf{Scalable hard encoding $\big(\m \rightarrow \c\big)$.} Our model requires hard encoding for generation once the model is trained. This process should have linear complexity in $M$.
\item \textbf{Scalable soft encoding $\big(p(\m)\rightarrow p(\c)\big)$.} Soft encoding is required during training for reparameterization. This process should also have linear complexity in $M$.
\item \textbf{Scalable soft decoding $\big(p(\c)\rightarrow p(\m)\big)$.} Our model employs \ac{siso} decoding during inference. This process should again be linearly complex in $M$.
\item \textbf{Differentiability.} Both encoding and decoding processes must be differentiable w.r.t. the inputs to enable gradient computation and backpropagation.
\end{itemize}

Since Shannon's landmark work \citep{shannon}, researchers have been developing feasible \ac{ecc} schemes that meet (or approach) the aforementioned properties for ever-increasing values of $M$ and $R$. This is driven by the high throughput demands in digital communications and limited power constraints in the case of wireless communications. In this regard, state-of-the-art \ac{ecc} schemes such as \ac{ldpc} codes \citep{ldpc} can be designed to meet the above criteria, leveraging their linear algebraic structure. Similarly, encoding and decoding algorithms for polar codes \citep{polarcodes} have complexity of $\mathcal{O}(D\log D)$, approaching this target. Additionally, in digital communications, soft decision decoding is known to provide savings of up to 2-3 dB of signal energy over hard decision decoding (binary input). For this reason, significant effort has been also made in the communications field to develop efficient and powerful \ac{siso} decoders, such as the sum-product algorithm (SPA) \citep{spa} for \ac{ldpc} codes.

\section{Hierarchical Coded-DVAE results}\label{polar_codes_results}

\begin{figure}[!b]
  \centering
  \includegraphics[width=0.25\linewidth]{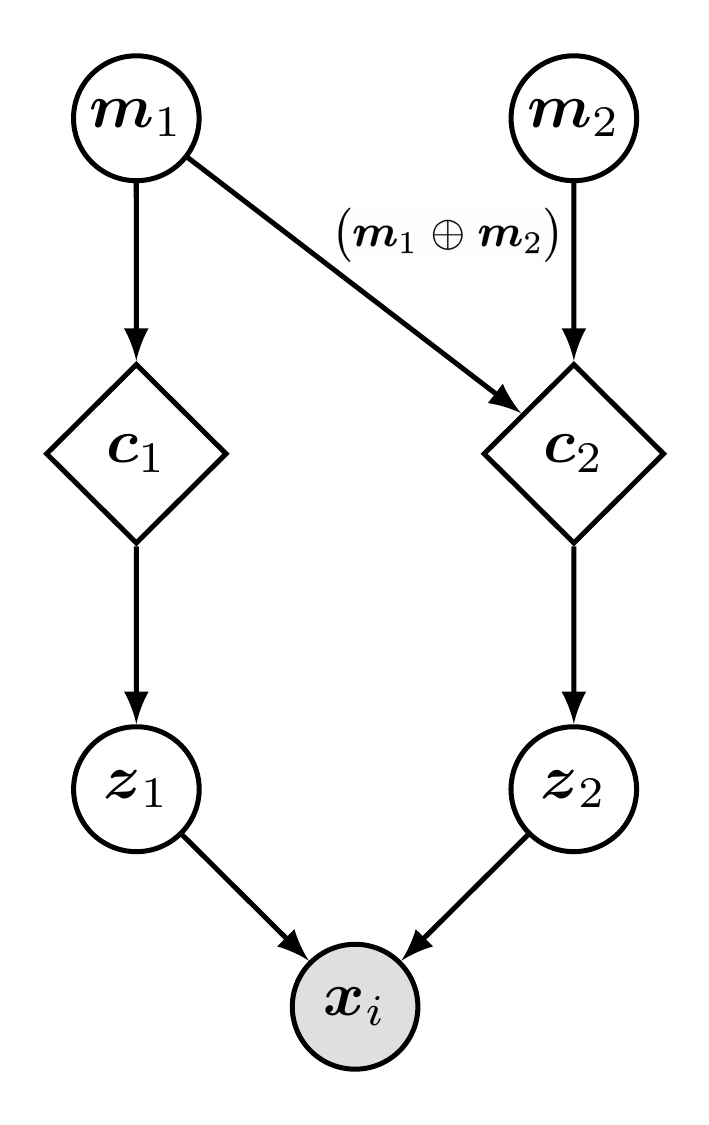}
  \caption{Graphical model of the hierarchical Coded VAE with two layers.}\label{fig:dag_hier}
\end{figure}

Inspired by polar codes \citep{polarcodes}, we present a hierarchical Coded-\ac{dvae} with two layers of latent bits, as illustrated in Fig. \ref{fig:dag_hier}. In this model, the latent bits $\m_1$ are encoded using a repetition code in the first layer, producing $\c_1$ and $\z_1$. Concurrently, the bits in the second layer, $\boldsymbol{m}_2$, are linearly combined with $\boldsymbol{m}_1$ following $\boldsymbol{m}_{1,2} = \boldsymbol{m}_1 \oplus \boldsymbol{m}_2$, considering a binary field or Galois field. The resulting vector is then encoded with another repetition code to produce $\boldsymbol{c}_2$,  which is subsequently modulated into $\boldsymbol{z}_2$. Finally, both $\boldsymbol{z}_1$ and $\boldsymbol{z}_2$ are concatenated and passed through the decoder network to generate $\boldsymbol{x}$. The model provides stronger protection for $\m_1$, as it appears in both branches of the generative model. Inference follows a similar approach to the one employed in the Coded-\ac{dvae}, incorporating the linear combination of $\m_1$ and $\m_2$ used in the second branch.

We adopt the same variational family as used in the standard Coded-VAE; however, in this case, we incorporate both hierarchical levels, leading to

\begin{equation}
q_{\e}(\boldsymbol{m}, \boldsymbol{z}|\boldsymbol{x}) = q_{\e}(\boldsymbol{m}_1|\boldsymbol{x})q_{\e}(\boldsymbol{m}_2|\boldsymbol{x})p(\boldsymbol{z}_1|\boldsymbol{c}_1)p(\boldsymbol{z}_2|\boldsymbol{c}_2),
\end{equation}

\noindent
where $q_{\e}(\boldsymbol{m}_1|\boldsymbol{x})$ is calculated following the same approach as in the Coded-\ac{dvae} with repetition codes, computing the all-are-zero and all-are-ones products of probabilities of the bits in $\boldsymbol{c}_1$ that are copies of the same message bit. The posterior $q_{\e}(\boldsymbol{m}_2|\boldsymbol{x})$ considering both the encoder's output and the inferred posterior distribution $q_{\e}(\boldsymbol{m}_1|\boldsymbol{x})$. Note that in this case, the decoder outputs the probabilities for both $\boldsymbol{c}_1$ and $\boldsymbol{c}_2$, with $\boldsymbol{c}_2$ being the encoded version of the linear combination $\boldsymbol{m}_{1,2} = \boldsymbol{m}_1 \oplus \boldsymbol{m}_2$. Consequently, we first obtain $q_{\e}(\boldsymbol{m}_{1,2}|\boldsymbol{x})$ following the same approach as in the Coded-\ac{dvae} with repetition codes, and determine $q_{\e}(\boldsymbol{m}_{2}|\boldsymbol{x})$ as

\begin{equation}
q_{\e}(\boldsymbol{m}_2|\boldsymbol{x}) = \prod_{j=1}^{M} \text{Ber} (p_j),
\end{equation}
\begin{equation}
p_j = q_{\e}(m_{1,2,j}=1|\boldsymbol{x})q_{\e}(m_{1,j}=0|\boldsymbol{x}) + q_{\e}(m_{1,2,j}=0|\boldsymbol{x})q_{\e}(m_{1,j}=1|\boldsymbol{x}).
\end{equation}

After obtaining $q_{\e}(\boldsymbol{m}_1|\boldsymbol{x})$ and $q_{\e}(\boldsymbol{m}_2|\boldsymbol{x})$, we recalculate the posterior bit probabilities for the linear combination $q'_{\e}(\boldsymbol{m}_{1,2}|\boldsymbol{x})$ as
\begin{equation}
q'_{\e}(\boldsymbol{m}_{1,2}|\boldsymbol{x}) = \prod_{j=1}^M \text{Ber}(q_j),
\end{equation}
\begin{equation}
q_j = q_{\e}(m_{1,j}=1|\boldsymbol{x})q_{\e}(m_{2,j}=0|\boldsymbol{x}) + q_{\e}(m_{1,j}=0|\boldsymbol{x})q_{\e}(m_{2,j}=1|\boldsymbol{x}).
\end{equation}

Next, we apply the soft encoding approach to incorporate the repetition code structure at both levels of the hierarchy. The posterior probabilities $q_{\e}(\boldsymbol{m}_1|\boldsymbol{x})$ are repeated to obtain $q_{\e}(\boldsymbol{c}_1|\boldsymbol{x})$, and the posterior probabilities $q_{\e}(\boldsymbol{m}_{1,2}|\boldsymbol{x})$ are repeated to produce $q_{\e}(\boldsymbol{c}_2|\boldsymbol{x})$. Utilizing the reparameterization trick from Eq. \eqref{CDF_dvae}, we sample $\boldsymbol{z}_1$ and $\boldsymbol{z}_2$, concatenate them to form $\boldsymbol{z}$, and pass this through the decoder to generate $p_{\boldsymbol{\theta}}(\boldsymbol{x}|\boldsymbol{z})$. The model is trained by maximizing the \ac{elbo}, given by

\begin{equation}
    \text{ELBO} = \E_{q_{\e}(\m,\z|\x)}\log p_{\t}(\x|\z) - \mathcal{D}_{KL}\big(q_{\e}(\m_1|\x)||p(\m_1)\big) - \mathcal{D}_{KL}\big(q_{\e}(\m_2|\x)||p(\m_2)\big),
\end{equation}

\noindent
where both $p(\m_1)$ and $p(\m_2)$ are assumed to be independent Bernoulli distributions with bit probabilities of $0.5$, consistent with the other scenarios.

\begin{figure}[!t]
     \centering
     \begin{subfigure}[b]{0.45\textwidth}
        \centering
         \includegraphics[scale=0.65]{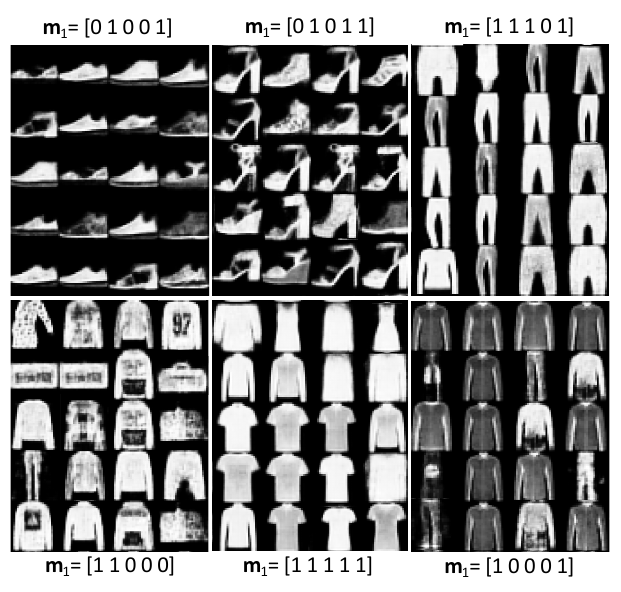}
         \caption{}
         \label{fig:hier_5_50}
     \end{subfigure}
     \begin{subfigure}[b]{0.45\textwidth}
        \centering
         \includegraphics[scale=0.652]{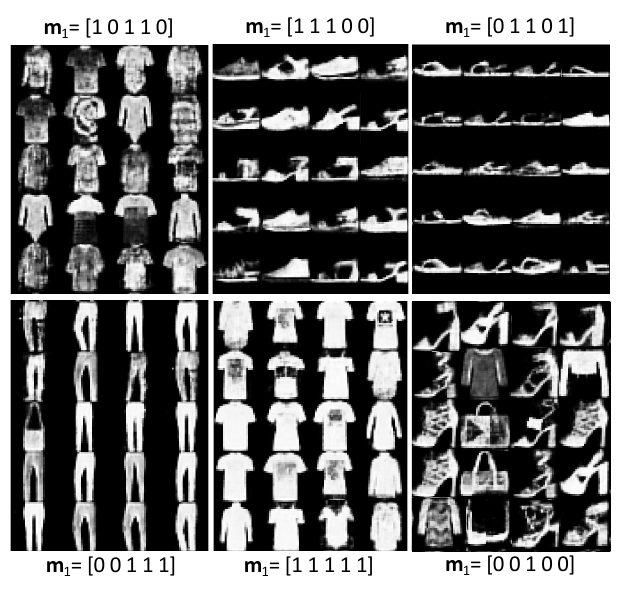}
         \caption{}
         \label{fig:hier_5_100}
     \end{subfigure}
    \caption{Examples of generated images using the hierarchical Coded-DVAE with (a) a 5/50 repetition code in each branch, and (b) a 5/100 repetition code in each branch. In all the examples provided, $\boldsymbol{m}_1$ was fixed while $\boldsymbol{m}_2$ was randomly sampled.}
    \label{fig:hier_generation}
\end{figure}

We obtained results on the FMNIST dataset using a model with 5 information bits per branch and repetition rates of $R=1/10$ and $R=1/20$. In this case, we applied the same code rate to both branches, although varying code rates could be used to control the level of protection at each hierarchy level. Tables \ref{tab:hier_general_metrics} and \ref{tab:hier_metrics} present the metrics obtained for the different configurations. Specifically, Table \ref{tab:hier_general_metrics} shows the overall metrics obtained with this structure, and Table \ref{tab:hier_metrics} compares the error metrics across the two hierarchy levels. As expected,  $\mathbf{m}_2$ shows poorer error metrics compared to $\mathbf{m}_1$, since the model provides more redundancy to $\mathbf{m}_1$ incorporating it in both branches. Although the overall metrics and generation quality are similar to those of the Coded-\ac{dvae} with 10 information bits (see tables in Figs. \ref{fig:rec_fmnist} and \ref{fig:generation_fmnist}), the introduced hierarchy results in a more interpretable latent space. In this setup, $\mathbf{m}_1$ captures global features (such as clothing types in the FMNIST dataset), while $\mathbf{m}_2$ controls individual features, as we can observe in Fig. \ref{fig:hier_generation}, where we show examples of the model’s generative outputs for fixed $\mathbf{m}_1$ and random samples of $\mathbf{m}_2$.

\begin{table}[!h]

    \centering
    \caption{Comparison of the obtained metrics for the Coded-DVAE with polar codes with different configurations, which we refer to as `hierarchical Coded-DVAE'.}\label{tab:hier_general_metrics}
    \begin{tabular}{cccccc}
      \toprule 
      \bfseries Model & \bfseries BER & \bfseries WER & \bfseries Acc & \bfseries Conf. Acc & \bfseries PSNR\\
      \midrule 
      hier. 5/50 & 0.099 & 0.400 & 0.753 & 0.800 & 17.130 \\
      hier. 5/100 & 0.050 & 0.330 & 0.784 & 0.870 & 17.513 \\
      \bottomrule 
    \end{tabular}
\end{table}

\begin{table}[!h]

    \centering
    \caption{Comparison of the obtained error metrics in the different hierarchy levels.}\label{tab:hier_metrics}
    \begin{tabular}{ccccc}
      \toprule 
      \bfseries Model & \bfseries BER $\m_1$ & \bfseries WER $\m_1$ & \bfseries BER $\m_2$ & \bfseries WER $\m_2$ \\
      \midrule 
      hier. 5/50 & 0.079 & 0.259 & 0.119 & 0.362  \\
      hier. 5/100 & 0.026 & 0.110 & 0.075 & 0.287 \\
      \bottomrule 
    \end{tabular}
\end{table}

\section{Ablation study}\label{ablation}

In this section, we conduct ablation studies on the hyperparameter $\beta$ of the model, responsible for regulating the decay of exponentials in the smoothing transformation, as well as on the number of trainable parameters in the models.

\subsection{Ablation study on the hyperparameter $\beta$}

Across all experiments, we have consistently configured the hyperparameter $\beta$, which controls the decay of exponentials in the smoothing transformation, to a value of 15. To illustrate its impact on the overall performance of the model, we conducted an ablation study on the value of this hyperparameter for both uncoded and coded cases.

The smoothing distribution employed for the reparameterization trick consists of two overlapping exponentials. The hyperparameter $\beta$ functions as a temperature term, regulating the decay of the distributions and, consequently, influencing the degree of overlapping. A lower $\beta$ value results in more overlapped tails, while a higher value leads to less overlapped distributions. A priori, we would like these distributions to be separated, allowing us to retrieve the true value of the bit and effectively use the latent structure of the model.

\subsubsection{Coded model}

We first evaluate the influence of the parameter $\beta$ in coded models. We take as a reference the coded model with 8 information bits and a rate $R=1/30$, and train it using $\beta = 5, 10, 15, 20$. We assess the performance of the model in reconstruction and

\begin{figure}[!hb]
  \centering
  \includegraphics[width=1\linewidth]{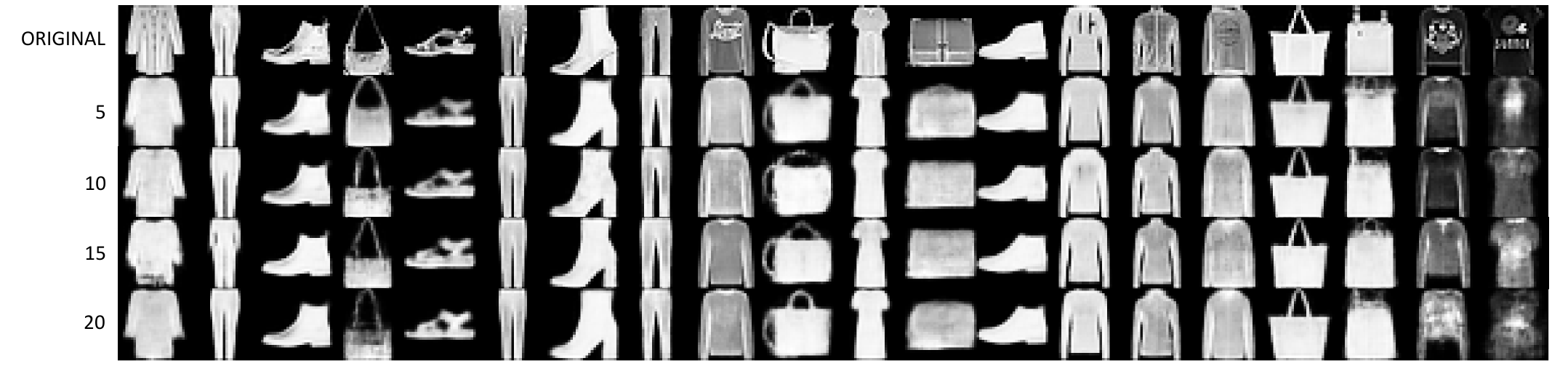}
  \caption{Example of reconstructed images obtained with different values of $\beta$ using the coded model with an 8/240 code. }\label{fig:rec_ablation_beta_coded}
\end{figure}

 generation tasks. We observe the model is fairly robust, achieving similar performance across configurations in most metrics. 

In Fig. \ref{fig:rec_ablation_beta_coded} we show examples of reconstructed images using the different configurations to assess reconstruction through visual examination, and Table \ref{tab:metrics_ablation_beta_coded} contains the associated reconstruction metrics. All the configurations achieve similar performances, although the models trained with $\beta = 10$ and $\beta = 15$ seem to be the best configurations for this scenario. Larger values may result in unstable training and inferior performance.

\begin{table*}[!t]

\centering
    \caption{Evaluation of reconstruction performance in FMNIST with different values of $\beta$ using the coded model with an 8/240 code.} \label{tab:metrics_ablation_beta_coded}
    \begin{tabular}{cccccccc}
      \toprule 
      \bfseries Model & \begin{tabular}{@{}c@{}}\bfseries{PSNR}\\ \bfseries{(train)}\end{tabular}& \begin{tabular}{@{}c@{}}\bfseries{Acc}\\ \bfseries{(train)}\end{tabular} & \begin{tabular}{@{}c@{}}\bfseries{Conf. Acc.}\\ \bfseries{(train)}\end{tabular} & \begin{tabular}{@{}c@{}}\bfseries{PSNR}\\ \bfseries{(test)}\end{tabular} & \begin{tabular}{@{}c@{}}\bfseries{Acc}\\ \bfseries{(test)}\end{tabular}  & \begin{tabular}{@{}c@{}}\bfseries{Conf. Acc.}\\ \bfseries{(test)}\end{tabular} & \bfseries Entropy\\
      \midrule 
      5  & 18.344 & 0.791 & 0.895 & 17.614 & 0.766 & 0.849 & 4.025\\
      10 & 19.106  & 0.822 & 0.904 & 17.737 & 0.793 & 0.872 & 4.023\\
      15 & 19.345 & 0.831  & 0.921 & 17.861 & 0.799 & 0.893 & 4.000\\
      20 & 18.797 & 0.809 & 0.887 & 17.837 & 0.787 & 0.877 & 3.810\\
      \bottomrule 
    \end{tabular}
\end{table*}

\begin{figure}[!t]
  \centering
  \includegraphics[width=0.8\linewidth]{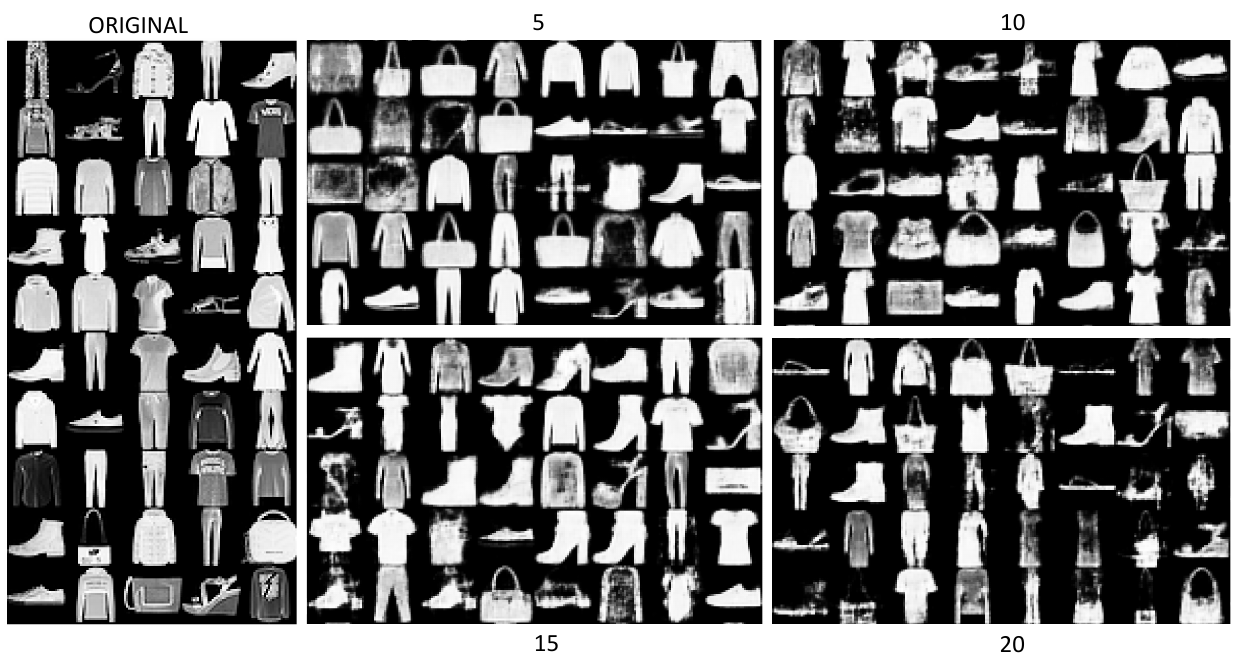}
  \caption{Example of randomly generated, uncurated images with different values of $\beta$ using the coded model with an 8/240 code. }\label{fig:gen_ablation_beta_coded}
\end{figure}

\begin{table}[!t]
\parbox{.45\linewidth}{

\centering
\caption{Evaluation of the \ac{ber} and  \ac{wer} in FMNIST with different values of $\beta$ using the coded model with an 8/240 code.}\label{tab:error_fmnist_ablation_beta_coded}
\begin{tabular}{ccc}
      \toprule 
      \bfseries Beta & \bfseries BER & \bfseries WER\\
      \midrule 
      5 & 0.150 & 0.726\\
      10 & 0.080 & 0.480\\
      15 & 0.037 & 0.231\\
      20 & 0.065 & 0.399\\
      \bottomrule 
\end{tabular}
}
\hfill
\parbox{.45\linewidth}{

\centering
\caption{Evaluation of the log-likelihood (LL) in FMNIST with different values of $\beta$ using the coded model with an 8/240 code.} \label{tab:loglik_ablation_beta_coded}
\begin{tabular}{ccc}
      \toprule 
      \bfseries Beta & \bfseries LL (train) & \bfseries LL (test) \\
      \midrule 
      5  & -228.448 & -234.629 \\
      10 & -229.379 & -237.495 \\
      15 & -231.679 & -238.459 \\
      20 & -229.627 & -235.927 \\
      \bottomrule 
\end{tabular}
}
\end{table}

Next, we evaluate the model in the image generation task. Fig. \ref{fig:gen_ablation_beta_coded} contains examples of randomly generated images using the different configurations. Table \ref{tab:error_fmnist_ablation_beta_coded} reports the obtained \ac{ber} and \ac{wer}, and Table \ref{tab:loglik_ablation_beta_coded} the estimated log-likelihood of the different values of $\beta$. The model trained with $\beta=15$ stands out in terms of error metrics, although achieves similar log-likelihood values as the model trained with $\beta=10$. Again, these two configurations appear to be the most suitable in this scenario.

\subsubsection{Uncoded model}

We first evaluate the influence of the parameter $\beta$ in uncoded models. We take as a reference the coded model with 8 information bits and train it using $\beta = 5, 10, 15, 20$. We assess the performance of the model in reconstruction and generation tasks. We observe that the uncoded model is also robust, achieving similar performance across configurations.

\begin{figure}[!h]
  \centering
  \includegraphics[width=0.95\linewidth]{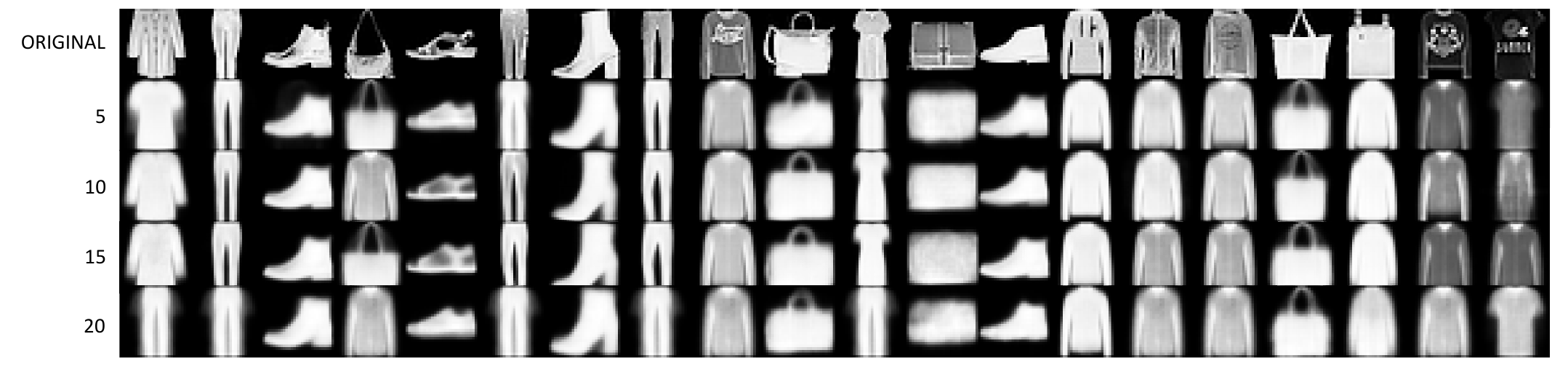}
  \caption{Example of reconstructed images obtained with different values of $\beta$ using an uncoded model 8 information bits. }\label{fig:rec_ablation_beta_uncoded}
\end{figure}

In Fig. \ref{fig:rec_ablation_beta_uncoded} we show examples of reconstructed images using the different configurations to assess reconstruction through visual examination, and Table \ref{tab:metrics_ablation_beta_uncoded} contains the associated reconstruction metrics. All the configurations achieve similar performances, although the models trained with $\beta = 10$ and $\beta = 15$ seem to be the best configurations for this scenario. Larger values may result in unstable training and inferior performance, as we can clearly observe in this case.

\begin{table*}[!htb]

\centering
    \caption{Evaluation of reconstruction performance in FMNIST with different values of $\beta$ using an uncoded model 8 information bits.} \label{tab:metrics_ablation_beta_uncoded}
    \begin{tabular}{cccccccc}
      \toprule 
      \bfseries Model & \begin{tabular}{@{}c@{}}\bfseries{PSNR}\\ \bfseries{(train)}\end{tabular}& \begin{tabular}{@{}c@{}}\bfseries{Acc}\\ \bfseries{(train)}\end{tabular} & \begin{tabular}{@{}c@{}}\bfseries{Conf. Acc.}\\ \bfseries{(train)}\end{tabular} & \begin{tabular}{@{}c@{}}\bfseries{PSNR}\\ \bfseries{(test)}\end{tabular} & \begin{tabular}{@{}c@{}}\bfseries{Acc}\\ \bfseries{(test)}\end{tabular}  & \begin{tabular}{@{}c@{}}\bfseries{Conf. Acc.}\\ \bfseries{(test)}\end{tabular} & \bfseries Entropy\\
      \midrule 
      5  & 14.239 & 0.503 & 0.501 & 14.237 & 0.503 & 0.491 & 0.231\\
      10 & 15.624  & 0.606 & 0.603 & 15.571 & 0.598 & 0.598 & 0.357\\
      15  & 15.644  & 0.601  & 0.602 & 15.598 & 0.594 & 0.595 & 0.467\\
      20 & 13.717 & 0.464 & 0.466 & 13.743 & 0.460 & 0.462 & 0.383\\
      \bottomrule 
    \end{tabular}
\end{table*}

\begin{figure}[!htb]
  \centering
  \includegraphics[width=0.8\linewidth]{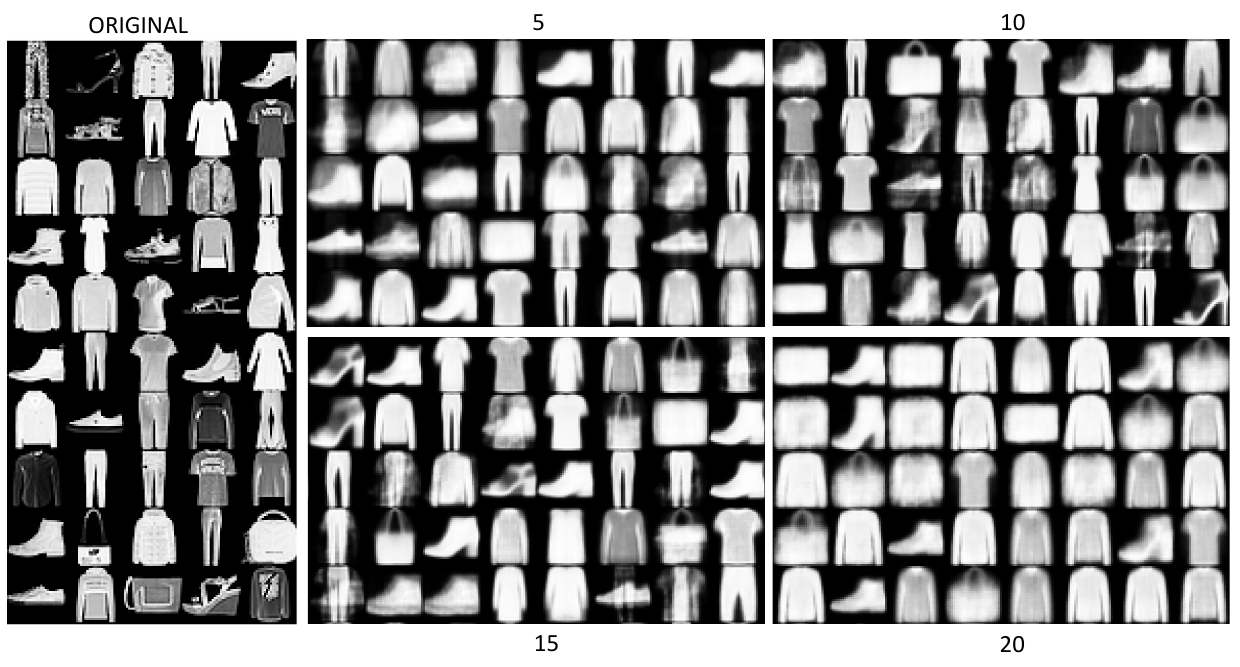}
  \caption{Example of randomly generated, uncurated images with different values of $\beta$ using an uncoded model 8 information bits.}\label{fig:gen_ablation_beta_uncoded}
\end{figure}

\begin{table}[!ht]
\parbox{.45\linewidth}{

\centering
\caption{Evaluation of the \ac{ber} and  \ac{wer} in FMNIST with different values of $\beta$ using an uncoded model 8 information bits. }\label{tab:error_fmnist_ablation_beta_uncoded}
\begin{tabular}{ccc}
      \toprule 
      \bfseries Beta & \bfseries BER & \bfseries WER\\
      \midrule 
      5 & 0.203 & 0.852\\
      10 & 0.086 & 0.384\\
      15 & 0.089 & 0.384\\
      20 & 0.278 & 0.939\\
      \bottomrule 
\end{tabular}
}
\hfill
\parbox{.45\linewidth}{

\centering
\caption{Evaluation of the log-likelihood (LL) in FMNIST with different values of $\beta$ using an uncoded model 8 information bits. } \label{tab:loglik_ablation_beta_uncoded}
\begin{tabular}{ccc}
      \toprule 
      \bfseries Model & \bfseries LL (train) & \bfseries LL (test) \\
      \midrule 
      5  & -256.431 & -257.983 \\
      10 & -247.507 &  -249.460 \\
      15 & -247.964 &  -249.880 \\
      20 & -272.460 &  -273.554 \\
      \bottomrule 
\end{tabular}
}
\end{table}

Next, we evaluate the model in the image generation task. Fig. \ref{fig:gen_ablation_beta_uncoded} contains examples of randomly generated images using the different configurations. Table \ref{tab:error_fmnist_ablation_beta_uncoded} reports the obtained \ac{ber} and \ac{wer}, and Table \ref{tab:loglik_ablation_beta_uncoded} the estimated log-likelihood of the different values of $\beta$. The models trained with $\beta=15$ and $\beta=10$ clearly outperform the other two in this task, generating more diverse and detailed images, and obtaining better error metrics and log-likelihood values.

\subsection{Ablation study on the number of trainable parameters}\label{ablation_parameters}

A consistent architecture was employed across all experiments, which is detailed in Section \ref{architecture}. However, since the introduction of the code alters the dimensionality of the latent space, it is necessary to adjust the encoder's output and the decoder's input. This results in an augmentation of the trainable parameters in the coded cases compared to their uncoded counterparts.

Given that a higher number of parameters usually results in better performance, we conducted an ablation study on the model's trainable parameters to confirm that the improved performance introduced by the coded models is not due to this factor. We adjusted the hidden dimensions of the encoder and decoder architectures to ensure both configurations (coded and uncoded) have roughly the same number of trainable parameters. We have conducted the ablation study using the uncoded model with 8 bits and the coded 8/240 model trained on FMNIST. 

We adjusted the encoder's last hidden dimension and the decoder's first hidden dimension to equalize the parameter count between the uncoded and coded models. This adjustment was straightforward since the last layers of the encoder and the first layers of the decoder are feed-forward layers. We kept the latent dimension of the model unchanged, ensuring that the modification solely pertained to the neural network architecture. 

\begin{table*}[!h]

\centering
    \caption{Parameter count.} \label{tab:parameter_count}
    \begin{tabular}{ccc}
      \toprule 
      \bfseries Model & \bfseries \# encoder parameters & \bfseries \# decoder parameters \\
      \midrule 
      uncoded 8  & 6,592,008  & 19,341,185\\
      uncoded 8 adjusted & 6,717,538 & 19,581,035\\
      coded 8/240  & 6,711,024 &  19,578,753 \\
      coded 8/240 adjusted  & 6,583,174 &  19,332,871 \\

      \bottomrule 
    \end{tabular}
\end{table*}

\subsubsection{Evaluation}

This section provides an empirical evaluation of the models trained with the adjusted parameter count, demonstrating that the enhanced performance observed in coded models does not result from an augmented number of trainable parameters. We found that the performance of the original and adjusted models is very similar, meaning that the conclusions drawn in the main text hold even in this scenario.

We first evaluate the reconstruction performance, measuring the \ac{psnr} and reconstruction accuracy in both train and test sets, which are included in Table \ref{tab:metrics_ablation_params}. Then, we assess generation measuring the  \ac{ber} and \ac{wer}, reported in Table \ref{tab:error_fmnist_ablation_params}. Finally, we compute the log-likelihood for train and test sets, shown in Table \ref{tab:loglik_ablation_params}.

In terms of reconstruction, both the original and adjusted models exhibit very similar performance, observed in both reconstruction quality and accuracy. However, the adjusted models show slightly inferior results than the original ones. For the coded model, this might be attributed to reduced flexibility when decreasing the number of parameters. As for the uncoded model, the increased complexity while maintaining the same low-dimensional latent vector may not provide enough expressiveness to leverage the added flexibility in the architecture, potentially causing the observed decrease in performance. We observe the same behavior when we evaluate the \ac{ber} and \ac{wer}.

Analyzing the results, especially the log-likelihood values shown in Table \ref{tab:loglik_ablation_params}, we can argue that increasing the flexibility in the architecture does not necessarily lead to improved performance in this scenario. The coded models exhibit similar performance with both the original and adjusted parameter counts, consistently outperforming the uncoded models. These results indicate that the performance enhancement is attributed to the introduction of \acp{ecc} in the latent space, rather than differences in the architecture required to handle the introduced redundancy.

\begin{figure}[!h]
  \centering
  \includegraphics[width=0.9\linewidth]{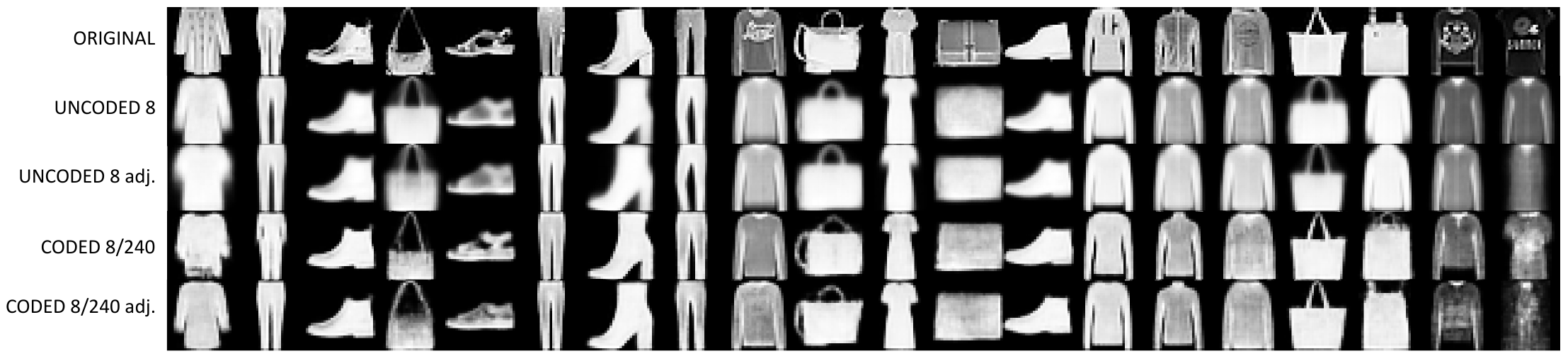}
  \caption{Example of reconstructed images obtained with different configurations. }\label{fig:rec_ablation_params}
\end{figure}
\begin{table*}[!h]

\centering
    \caption{Evaluation of reconstruction performance in FMNIST with the adjusted parameter count.} \label{tab:metrics_ablation_params}
    \begin{tabular}{cccccccc}
      \toprule 
      \bfseries Model & \begin{tabular}{@{}c@{}}\bfseries{PSNR}\\ \bfseries{(train)}\end{tabular}& \begin{tabular}{@{}c@{}}\bfseries{Acc}\\ \bfseries{(train)}\end{tabular} & \begin{tabular}{@{}c@{}}\bfseries{Conf. Acc.}\\ \bfseries{(train)}\end{tabular} & \begin{tabular}{@{}c@{}}\bfseries{PSNR}\\ \bfseries{(test)}\end{tabular} & \begin{tabular}{@{}c@{}}\bfseries{Acc}\\ \bfseries{(test)}\end{tabular}  & \begin{tabular}{@{}c@{}}\bfseries{Conf. Acc.}\\ \bfseries{(test)}\end{tabular} & \bfseries Entropy\\
      \midrule 
      uncoded 8  & 15.644  & 0.601  & 0.602 & 15.598 & 0.594 & 0.595 & 0.659\\
      uncoded 8 adj. & 15.530  & 0.586  & 0.586 & 15.491 & 0.581 & 0.580 & 0.449\\
      coded 8/240 & 19.345 & 0.831  & 0.921 & 17.861 & 0.799 & 0.893 & 4.609\\
      coded 8/240 adj. & 19.383 &  0.828 & 0.883 & 17.771 & 0.792 & 0.828 & 3.952\\
      \bottomrule 
    \end{tabular}
\end{table*}

\begin{figure}[!h]
  \centering
  \includegraphics[width=0.8\linewidth]{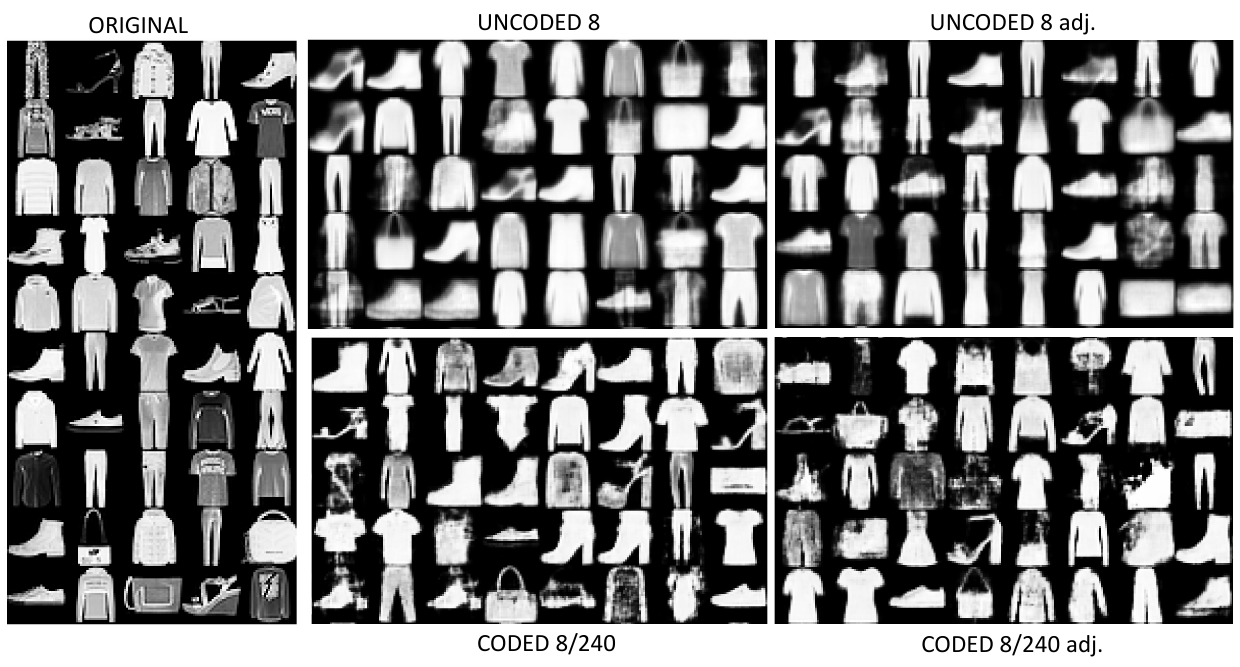}
  \caption{Example of randomly generated, uncurated images using different model configurations. }\label{fig:gen_ablation_params}
\end{figure}

\begin{table}[!h]
\parbox{.47\linewidth}{

\centering
\caption{Evaluation of the \ac{ber} and  \ac{wer} in FMNIST with the adjusted parameter count.}\label{tab:error_fmnist_ablation_params}
\begin{tabular}{ccc}
      \toprule 
      \bfseries Model & \bfseries BER & \bfseries WER\\
      \midrule 
      uncoded 8 & 0.089 & 0.384\\
      uncoded 8 adj. & 0.125 & 0.561\\
      coded 8/240 & 0.037 & 0.231\\
      coded 8/240 adj. & 0.064 & 0.399\\
      \bottomrule 
\end{tabular}
}
\hfill
\parbox{.47\linewidth}{

\centering
\caption{Evaluation of the log-likelihood (LL) in FMNIST with the adjusted parameter count.} \label{tab:loglik_ablation_params}
\begin{tabular}{ccc}
      \toprule 
      \bfseries Model & \bfseries LL (train) & \bfseries LL (test) \\
      \midrule 
      uncoded 8  & -247.964 &  -249.880 \\
      uncoded 8 adj. & -250.543 &  -252.408 \\
      coded 8/240 & -231.679 & -238.459 \\
      coded 8/240 adj. & -229.283 & -238.302 \\
      \bottomrule 
\end{tabular}
}
\end{table}

\newpage
\section{Evaluating log-likelihood using the soft-encoding model}\label{augmented_model}

The reparameterization trick introduced in \eqref{CDF_dvae} requires that bits are independent, a condition not met in $\c$ once the code's structure is introduced. To address this issue, during training, we employ a \emph{soft encoding strategy}. We assume the bits in $\c$ are independent and equally distributed according to $q_{\e}(\m|\x)$.  Therefore, instead of directly repeating sampled bits in $\m$ to obtain $\c$ following $\c = \m^T\mathbf{G}$, we repeat the posterior probabilities for the copies of the same bit and sample $\z$ using the reparameterization trick in \eqref{CDF_dvae}. Remark that, despite the soft encoding assumption during training, the generative results in Fig. \ref{fig:generation_fmnist}, \ref{fig:cifar_imagenet_gen}, \ref{fig:generation_fmnist_supp}, \ref{fig:generation_mnist}, \ref{fig:gen_cifar_sup}, \ref{fig:gen_cifar_sup}, and \ref{fig:gen_imagenet_supp} are obtained through \emph{hard encoding}. Namely, we sampled an information word $\m$ and obtained $\c$ by repeating its bits. 

While the hard-encoding images are visually appealing, to evaluate the Coded-\ac{dvae} LL for a given image $\x$, we have to leverage the soft-encoding model since it is unrealistic for a sample from the soft-encoding model to have equally repeated bits. In the soft-coded model, we sample bit probabilities from a prior distribution that we model through the product of $M$ independent Beta distributions, and we use a proposal distribution model similar to the Vamp-prior in \cite{VampPrior}. Namely, a mixture model with components given by $q(\m|\x)$ for different training points.

In the main text and Section \ref{ablation}, we report the \ac{ll} values for different model configurations trained on the FMNIST dataset. Table \ref{tab:loglik_mnist} presents the results obtained for the MNIST dataset. Due to their simplicity, these datasets do not require high-dimensional latent spaces, and competitive results can be achieved with just 8 or 10 information bits. However, for more complex datasets like CIFAR10 or Tiny ImageNet, high-dimensional latent spaces are necessary to capture spatial information, colors, and textures. For these high-dimensional datasets, we were unable to obtain valid log-likelihood estimates because the number of samples needed for importance sampling to converge was too large. However, given the difference in the level of detail in the reconstructed and generated images between coded and uncoded models (see Sections \ref{cifar_results_supp} and \ref{tinyimagenet_results_supp}), coded models are expected to better approximate the true marginal likelihood of the data.

\section{Connection to previous works on VAEs as source coding methods}\label{source_coding}

Effective learning of low-dimensional discrete latent representations from complex data is a technically challenging task. In this work, we propose a novel method to improve inference in discrete \acp{vae} within a fully probabilistic framework, introducing a new perspective on the inference problem. While previous studies have analyzed \acp{vae} using \ac{rd} theory \citep{chen2022variational, townsend2019practical, vqvae}, our approach stems from a different yet complementary perspective that remains compatible with these works.

In the literature, \acp{vae} are often interpreted as lossy compression models (e.g., \acp{vae} as source coding methods), but our contribution is best understood from a generative perspective. We conceptualize inference via $q(\boldsymbol{m}|\boldsymbol{x})$ as a decoding process, where the goal is to recover the discrete latent variable from the observed data. We sample a latent vector $\boldsymbol{m}$, generate a data point $\boldsymbol{x}$ (e.g., an image), and aim to minimize the error rate in recovering $\boldsymbol{m}$ from $\boldsymbol{x}$. Achieving this requires the variational approximation $q(\boldsymbol{m}|\boldsymbol{x})$ to closely match the true posterior $p(\boldsymbol{m}|\boldsymbol{x})$.We demonstrate that incorporating redundancy into the generative process reduces errors in estimating $\m$, leading to a more accurate latent posterior approximation. This improvement directly enhances our model’s overall performance, as confirmed by our experimental results. Additionally, error rates and variational gaps are related through bounds derived from mismatch hypothesis testing. These bounds demonstrate that minimizing the variational gap by maximizing the \ac{elbo} effectively tightens an upper bound on the error rate \citep{error_bounds_f_divergence, error_bounds_refined}.

In \ac{rd} theory, the primary focus is on compression within the latent space, typically analyzed from an encoder/decoder and reconstruction (distortion) perspective. RD theory establishes theoretical limits on achievable compression rates and describes how practical models may diverge from these limits. Using \ac{rd} practical compression methods, \citet{townsend2019practical} demonstrates how asymmetric numeral systems (ANS) can be integrated with a \ac{vae} to improve its compression rate by jointly encoding sequences of data points, bringing performance closer to \ac{rd} theoretical limits. Similarly, \citet{chen2022variational} shows that a complex prior distribution in a \ac{vae} using an autoregressive invertible flow narrows the gap between the approximate and the true posterior distribution, thereby enhancing the overall performance of the \ac{vae}. We note that even using an independent prior, the hierarchical code structure outlined in Section \ref{polar_codes_results} naturally decouples information across different latent spaces at various conceptual levels. Specifically, the most relevant information (class label) is captured by the most protected latent space, while the other space captures fine-grained features. This effect is not straightforward to enforce through direct design of a more complex prior.

Our method complements all these efforts by introducing redundancy in the generative pathway to enhance variational inference, leading to more accurate latent posterior approximations. Notably, even when using a complex prior distribution, \acp{ecc} can still be leveraged to enhance inference and overall model performance. In this work, we demonstrate that our approach yields more robust models even with a simple, fixed independent prior, as evidenced by improved \ac{ll}, generation quality, and reconstruction metrics. Moreover, in Section \ref{polar_codes_results}, we show that integrating a hierarchical \ac{ecc} with the same independent prior leads to even greater performance gains.

It is also important to discuss the \ac{vqvae} \citep{vqvae}, one of the most relevant works related to our approach, which is well-known for effectively learning compressed discrete representations of complex data. We believe that highlighting key differences between \acp{vqvae} and our method can provide useful insights. A key distinction lies in the structure and dimensionality of the latent space. In image modeling, \acp{vqvae} typically employ a latent matrix where indices correspond to codewords in a codebook. This design allows different codewords to capture specific patches of the original image, improving reconstruction and generation (the latter through an autoregressive prior on the latent representations).  However, representing each data point as a grid of embeddings complicates interpretability. In contrast, our method encodes the entire image into the latent space, rather than partitioning it into patches. Another major difference is the nature of the encoder. \acp{vqvae} rely on a non-probabilistic encoder that maps inputs to the nearest latent code using a distance-based metric. This deterministic mapping limits the model’s ability to capture and quantify uncertainty in the latent space. Our approach, on the other hand, operates within a fully probabilistic framework, allowing for uncertainty quantification in the learned latent representations. Additionally, our method is fully differentiable, enabling seamless gradient computation and backpropagation for end-to-end training. Our approach also allows us to leverage the reparameterization trick introduced in DVAE++ \citep{dvae++}, which we found to be more stable than continuous relaxations like the Gumbel-Softmax used in \acp{vqvae} with stochastic quantization \citep{williams2020hierarchical}.

\section{Variational inference at codeword level}\label{coded_word}


Here, we present an alternative variational family that is valid for any ECC, including random codes. We assume that we have a deterministic mapping of the form $\c=\mathcal{C}(\m)$. We assume a variational family of the form

\begin{equation}\label{VI1}
    q_{\e}(\c,\z|\x) = q_{\e}(\c|\x)p(\z|\c),
\end{equation}
\begin{equation}\label{VI2}
    q_{\e}(\c|\x) \propto p(\c)q_{\e}^{u}(\c|\x),
\end{equation}
\begin{equation}\label{VI3}
    q_{\e}^{u}(\c|\x) = \prod_{j=1}^{M/R} \text{Ber}(g_{j,\e}(\x)),
\end{equation}

\noindent where $g_{\e}(\x)$ represents the output of the encoder with a parameter set denoted as $\e$. Note that $q_{\e}^{u}(\c|\x)$ corresponds to the \textit{uncoded} posterior, which we subsequently constrain using the prior distribution $p(\c)$ over the code words to obtain the \textit{coded} posterior $q_{\e}(\c|\x)$. Then, the \textit{coded} posterior distribution can be defined as a categorical distribution over the set of codewords $\mathcal{C}(\m)$, which is given by

\begin{equation}
    \begin{gathered}
        q_{\e}(\c|\x) = \text{Cat}\Big(\Big[\frac{1}{W}q_{\e}^{u}(\c_1|\x),\ldots, \frac{1}{W}q_{\e}^{u}(\c_{2^M}|\x)\Big]\Big)\\
        = \frac{1}{W}\prod_{\c_i \in \mathcal{C}(\m)}\prod_{j=1}^{M/R}g_{j,\e}(\x)^{c_{i,j}}(1-g_{j,\e}(\x))^{(1-c_{i,j})},
    \end{gathered}
\end{equation}

\noindent
where $W = \sum_{\c_i \in \mathcal{C}(\m)}q_{\e}^{u}(\c_i|\x)$ is a constant for normalization.

Inference is done by maximizing the \ac{elbo}, which can be expressed as
\begin{equation}
    \begin{gathered}
    \text{ELBO} = \int  q_{\e}(\c,\z|\x) \log\left(\frac{p_{\t}(\x,\z,\c)}{q_{\e}(\c,\z|\x)}\right)d\c d\z \\
    = \E_{q_{\e}(\c,\z|\x)}\log\left(\frac{p_{\t}(\x|\z)p(\z|\c)p(\c)}{q_{\e}(\c|\x)p(\z|\c)}\right) \\
    = \E_{q_{\e}(\c,\z|\x)}\log p_{\t}(\x|\z) - \mathcal{D}_{KL}\big(q_{\e}(\c|\x)||p(\c)\big).
    \end{gathered}\label{elbo_cvae_word}
\end{equation}


In this case, due to the inability to compute the \ac{kl} Divergence in closed form, we approximate it via Monte Carlo, sampling from the categorical distribution $q_{\e}(\c|\x)$. The reconstruction term also needs to be approximated via Monte Carlo. Since the use of channel coding introduces structural dependencies among the components of the vectors $\c$, we can no longer assume their independence. Consequently, the formulation of the smoothing transformation as independent mixtures introduced in the \ac{dvae} is no longer applicable. Hence, this approach involves sampling $\c'$ from the categorical distribution $q_{\e}(\c|\x)$ and subsequently applying the transformation over the sampled word. Thus, we obtain a smooth transformation for each sample $\c'$ using the inverse \acp{cdf} of $p(z_j|c_j=0)$ and $p(z_j|c_j=1)$, which are given by

\begin{equation}
F^{-1}_{p(z_j|\c'_j=0)}(\rho) = -\frac{1}{\beta}\log \big(1-\rho(1-e^{-\beta})\big), 
\end{equation}

\begin{equation}
F^{-1}_{p(z_j|\c'_j=1)}(\rho) = \frac{1}{\beta}\log \big(\rho(1-e^{-\beta})+e^{-\beta}\big)+1.
\end{equation}

These are differentiable functions that convert samples $\rho$ from a uniform distribution $\mathcal{U}(0,1)$ into a sample from $q_{\eta}(\z,\c=\c'|\x)$ following

\begin{equation}
    \begin{gathered}
    q_{\eta}(\z,\c=\c'|\x) = \\=\prod_{j=1}^{M/R}\Big[F^{-1}_{p(z_j|\c'_j=1)}(\rho)^{c'_j}+F^{-1}_{p(z_j|\c'_j=0)}(\rho)^{(1-c'_j)}\Big].
    \end{gathered}
\end{equation}

Thus, we can apply the reparameterization trick to obtain samples from the latent variable $\z$ and optimize the reconstruction term of the \ac{elbo} with respect to the parameters $\t$ of the decoder.

The \ac{kl} Divergence term is approximated via Monte Carlo, drawing samples from $q_{\e}(\c|\x)$. Since it is not possible to backpropagate through discrete variables, we approximate the gradients with respect to the parameters of the encoder using the REINFORCE leave-one-out estimator \citep{reinforce_loo_1, reinforce_loo_2}, given by

\begin{equation}
    \begin{gathered}
        \widehat{g}_{LOO} = \\
        = \frac{1}{S-1}\Bigg[\sum_{s=1}^{S}f_{\e}\big(\z^{(s)},\c^{(s)}\big)\bigtriangledown_{\e}\log q_{\e}\big(\c^{(s)}|\x\big) - \overline{f}_{\e}\sum_{s=1}^{S}\bigtriangledown_{\e}\log q_{\e}\big(\c^{(s)}|\x\big)\Bigg],
    \end{gathered}
    \label{gloo}
\end{equation}

where 
\begin{equation}
f_{\e}(\z^{(s)},\c^{(s)}) = \log\Bigg(\frac{q_{\e}(\c^{(s)}|\x)}{p_{\t}(\x|\z^{(s)})p(\c^{(s)})}\Bigg),
\end{equation}

\begin{equation}
\overline{f_{\e}} = \frac{1}{S}\sum_{s=1}^{S}f_{\e}(\z^{(s)},\c^{(s)}).
\end{equation}

Defining a distribution over the codebook can seem intuitive, but scalability becomes challenging as the size of the codebook increases. The reason for this is that the posterior distribution must be evaluated for all codewords during both inference and test time. However, it can still provide a bound, enabling the utilization of more complex codes with theoretical guarantees that can outperform the previously proposed repetition codes.

\begin{algorithm}[!h]
   \caption{Training the model with inference at codeword level.}
   \label{training_word}
\begin{algorithmic}[1]
   \STATE {\bfseries Input:} training data $\x_i$, codebook.
   \REPEAT
   \STATE $q_{\e}^{u}(\c|\x_i) \gets$ forward encoder $g_{\e}(\x_i)$ 
   \STATE $q_{\e}(\c|\x_i) \gets$ evaluate $q_{\e}^{u}(\c|\x_i)$ over the codebook and normalize 
   \STATE $\tilde{\c} \gets $ sample from $q_{\e}(\c|\x_i)$
   \STATE $\z \gets$ modulate $\tilde{\c}$
   \STATE $p_{\t}(\x|\z) \gets$ forward decoder $f_{\t}(\z)$
   \STATE Compute ELBO according to \eqref{elbo_cvae_word}
   \STATE Compute encoder's gradients according to \eqref{gloo}
   \STATE $\t, \e \gets Update(ELBO)$ 
   \UNTIL convergence
\end{algorithmic}
\end{algorithm}

\section{Computational resources}\label{computational_resources}

All the experiments in this paper were conducted on a single GPU. Depending on availability, we used either a Titan X Pascal with 10GB of RAM, a Nvidia GeForce GTX with 10GB of RAM, or a Nvidia GeForce RTX 4090 with 24GB of RAM. Since training times varied significantly based on the hardware used, we were unable to provide comparable training times.

\section{Coded-DVAE scheme}

\begin{figure}[!h]
  \centering
  \includegraphics[width=1\linewidth]{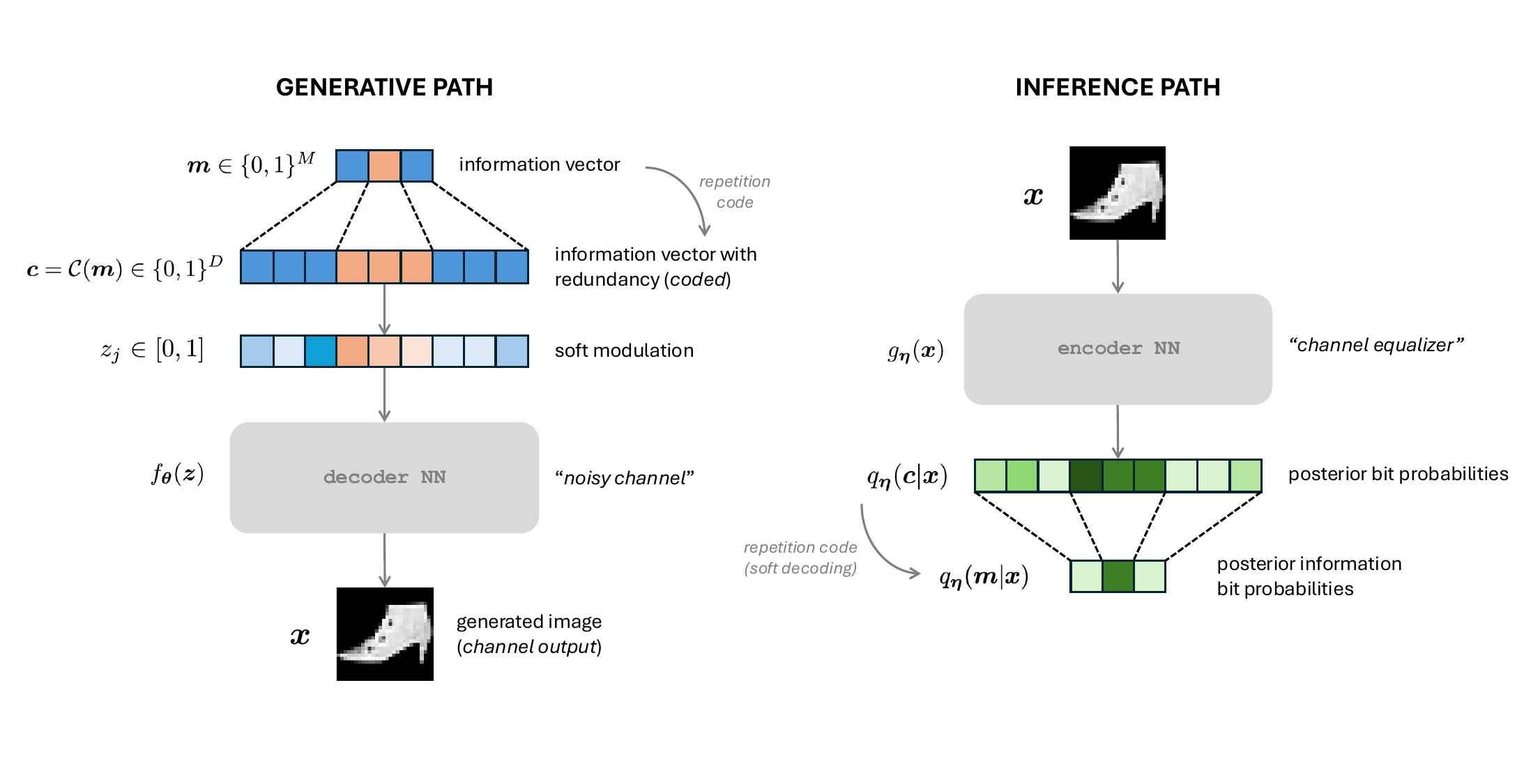}
  \caption{Graphic representation of the Coded-DVAE, illustrating both the generative and inference paths.}\label{fig:coded_scheme}
\end{figure}

\newpage
\section{Hierarchical Coded-DVAE scheme}

\begin{figure}[!h]
  \centering
  \includegraphics[width=1\linewidth]{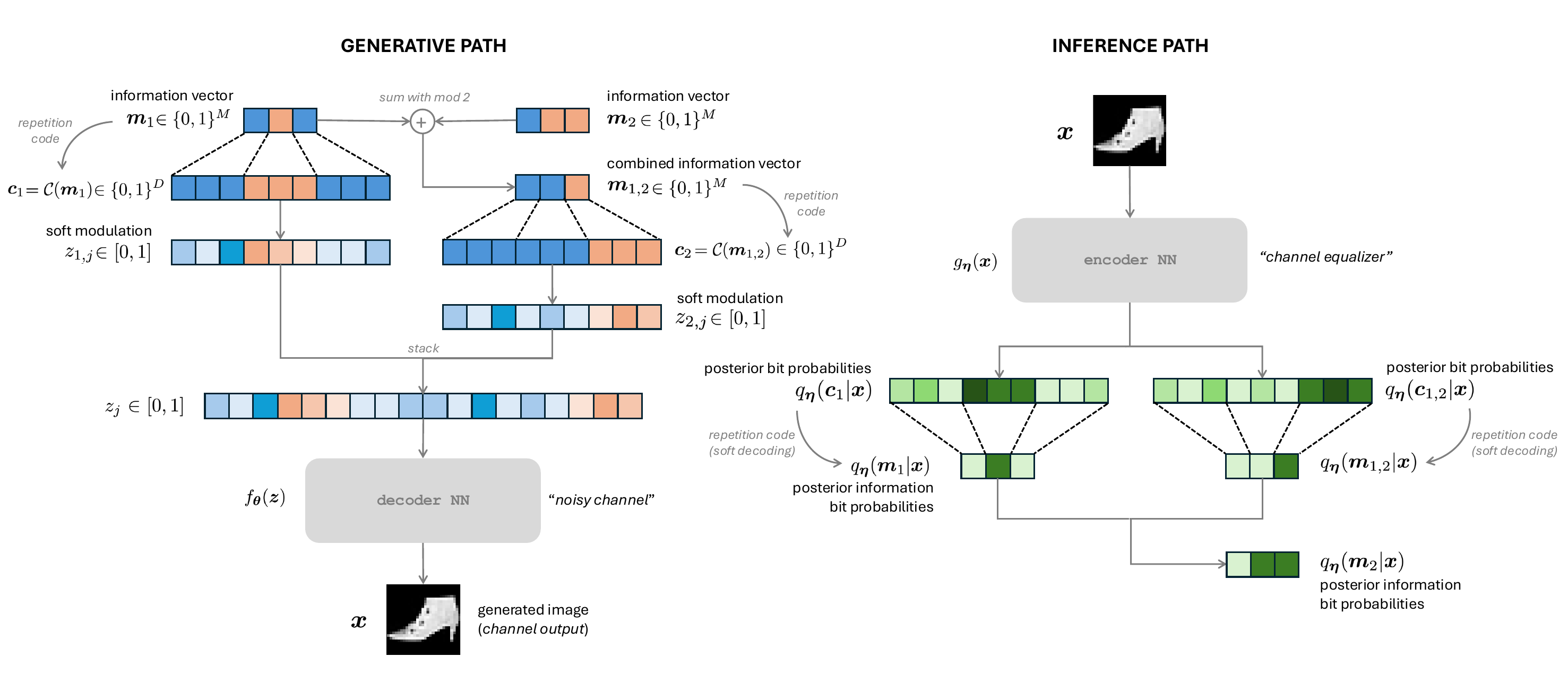}
  \caption{Graphic representation of the hierarchical Coded-DVAE, illustrating both the generative and inference paths.}\label{fig:hier_coded_scheme}
\end{figure}

\end{document}